\documentclass[lettersize,journal]{IEEEtran}
\usepackage{amsmath,amsfonts}
\usepackage{algorithmic}
\usepackage{algorithm}
\usepackage{array}
\usepackage[caption=false,font=footnotesize,labelfont=rm,textfont=rm]{subfig}
\usepackage{textcomp}
\usepackage{stfloats}
\usepackage{tabularx}
\usepackage{multirow} 
\usepackage{url}
\usepackage{verbatim}
\usepackage{graphicx}
\usepackage{pifont}
\usepackage{booktabs}
\usepackage{makecell}
\usepackage{arydshln}
\usepackage{bbding}
\usepackage{float}
\usepackage{color}
\usepackage{cite}
\begin{document}

\title{AMFD: Distillation via Adaptive Multimodal Fusion for Multispectral Pedestrian Detection}

\author{Zizhao Chen, Yeqiang Qian,~\IEEEmembership{Member,~IEEE,} Xiaoxiao Yang~\IEEEmembership{Member,~IEEE,}

Chunxiang Wang,~\IEEEmembership{Member,~IEEE,} and Ming Yang,~\IEEEmembership{Member,~IEEE}
\thanks{This work is supported by the National Natural Science Foundation of
China under Grant 62473253/62173228/62373250. (Corresponding author: Yeqiang Qian; Ming Yang.)}
\thanks{Zizhao Chen, Yeqiang Qian, Xiaoxiao Yang, Chunxiang Wang and Ming Yang are with the Department of Automation, Shanghai Jiao Tong University, Key Laboratory of System Control and Information Processing, Ministry of Education of China, Shanghai, 200240, China (email: qianyeqiang@sjtu.edu.cn; mingyang@sjtu.edu.cn).}}

\markboth{Journal of \LaTeX\ Class Files,~Vol.~14, No.~8, August~2021}%
{Shell \MakeLowercase{\textit{et al.}}: A Sample Article Using IEEEtran.cls for IEEE Journals}

\IEEEpubid{0000--0000/00\$00.00~\copyright~2025 IEEE}

\maketitle

\begin{abstract}
Multispectral pedestrian detection has been shown to be effective in improving performance in complex illumination scenarios. 
However, prevalent double-stream networks in multispectral detection employ two separate feature extraction branches for multi-modal data, leading to nearly double the inference time compared to single-stream networks utilizing only one feature extraction branch.
This increased inference time has hindered the widespread employment of multispectral pedestrian detection in embedded devices for autonomous systems. 
To efficiently compress multispectral object detection networks, we propose a novel distillation method, the Adaptive Modal Fusion Distillation (AMFD) framework. Unlike traditional distillation methods, the AMFD framework fully leverages the original modal features from the teacher network, thereby significantly enhancing the performance of the student network.
Specifically, a Modal Extraction Alignment (MEA) module is utilized to derive learning weights for student networks, integrating focal and global attention mechanisms. 
This methodology enables the student network to acquire optimal fusion strategies independent from that of teacher network without necessitating an additional feature fusion module. Furthermore, we present the SMOD dataset, a well-aligned challenging multispectral dataset for detection.
Extensive experiments on the challenging KAIST, LLVIP, SUNRGB-D and SMOD datasets are conducted to validate the effectiveness of AMFD. The results demonstrate that our method outperforms existing state-of-the-art methods in both reducing log-average Miss Rate and improving mean Average Precision. The code is available at https://github.com/bigD233/AMFD.git.

\end{abstract}


\section{Introduction}
\IEEEPARstart{P}edestrian detection is a crucial problem in computer vision, with applications ranging from autonomous vehicles\cite{yang2018real} to surveillance systems\cite{bilal2016low}. 
Modern studies using visible images perform well under regular light conditions. 
However, due to susceptibility to lighting conditions, visible light exhibits poor detection performance in complex low-light scenarios.
To mitigate this limitation, thermal infrared images are introduced to provide supplementary data, leading to the exploration of multispectral pedestrian detection\cite{hwang2015multispectral} as a potent solution.
\begin{figure}
  \centering
  \includegraphics[width=\linewidth]{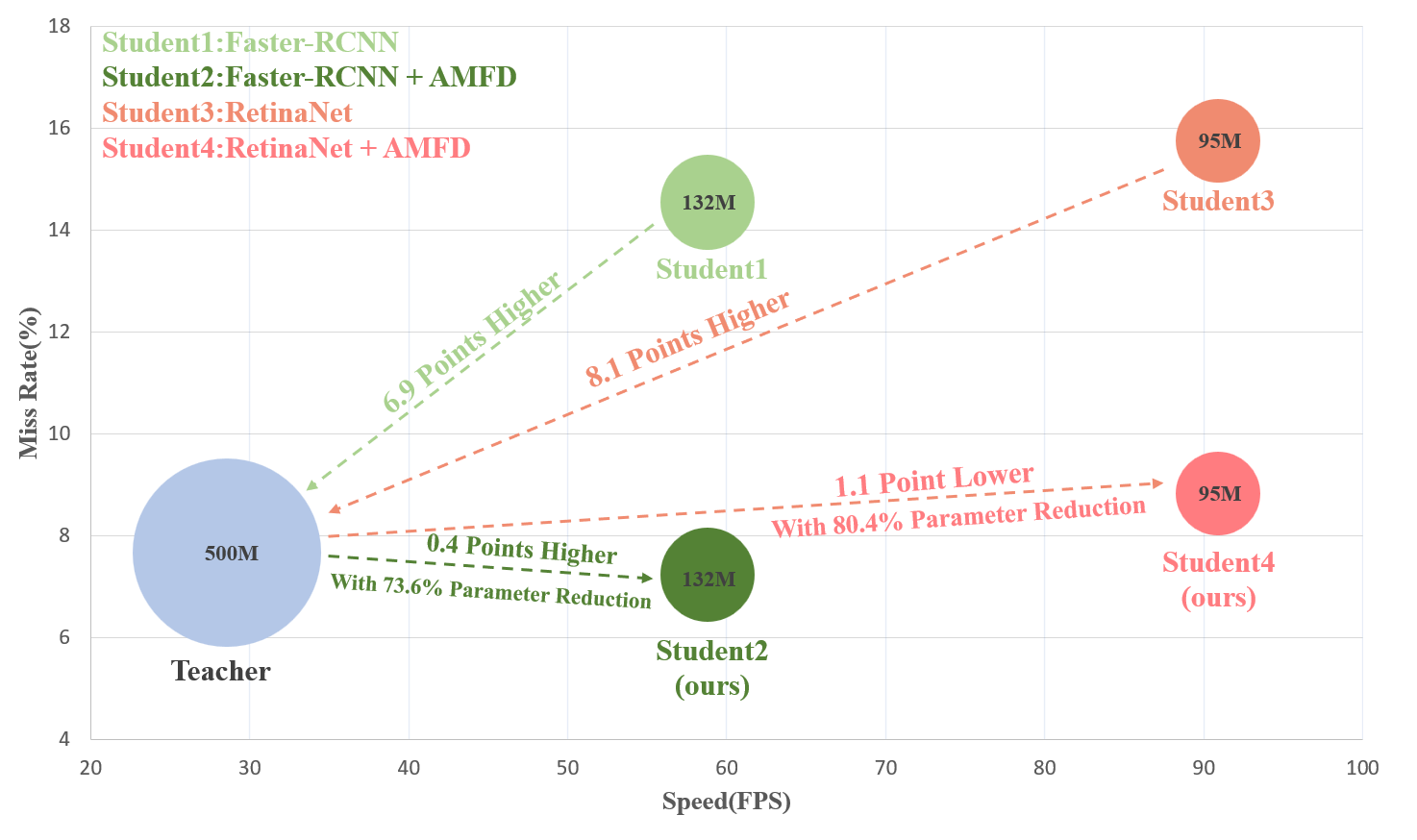}
  \caption{The experimental results of pedestrian detection on KAIST\cite{hwang2015multispectral} dataset. The teacher is a two-stream network with a complex fusion module and ResNet50 backbone\cite{2016Deep}. 
  The students are single-stream networks(Faster-RCNN\cite{2017Faster} and RetinaNet\cite{lin2017focal}) with simple image-level fusion and ResNet18 backbone\cite{2016Deep}. }
  \label{fig:res}
\end{figure}

Multi-modal feature fusion is the key to multispectral detection. 
Previous works\cite{xu2017learning, zhang2021guided, chen2022multimodal, yang2022baanet, chen2023attentive, zhang2023tfdet, li2023multiscale, li2022confidence} have explored various fusion strategies: early, mid, and late fusion. 
In the mid-fusion framework, thermal and RGB features are extracted independently and subsequently fused at the intermediate stage of the two-stream network.
Recent research employing the mid-fusion strategy has shown superior performance in multispectral detection. 
Nevertheless, the adoption of mid-fusion strategies in two-stream networks incurs significant computational overhead, while nearly doubling the inference speed compared to single-stream networks, thereby posing challenges for deployment on embedded devices\cite{kruthiventi2017low}. 

\begin{figure}[h]
  \centering
        \subfloat[RGB image input]{
        \includegraphics[width=0.415\linewidth]{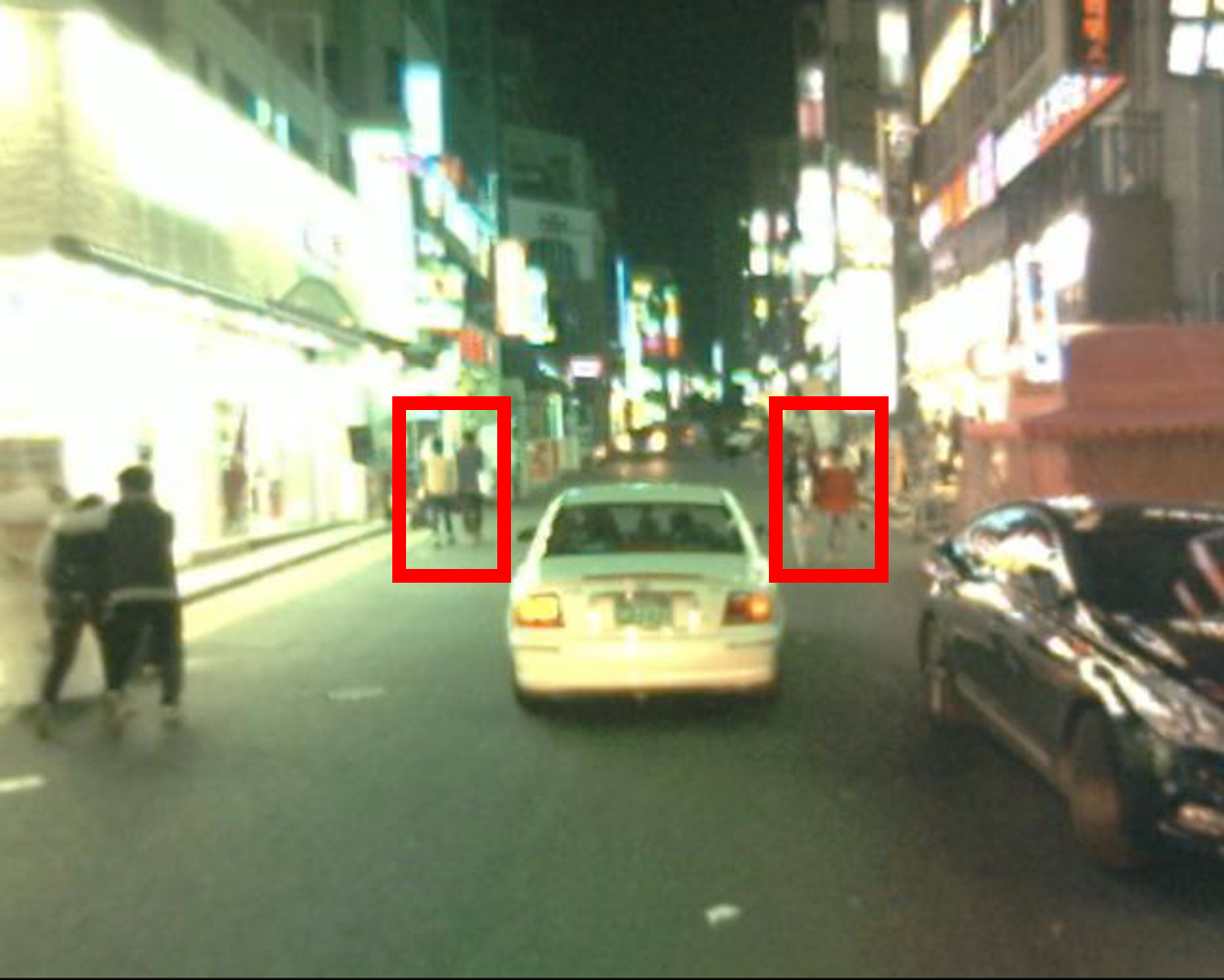}%
        \label{fig: sub_figure1}
        }
	\subfloat[teacher network]{
        \includegraphics[width=0.45\linewidth]{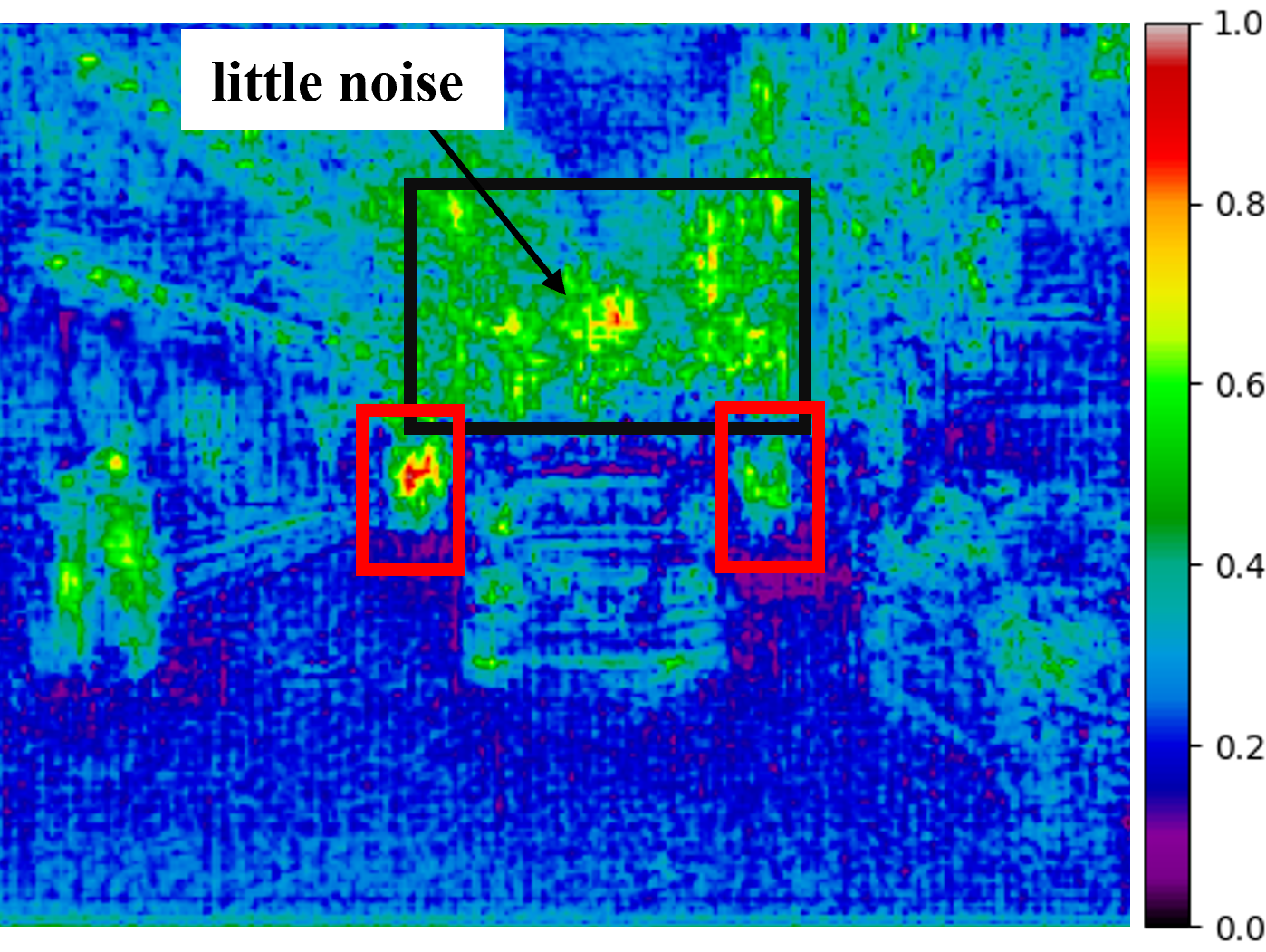}%
        \label{fig: sub_figure2}
        }
        
        \subfloat[distill fusion feature(student)]{
        \includegraphics[width=0.45\linewidth]{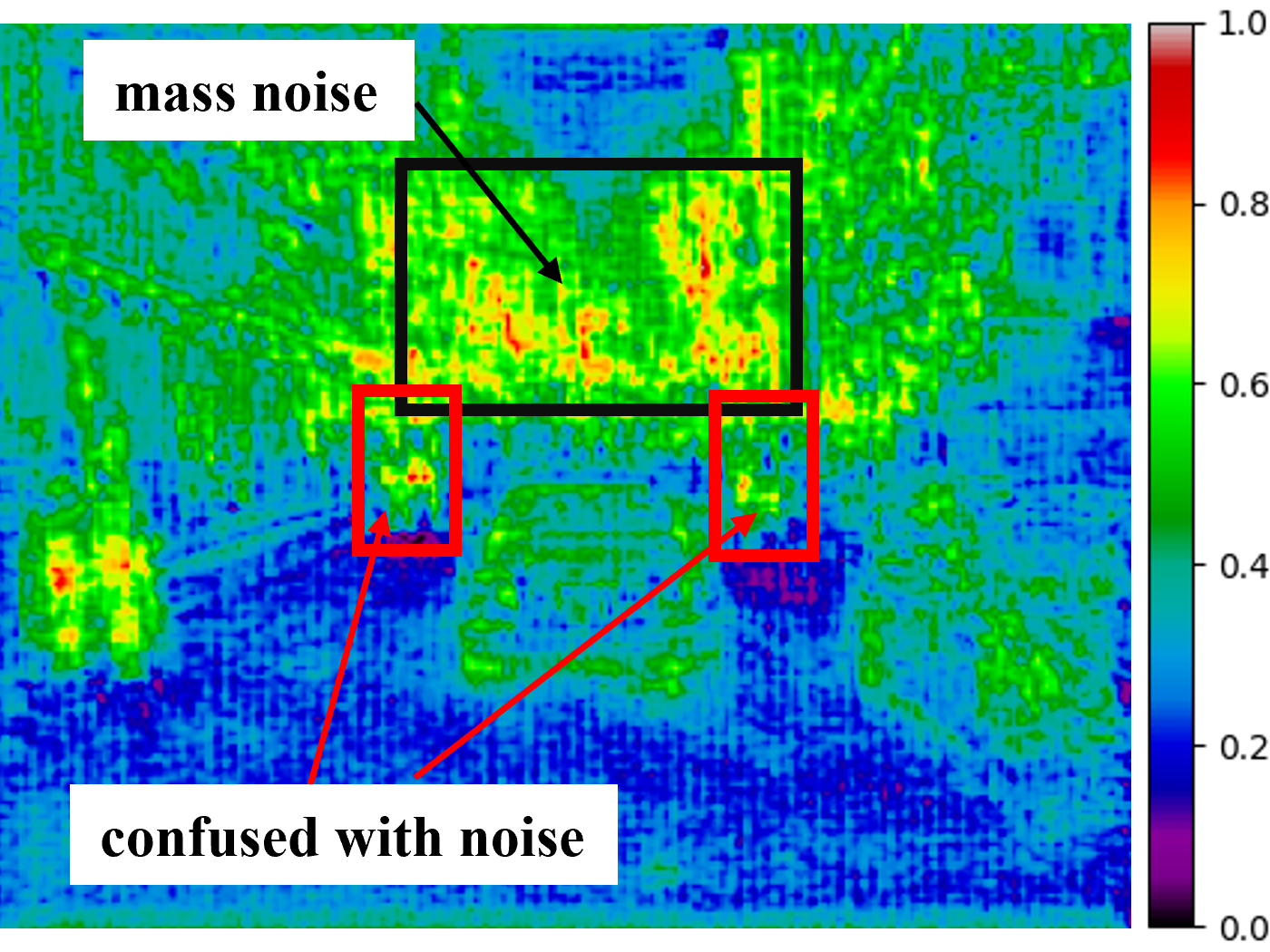}%
        \label{fig: sub_figure3}
        }
        \subfloat[ours(student)]{
        \includegraphics[width=0.45\linewidth]{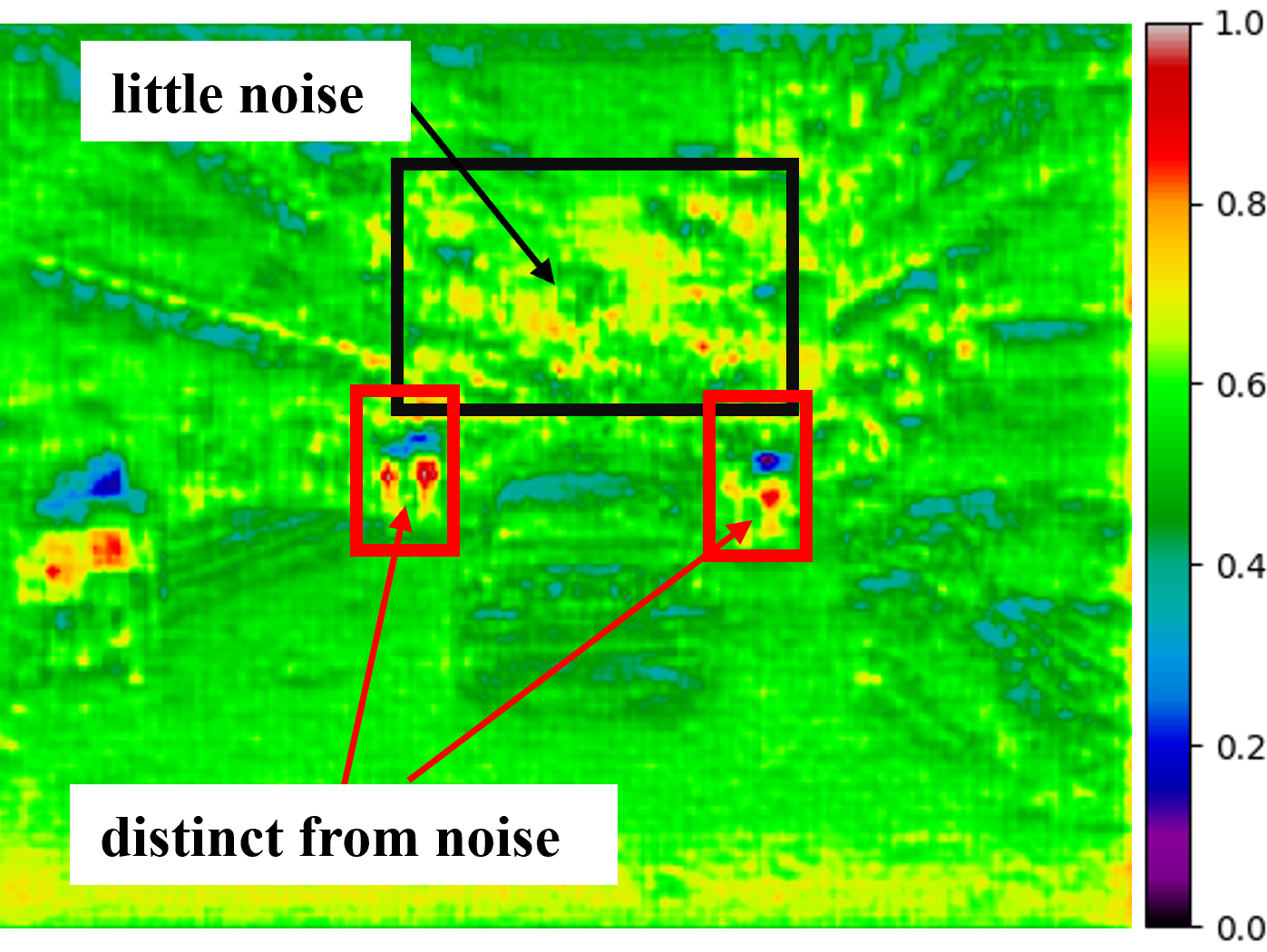}%
        \label{fig: sub_figure4}
        }
	\caption{
 Based on the fusion feature maps of different networks, we  obtain the spatial attention of the feature map.
 We can see that (c) is similar to (b), but the noise in the black box is not suppressed. Noise makes pedestrians in the two red boxes in figure (a) not easily recognizable in figure (c).  Our method shown in (d) no longer follows the fusion strategy of teacher can well represent the pedestrian features.}
	\label{spatial_attention}
\end{figure}

\IEEEpubidadjcol

Lightweight single-stream networks can infer in real time but lack the ability to handle complex scenarios. Complex two-stream networks have good performance, but are not easy to deploy and use in real time. In search of a balance between time and performance, many researchers focus on model compression through knowledge distillation, a technique aimed at transferring information from a frozen large teacher network to a compact training student network, thereby reducing model inference time.
Recently, many feature-based knowledge distillation methods\cite{luo2018graph, garcia2019learning, liu2021deep, zhang2022low} have been proposed for the field of multispectral object detection. 
However, these methods face a challenge in transferring the knowledge of the fusion feature from specific modules of the teacher network to the student network. 
Since complex feature fusion modules are often present in teacher networks but not in simple student networks, this increases the capacity gap\cite{mirzadeh2020improved, gao2021residual} between teacher and student networks. Moreover, direct distillation of the fusion features ignores much potentially useful information in the original modal features of the teacher network.
As a result, the student network struggles to effectively assimilate the fusion strategy of the teacher network.
As shown in Fig.\ref{fig: sub_figure3}, directly learning the fusion features of the teacher network will introduce more noise into the student network, indicating that the fusion strategy suitable for the teacher network may not necessarily be suitable for the small-capacity student network.

To address the above issues, we propose an \textbf{A}daptive \textbf{M}odal \textbf{F}usion \textbf{D}istillation framework (AMFD). 
Specifically, a fusion distillation architecture is introduced.
Instead of directly distilling fusion features from the teacher network to the student network, we simultaneously distill the thermal and RGB features of the teacher network to the fusion features of the student network. 
This approach aims to enable the student network to develop its own efficient fusion strategy during training. Drawing inspiration from GcBlock\cite{cao2019gcnet}, we design the modal extraction alignment (MEA) module, which obtains distillation losses based on focal and global attention mechanisms. The fusion features of the student network are learned from thermal and RGB features through distillation losses calculated by the MEA module. 
This approach liberates the student network from the constraints of the teacher network and significantly enhances the performance of the student network.

In summary, our contributions are as follows.

$\bullet$ We innovatively propose the Adaptive Modal Fusion Distillation method, which front-loads the distillation location through the fusion distillation architecture. Two Modal Extraction Alignment (MEA) modules are also used to extract infrared and visible modal features, respectively, so that the student network can efficiently generate fusion strategies independent of the instructor network, thus efficiently compressing the teacher network to a compact student network, which reduces the inference speed.

$\bullet$ We present the SJTU Multispectral Object Detection (SMOD) dataset for detection. Within this dataset, 8042 pedestrians, 10478 riders, 6501 bicycles, and 6422 cars are annotated. The degree of occlusion of all objects is meticulously annotated. The low sampling rate dataset has dense rider and pedestrian objects and a very large number of challenging and complex occlusion scenarios. Our dataset is available on Kaggle\footnote{https://www.kaggle.com/datasets/zizhaochen6/sjtu-multispectral-object-detection-smod-dataset}. 

$\bullet$ 
 The experiments on KAIST\cite{hwang2015multispectral}, LLVIP\cite{jia2021llvip} and our SMOD datasets show that student networks can achieve excellent distillation results without adding additional modules. 
 Our AMFD distillation method on the KAIST\cite{hwang2015multispectral} dataset can reduce the log-average miss rate ($\textit{MR}^{-2}$) of the student network to 0.4\% lower than that of the teacher network, while the inference time is reduced to half of that of the teacher network. 
 On the LLVIP\cite{jia2021llvip} dataset, our method improves the COCO Mean Average Precision (mAP) of the student network by 2.7\%. 
On the SMOD and SUNRGB-D\cite{song2015sun} dataset, we validate the effectiveness and flexibility of the fusion distillation architecture.

\section{Related WORK}

\begin{table*}[h]
\centering
\caption{Comparison of SMOD and existing datasets including FLIR Thermal Dataset, CVC-14, KAIST Multispectral Dataset, LLVIP Dataset and $\text{M}^3\text{FD}$ Dataset. The “Person Density” refers to the number of pedestrians and cyclists higher than 35 pixels per image. The “Occ(FG)” and "Occ(\%)" refer to whether the dataset is annotated with a fine-grained degree of occlusion and occlusion object percentage, respectively.}
\label{tab:compare_dataset}
\begin{tabular}{@{}lccccccccc@{}}
\toprule
Dataset & Img Pairs & Sampling Rate & Year & Driving & Aligned & Occ(FG) & Occ(\%) & Person Density & Bicycle Density \\ \hline
LLVIP & 16836 & 1FPS & 2021 & \XSolidBrush & \Checkmark & \XSolidBrush & \XSolidBrush &2.51 & \XSolidBrush\\ 
$\text{M}^3\text{FD}$ & 4200 &\textbf{ --} & 2022 & \XSolidBrush & \Checkmark & \XSolidBrush & \XSolidBrush & 1.22 &  0.11 \\
\hdashline
\rule{0pt}{11pt}CVC-14 & 8518 & 10FPS & 2016 & \Checkmark & \XSolidBrush & \XSolidBrush & \XSolidBrush & 0.80  & \XSolidBrush \\
FLIR & 10228 & 30FPS &2018 & \Checkmark & \XSolidBrush & \XSolidBrush & \XSolidBrush & 1.52 &  0.43 \\
KAIST & 95328 &20FPS&  2015 & \Checkmark & \Checkmark & \XSolidBrush & 21.4\% & 0.62 & \XSolidBrush \\
\textbf{SMOD} & 8676 & \textbf{2.5FPS} & \textbf{2024} & \Checkmark & \Checkmark & \Checkmark & \textbf{49.2\%} &\textbf{2.14} & \textbf{0.79} \\
   \bottomrule
\end{tabular}
\end{table*}

\subsection{Multispectral Pedestrian Detection}
Multispectral pedestrian detection, which combines RGB and thermal images, has attracted considerable attention for its ability to maintain robust detection performance under varying lighting conditions. 
Recent research has concentrated on refining multispectral feature fusion techniques. 
Zhou et al. proposed MBNet\cite{zhou2020improving} to mitigate the imbalance between different modalities. 
GAFF\cite{zhang2021guided}, introduced by Zhang et al., integrates multispectral features using both inter- and intra-modality attention mechanisms. 
BAANet\cite{yang2022baanet} leverages modality correlation to recalibrate attention mechanisms for two modalities. 
MSDS-RCNN\cite{li2018multispectral} enhances pedestrian detection performance by simultaneously conducting pedestrian detection and segmentation. 
DCMNet\cite{xie2022learning} focuses on learning both local and non-local information within multispectral features. 
Some studies employing late fusion strategies have also demonstrated remarkable performance. 
For example, ProbEn3\cite{chen2022multimodal} explores late fusion of single-modal detections through detector ensembling. 
However, these methods primarily rely on two-stream networks with various additional modules, inevitably increasing the inference time and posing challenges for deployment on embedded devices.

\subsection{Knowledge Distillation}

Knowledge distillation (KD), pioneered by Hinton et al.\cite{hinton2015distilling}, is a crucial technique for model compression while preserving network architecture. FitNet\cite{adriana2015fitnets} and Zagoruyko et al.\cite{zagoruyko2016paying} have, respectively, highlighted the role of intermediate layer semantics and the use of unsupervised attention maps in guiding student models. KD's applications extend beyond image classification to object detection. For example, Wang et al.\cite{wang2019distilling} introduced fine-grained masking to distill regions delineated by ground-truth bounding boxes, while Guo et al.\cite{guo2021distilling} highlighted the significance of distilling both foreground and background information separately, leading to improved student performance. 

For multispectral detection, Liu et al.\cite{liu2021deep} proposed a method that employs distinct distillation losses at multiple levels, encompassing the perspective of feature, detection, and segmentation. Zhang et al.\cite{zhang2022low} designed a knowledge transfer module to transfer knowledge from the fusion features extracted by GAFF\cite{zhang2021guided} of the teacher network.  However, these methods often rely on teacher network fusion features and additional student network modules, contrary to the goal of minimizing inference time. Our framework diverges by distilling original modal features, enabling robust student network performance without added inference latency.

\subsection{Multispectral Pedestrian Dataset}

In recent years, many multispectral pedestrian detection datasets have been proposed. Both CVC-14\cite{gonzalez2016pedestrian} and FLIR are multispectral datasets proposed for automated driving pedestrian detection tasks. However, the visible and infrared image pairs in these two datasets are not well aligned spatially and temporally. The KAIST\cite{hwang2015multispectral} dataset provides well-aligned picture pairs, annotated pedestrian occlusions, and is now widely used for multispectral pedestrian detection tasks. However, its original data sampling rate is too high, resulting in little variation in consecutive frame images, thus requiring data cleaning when in use. Furthermore, the KAIST dataset lacks a quantitative definition for the annotation of the degree of occlusion. In addition to the above dataset in driving angle, LLVIP is a well-aligned multispectral dataset in surveillance angle. Most of the images in the LLVIP dataset are collected from a medium distance, so the dataset lacks small-target pedestrian. The $\text{M}^3\text{FD}$ dataset\cite{liu2022target} covers four major scenarios with various environments, having a wide range of pixel variations. However, this also results in a low number of pedestrians in the data.

In addition to this, we find that the density of "effective" person (defined as pedestrians with a height higher than 35 pixels) in the dataset in the driving view are all relatively low. Tab.\ref{tab:compare_dataset} shows the details of these datasets and our proposed SMOD dataset. As shown in Table 1, the SMOD dataset with dense "effective" person objects contains a large number of campus road scenes with complex occlusions, and we therefore perform fine-grained annotations on the occluded objects. This facilitates the validation of the robustness and adaptability of the pedestrian detection model.

\begin{figure*}[h]
  \includegraphics[width=\textwidth]{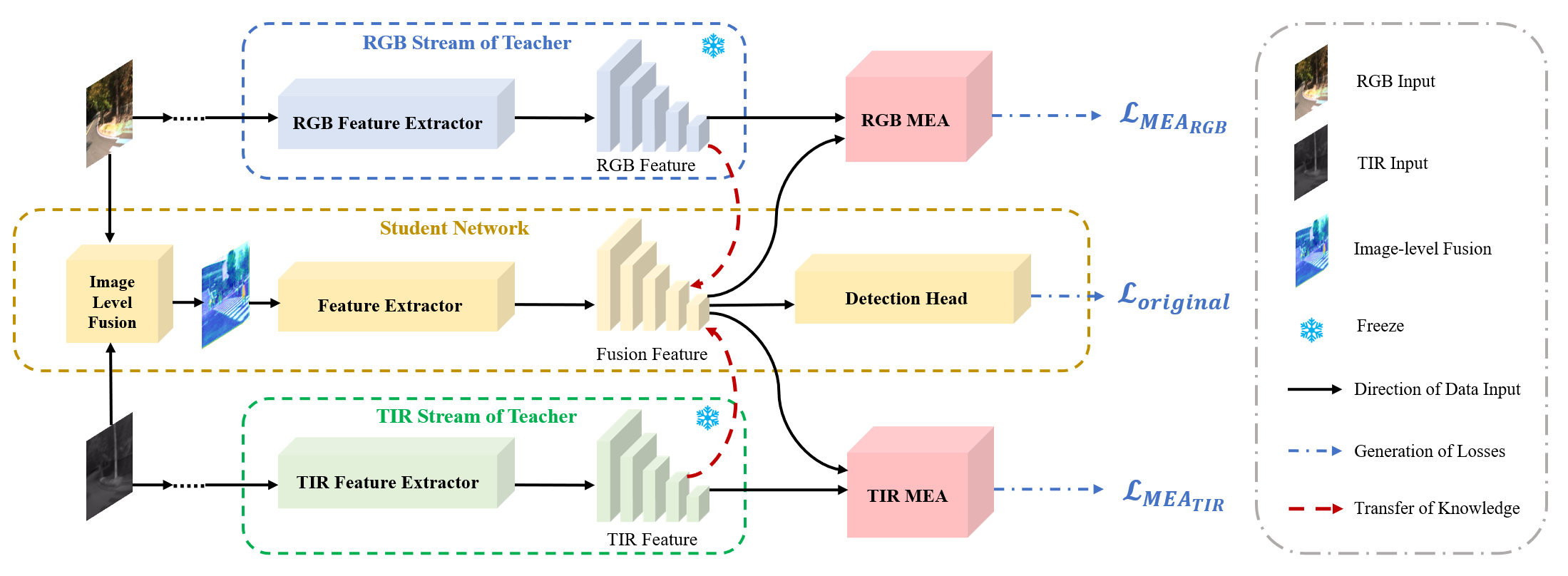}
  \caption{The overall architecture of the proposed distillation framework. 
  Firstly, we use a frozen two-stream network with complex feature extractor as the teacher network. The student network is a single-stream network with a simple image-level fusion module. 
  Then during training, the framework distills the knowledge of RGB and TIR features of the teacher network into the fusion features of the student network through the fusion distillation architecture. 
  The fusion distillation architecture contains two modal extraction alignment modules (MEA) to adaptively extract the difference between the fusion feature of the student network and the RGB, TIR feature of the teacher network. 
  At last, the student network is optimized by two MEA losses produced by MEA and the original detection loss. }

  \label{fig:overall}
\end{figure*}

\section{method}
\subsection{Overall Architecture}
As shown in Fig.\ref{fig:overall}, the two-stream teacher network has two feature extractors, which contains the backbone and feature pyramid networks (FPN)\cite{lin2017feature}. 
The teacher network is recognized as having excellent feature extractor and fusion feature. In contrast, the student network has only one feature extractor compared to the teacher network. The input of the backbone is the fusion of RGB and thermal infrared (TIR) images. 
Notably, the single stream with simple image-level fusion input has been shown to perform poorly in multispectral pedestrian detection. Hence, our objective is to enhance the performance of student networks with small capacity.

The distillation via Adaptive Modal Fusion (AMFD) focuses on the RGB and TIR features. To exploiting potentially useful information in original modal features and efficiently guide the student network in formulating a fusion strategy, we leverage a fusion distillation architecture. This approach involves distilling multi-modal feature knowledge from the teacher network into the fusion feature of the student network. The Modal Extraction Alignment (MEA) module aligns the feature maps using global and focal attention mechanisms. These MEA modules dynamically generate channel-wise weights to learn the useful part of the RGB and TIR features. Then the student network is optimized by the MEA losses and the original detection loss.  Consequently, student networks can generate more suitable fusion strategies independently of teacher networks.

\begin{figure}[h]
  \centering
  \includegraphics[width=\linewidth]{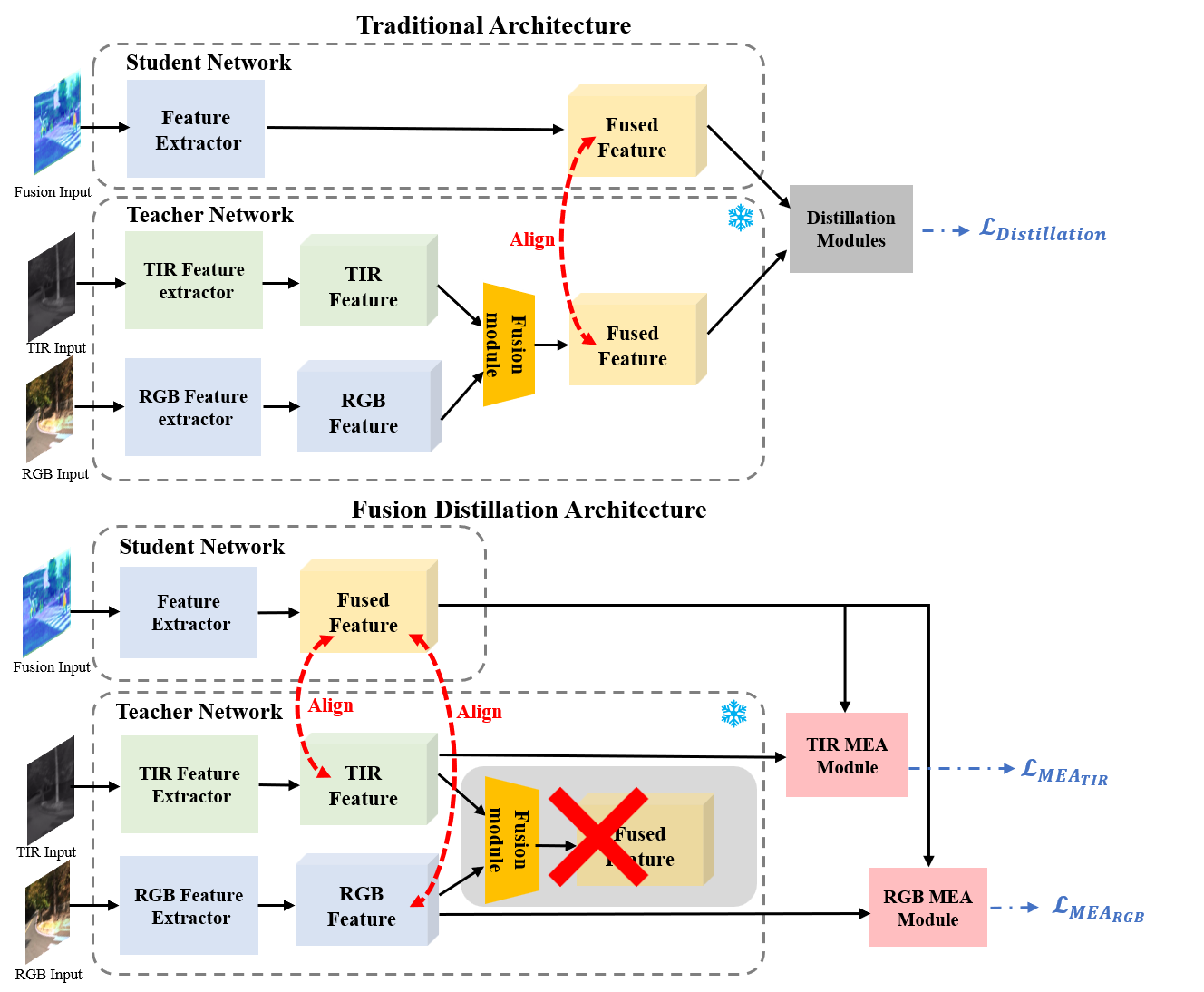}
  \caption{The comparison between fusion distillation architectures and traditional architecture. The position of the feature for distillation is advanced from the fusion feature to the original modal feature.}
  \label{fig:multi-to-one}
\end{figure}

\subsection{Fusion Distillation Architecture}

Typically, traditional methods select features at the same stage for distillation. That is, the student network will mimic the features generated by identical modules of the teacher network during the distillation process. Thus, distilling the fusion features of the teacher network to the fusion features of the student network naturally becomes the dominant distillation architecture. Based on this architecture, many approaches\cite{liu2021deep}, \cite{zhang2022low} will add some specific modules to the single-stream student network to improve distillation results. However, this situation limits the structure of the student network and also increases the complexity of the student network, which is not conducive to its deployment.

To maintain the flexibility of the student network structure while allowing it to fully utilize the original modal features, we propose a fusion distillation architecture. As shown in Fig.\ref{fig:multi-to-one}, the fusion features of the student network are simultaneously aligned with the TIR and RGB features of the teacher network. We denote the RGB feature and the TIR feature, which is the output of its feature pyramid network(FPN\cite{lin2017feature}) as $x_R$ and $x_T$. The $\mathcal F(\cdot)$ denotes feature fusion module while the $\mathcal L(\cdot)$ denotes the loss function of the distillation. The distillation loss of the traditional architecture is defined as
\begin{equation}
  L_{traditional}  =  \mathcal L(\mathcal F(x_R, x_T),\widetilde{x}_F),
\end{equation}
where the $\widetilde{x}_F$ is the fusion feature of the student network. While the fusion distillation architecture no longer relies on the feature fusion module of the teacher network. The distillation loss of the fusion distillation architecture is:
\begin{equation}
  L_{fusion}  =  \mathcal L(x_R, \widetilde{x}_F) + \mathcal L(x_T, \widetilde{x}_F).
\end{equation}

This loss indicates that during distillation, the fusion features of the student network must emulate both the TIR and RGB features. In contrast to the distillation loss of the traditional architecture, this loss will not lose potentially useful information during feature fusion. Consequently, the student network can fully leverage unfused modal features to devise fusion strategies independent of those of the teacher networks.

\begin{figure*}[h]
\centering
\subfloat[]{\includegraphics[height=2.5in]{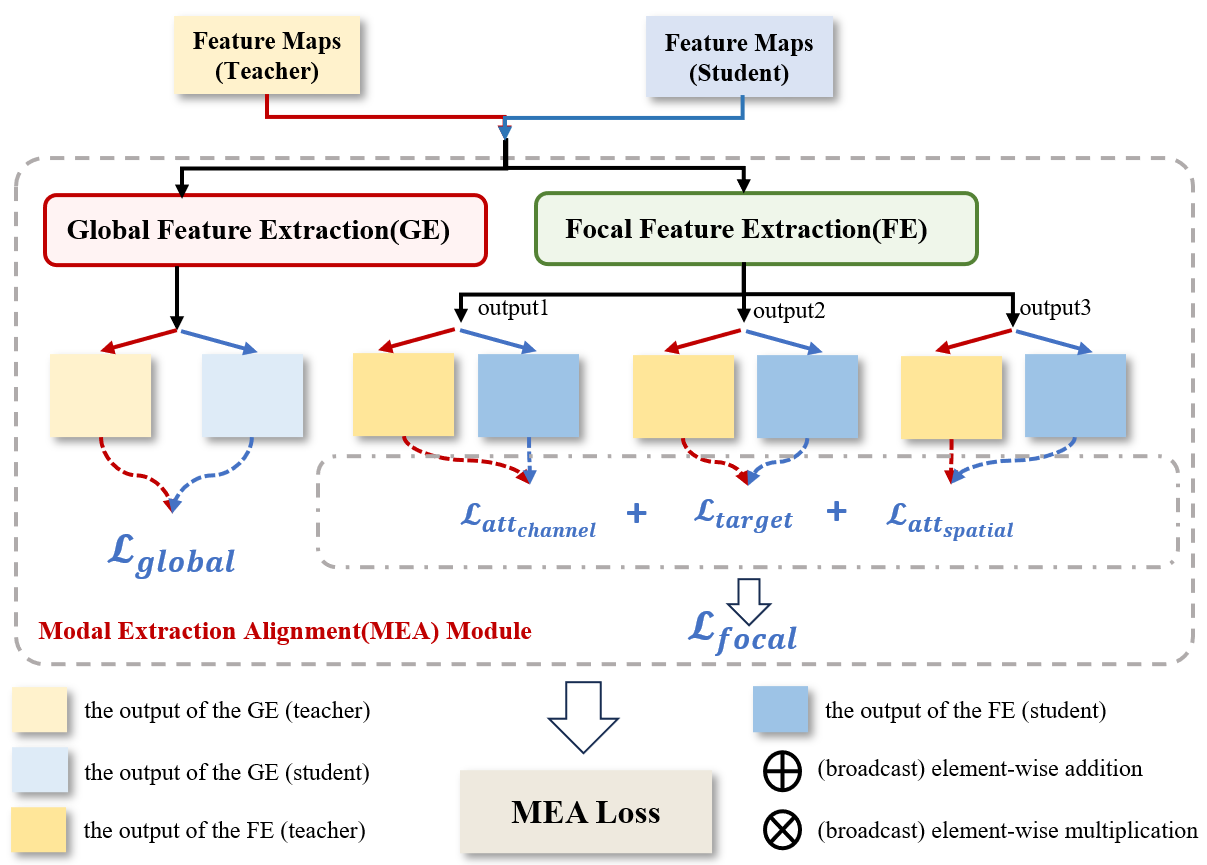}%
\label{fig:MEA_general}}
\hfil
\subfloat[]{\includegraphics[height=2.5in]{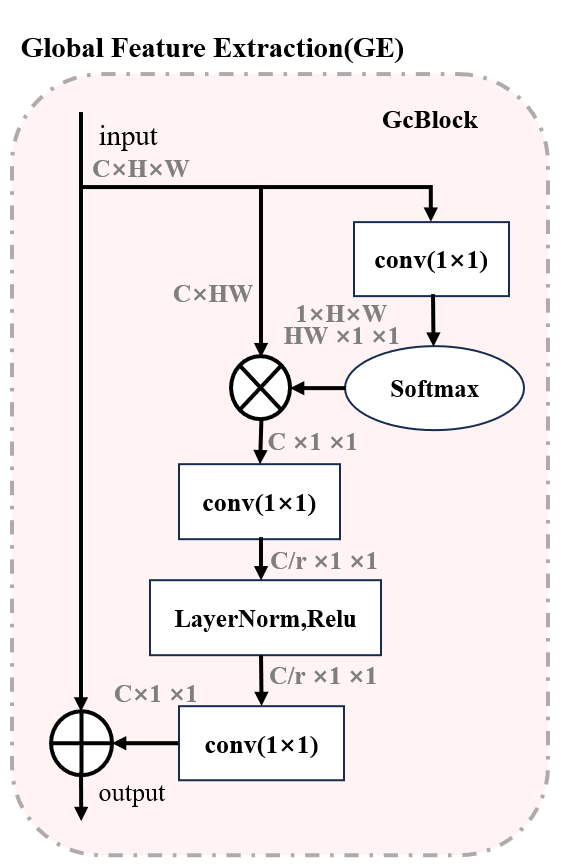}%
\label{fig:GE}}
\subfloat[]{\includegraphics[height=2.5in]{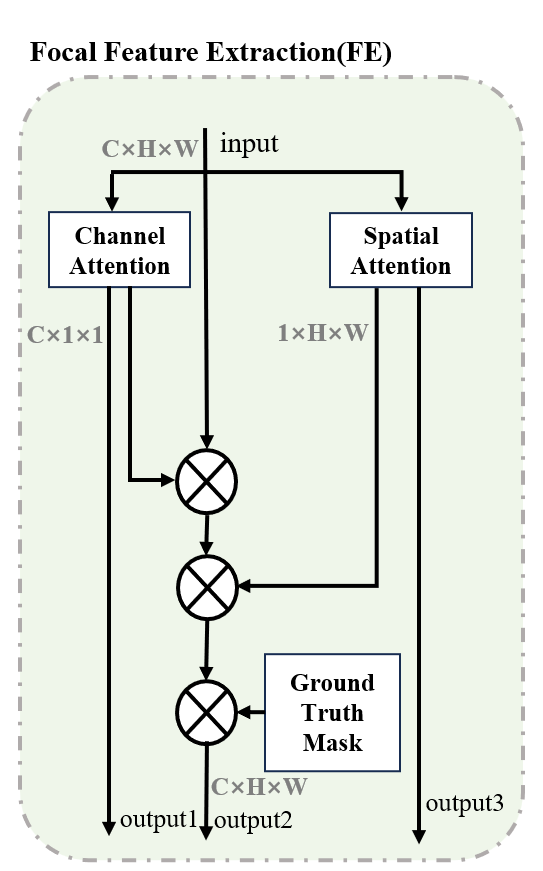}%
\label{fig:FE}}
\caption{The structure of our modal extraction alignment (MEA) module. (a) shows the overall structure of the MEA module. Both input features go through two modules, Global Feature Extraction (GE) and Focal Feature Extraction (FE), which ultimately form a MEA loss consisting of the global loss and focal loss. (b) and (c) show the details of the GE and FE. GE generates a $C\times1\times 1$ weight and this weight is added to the original feature map by broadcast channel-wise addition, aiming to obtain a feature map with better global relation.}
\label{fig:MEA}
\end{figure*}

\subsection{Modal Extraction Alignment Module (MEA)}

Designing an effective distillation loss is a key issue in the field of object detection distillation. The extreme imbalance between the foreground and the background is a key issue in the field of object detection\cite{lin2017focal} and distillation. To address this issue, we extract the global and focal knowledge separately. As shown in Fig.\ref{fig:MEA}, the modal extraction alignment module (MEA) consists of two parts: Global Feature Extraction and Focal Feature Extraction. In the following subsection we will introduce these two modules, respectively.

\subsubsection{\textbf{Global Feature Extraction}}

Global information distillation is the most direct and efficient distillation method. This permits maximum retaining of global information and also allows more globally relevant dark knowledge to be transferred from the teacher network to the student network. However, the features of the student network must be aligned with the features of multiple modalities at the same time because of the fusion distillation architecture. This requires us to pay attention to different information according to different modalities in the global distillation. Hence, it is important to merge different information in the global distillation corresponding to different modalities.

To enhance the extraction of global relations, we use GcBlock\cite{cao2019gcnet} to ensure that the student network possesses the same global relations as the teacher network. The GcBlock here is to extract important global relations in different modalities and make the student network adaptively learn these relations. The GcBlock will generate a channel-wise weight:
\begin{equation}
    \mathcal{W}(x) = \mathcal F_{conv3}(\mathcal L \mathcal R(\mathcal F_{conv2}( \sum_{j=1}^{N_p} \frac{e^{\mathcal F_{conv1}(x_j)}}{\sum_{m = 1}^{N_p} e^{\mathcal F_{conv1}(x_m)}} x_j)))  .
    \label{equ: channel_weights}
\end{equation}
And add this weight to the input feature map, then we get the output feature map with global relations:
\begin{equation}
  \mathcal G(x) =x \oplus \mathcal{W}(x)  ,
\end{equation}
where $\mathcal F_{conv}$ denotes the 1×1 convolution layers, $\mathcal L \mathcal R$  denotes the layer normalization and relu activation. $N_p$ is the number of pixels of the feature $x$, and $x_j,x_m$ is the slice of the feature map ($C\times 1\times 1$, C is the number of channels in the feature map). $\oplus$ denotes the broadcast element-wise addition. 

So for the RGB, TIR feature maps $x_R$, $x_T$ of teacher network and the fusion feature $\widetilde{x}_F$ of the student network, the global feature extraction loss is:
\begin{align}
    & \mathcal L_{global_{TIR}}  = \lambda_1 \sum (\mathcal G_1(x_T) - \mathcal G_1(\widetilde{x}_F))^2,  \\
    & \mathcal L_{global_{RGB}}  = \lambda_2 \sum (\mathcal G_2(x_R) - \mathcal G_2(\widetilde{x}_F))^2,  \\
    & \mathcal L_{global}  =  \mathcal L_{global_{TIR}} + \mathcal L_{global_{RGB}}\quad \text{,}
\end{align}
where $\mathcal G_1, \mathcal G_2$ are the GcBlocks in two MEA modules used to distill the TIR and RGB feature of the teacher network respectively. $\lambda_1$ and $\lambda_2$ are hyper-parameters to balance the loss.

\subsubsection{\textbf{Focal Feature Extraction}}

Focusing solely on global relations is obviously insufficient. The global distillation process carries the risk of introducing noise, which could adversely affect the features of the target area. To enable the student network to acquire more precise knowledge about the features of the target area, we emphasize crucial pixels and attention maps in the feature map. Building upon this, we introduce focal feature extraction.

Obviously the areas we focus on are areas where pedestrians are present. So we select the area based on the ground-truth bounding boxes. These areas are marked by a mask $M$:
\begin{equation}
M_{i,j} = \left\{
\begin{aligned}
& \frac{1}{h_b w_b} & \text{, if}(i,j)\in b\\
& 0 & \text{, Otherwise}\\
\end{aligned}
\right.,
\end{equation}
where the $b$ denotes the ground-truth boxes corresponding to the feature map scale, $h_b, w_b$ are the height and width of the ground-truth box with the largest area that this pixel belongs to. The weight $\frac{1}{h_b w_b}$ balances the composition of the losses. This way, the losses generated for targets with larger areas do not account for the majority of the losses.


The mask $M$ selects the regions that are important to us. However, the weights for the areas within a ground-truth box are uniform both spatially and across channels. Inspired by the attention mechanism, we will calculate the focal feature loss according to the channel and spatial attention map. This involves calculating the absolute mean values on both pixel and channel dimensions, followed by generating weights using softmax.

\begin{align}
    A^S(x) & = H\cdot W \cdot softmax(\frac{1}{C}\sum_{c =1}^C  |x_{c}|) , \\
    A^C(x) & = C\cdot softmax(\frac{1}{HW}\sum_{i =1}^H \sum_{j=1}^W |x_{i,j}|),
\end{align}
where the $A^S(x),A^C(x)$ are the spatial and channel attention maps of the feature $x$. Then the loss after spatial, channel attention and mask weighting of target area is:

\begin{align}
     \mathcal L_{target_{RGB}} & = \alpha_1 \sum_{k =1}^C \sum_{i =1}^H \sum_{j=1}^W M_{i,j} A_{i,j}^{S_R} A_k^{C_R} (x_R - \widetilde{x}_F)^2 ,\\
      \mathcal L_{target_{TIR}} & = \alpha_2 \sum_{k =1}^C \sum_{i =1}^H \sum_{j=1}^W M_{i,j} A_{i,j}^{S_T} A_k^{C_T} (x_T - \widetilde{x}_F)^2,
\end{align}
where $\alpha_1$ and $\alpha_2$ are hyper-parameters to balance the loss. Note that $A^{S_R}, A^{C_R}$ and $A^{S_T}, A^{C_T}$ are the spatial and channel attention for RGB and TIR feature of the teacher network, respectively.

Although we obtain the loss of important regions using spatial and channel attention of the teacher network, we do not constrain spatial and channel attention of the student network. In order for the student network to learn spatial attention and channel attention of the two modalities of the teacher network in a integrated manner, we design the attention loss:
\begin{align}
     \mathcal L_{att_{RGB}} & = \gamma_1 (l(A^{S_R},A^{S_F})+l(A^{C_R},A^{C_F})) ,\\
     \mathcal L_{att_{TIR}} & = \gamma_2 (l(A^{S_T},A^{S_F})+l(A^{C_T},A^{C_F})),
\end{align}
where $\gamma_1,\gamma_2$ are hyper-parameters to balance the loss, and $l$ denotes the L1 loss. In this way, the spatial and channel attention $A^{S_F}, A^{C_F}$ of the student network can learn from teacher network and generate fusion attention that is suitable for itself. Then the loss generated by the focal feature extraction is:

\begin{equation}
\begin{aligned}
    \mathcal L_{focal} = &  \mathcal L_{focal_{RGB}} +  \mathcal L_{focal_{TIR}}  \\
    = & \mathcal L_{target_{RGB}} + \mathcal L_{att_{RGB}} + \mathcal L_{att_{TIR}}  + \mathcal L_{target_{TIR}} .
\end{aligned}
\end{equation}

\subsubsection{\textbf{Total MEA Loss}}
The total loss is generated by two MEA modules. The two MEA modules transfer the knowledge of the RGB feature and the TIR feature from the teacher network to the student network. The student network generates the fusion strategy during training optimized by the MEA loss:
\begin{equation}
\begin{aligned}
    \mathcal L_{MEA} & = \mathcal L_{MEA_{RGB}} + \mathcal L_{MEA_{TIR}}  \\
    & = \mathcal L_{global_{RGB}}+ \mathcal L_{focal_{RGB}} +\mathcal L_{global_{TIR}}+\mathcal L_{focal_{TIR}} .
\end{aligned}
\end{equation}

\subsection{Overall Loss}
The total loss of the adaptive modal fusion distillation is:
\begin{equation}
     \mathcal L = \mathcal L_{original} + \mathcal L_{MEA},
\end{equation}
where $\mathcal L_{original}$ is the original detection loss of the detector. Our frame work only need the feature after the FPN of the teacher network. So, the framework can be used in many types of detectors.

\section{SMOD Dataset}
We propose the SJTU Multispectral Object Detection (SMOD) Dataset for detection. 
\subsection{Image Capture and Registration}

\begin{figure}[h]
\centering
\subfloat[different field of views]{\includegraphics[height=1.0in]{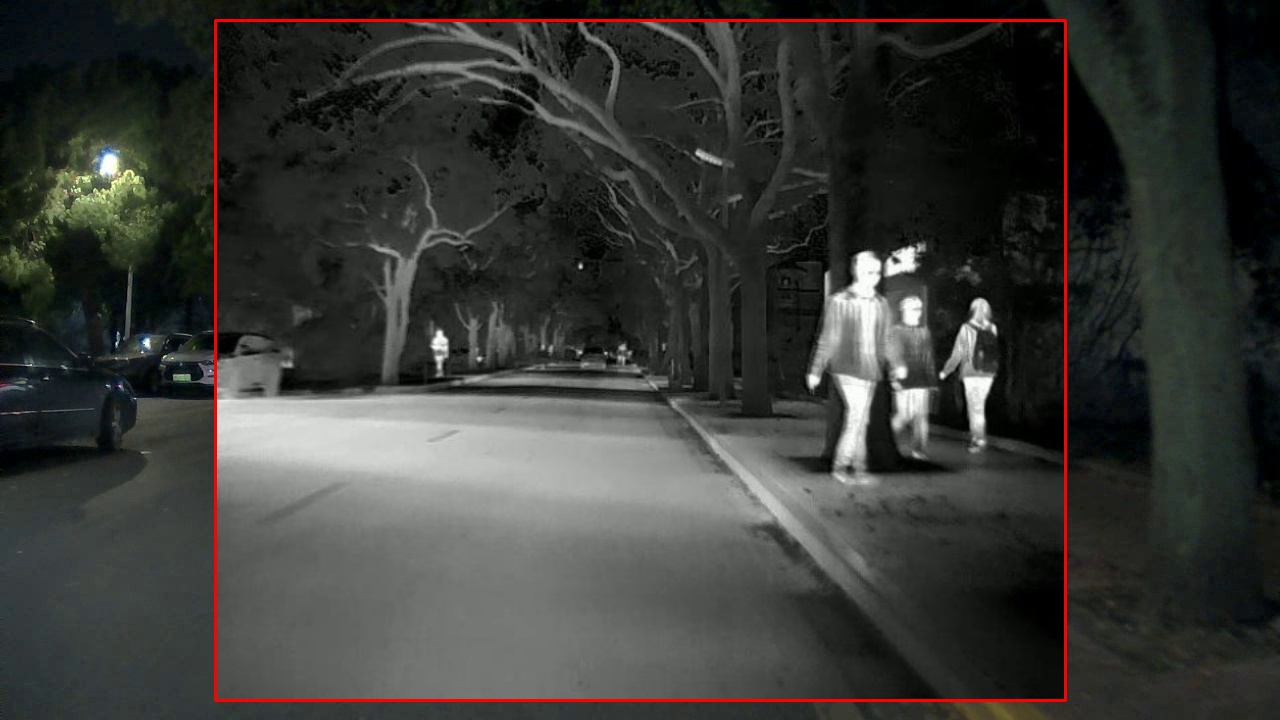}%
\label{fig:different_field}}
\hfil
\subfloat[clipped visible images]{\includegraphics[height=1.0in]{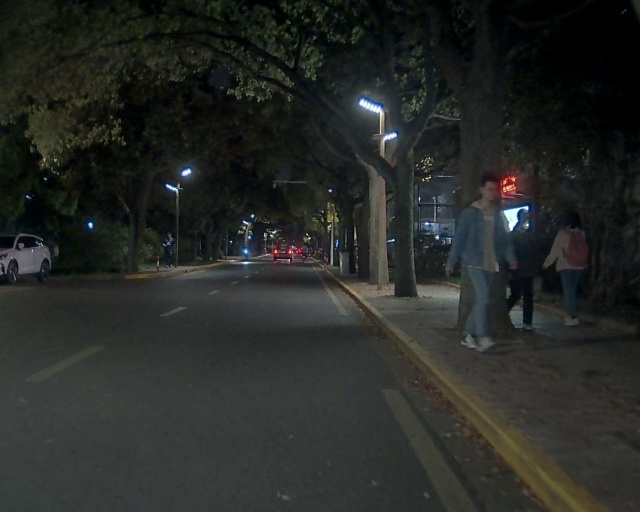}%
\label{fig:rgb_cut}}
\caption{Image capture and image registration. The visible light camera and the infrared camera has the different field of views shown in (a). The image pairs are aligned after clipped and registered the visible images shown in (b).}
\label{fig:SMOD_registration}
\end{figure}

The images are captured by the Asens FV6, a binocular vehicle camera platform that consist of a visible light camera and a infrared camera. Due to the different field of view of different sensor cameras, we clip and register visible images to make image pairs strictly aligned. Fig.\ref{fig:SMOD_registration} shows the different field sizes of different cameras and the visible image after registration. We also provide unregistered image pairs and annotations for researchers to study visible and infrared image registration. The dataset contains 5378 pairs of images of daytime scenes taken at 3pm and 3298 pairs of images of nighttime scenes taken at 7pm.

\subsection{Annotations and Advantages}
We annotate four categories of targets: pedestrians, riders, cars, and bicycles. In particular, we annotate the degree of occlusion for each type of object. As shown in Tab.\ref{tab:object_occlusion}, we categorize occlusion into four categories based on the percentage of occlusion: no occlusion (NO), light occlusion (LO), moderate occlusion (MO) and heavy occlusion (HO).  

\begin{table}[h]
\centering
\caption{Number of each type of object in the SMOD dataset with different degrees of occlusion. The "$O$" denotes the percentage of the object that is occluded.}
\label{tab:object_occlusion}
\begin{tabular}{@{}lcccc@{}}
\toprule
                  & NO & LO & MO & HO\\
         Object  & $O=$0\% &$O<$30\% & 30\%$\le O<$60\% & 60\%$\le O<$90\%\\ \midrule \hline
pedestrian & 4007& 1902& 1285& 848 \\
rider   &  8151 & 1024 & 721 & 582 \\
bicycle & 3039 & 1852 & 1762 & 1279  \\ 
car & 1819 & 1118 & 1595 & 1890 \\
                  \hline \bottomrule
\end{tabular}
\end{table}

Tab.\ref{tab:compare_dataset} shows the comparison of SMOD and the existing datasets mentioned in Section 2. Our SMOD dataset has the following advantages:

$\bullet$ Since Campus roads have an abundance of pedestrians and riders, most of the image pairs contain very dense and abundant objects, especially bicycle and person. There are also complex occlusion relationships between these objects, making the scene in the SMOD dataset even more challenging.

$\bullet$ The SMOD dataset provides fine-grained annotation of the degree of occlusion. We quantitatively categorize the degree of occlusion into four categories based on the percentage of the occluded area of the object. Clear occlusion annotations can effectively evaluate the robustness and safety of the model, which is important for autonomous driving.

\begin{figure}[h]
\centering
\subfloat{\includegraphics[width=0.45\linewidth]{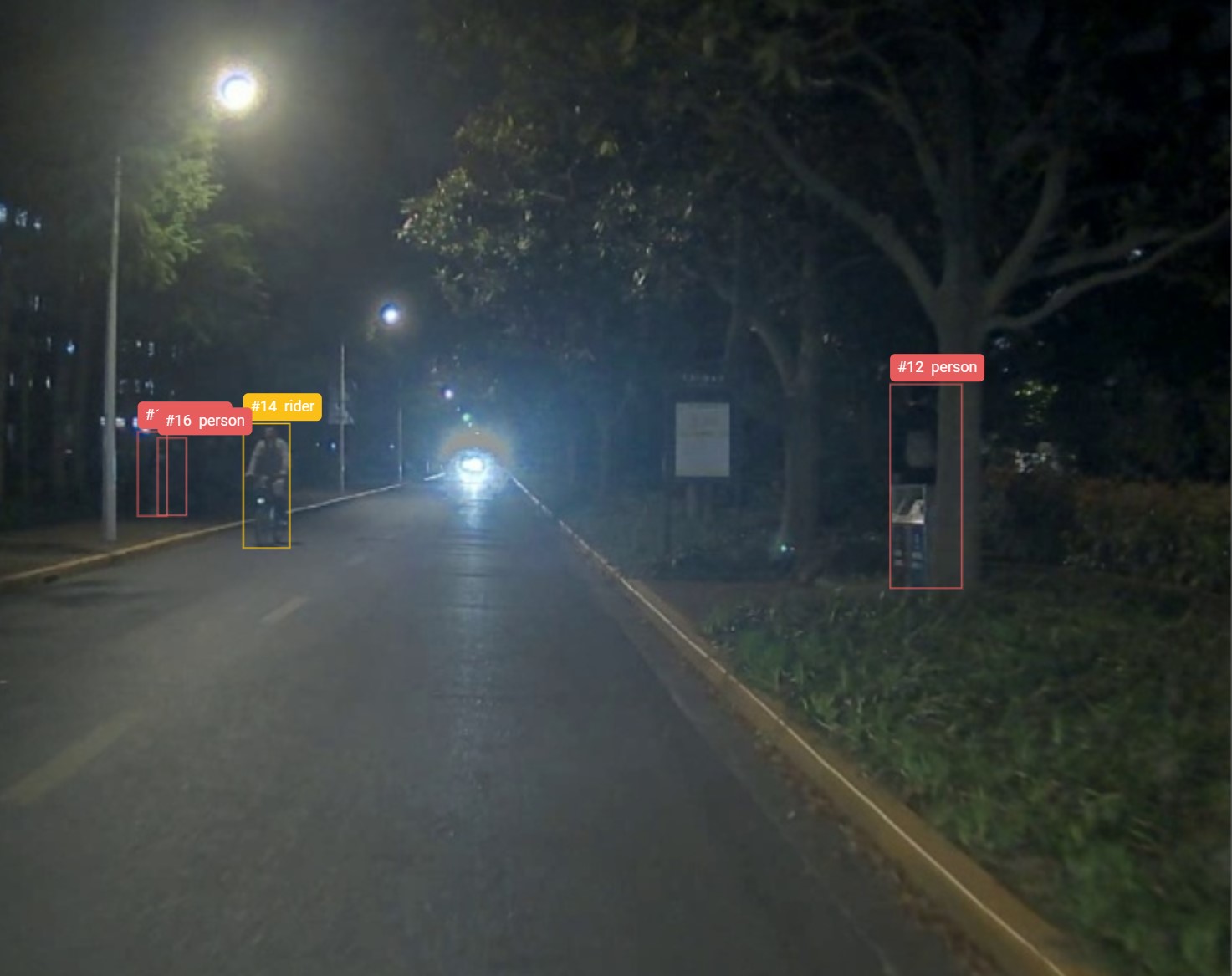}%
\hspace{1mm}}
\subfloat{\includegraphics[width=0.45\linewidth]{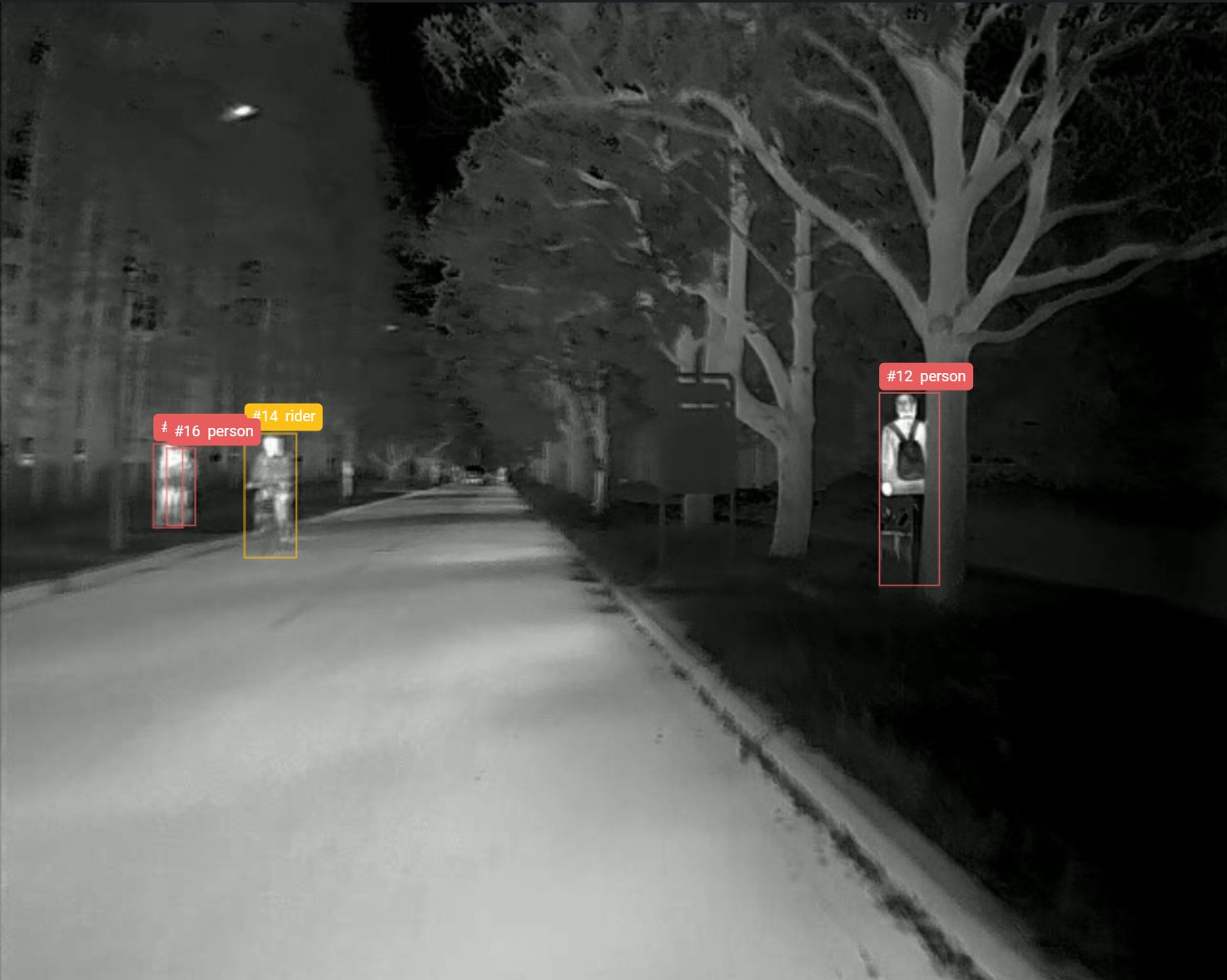}%
}
\vskip -8.2pt
\subfloat{\includegraphics[width=0.45\linewidth]{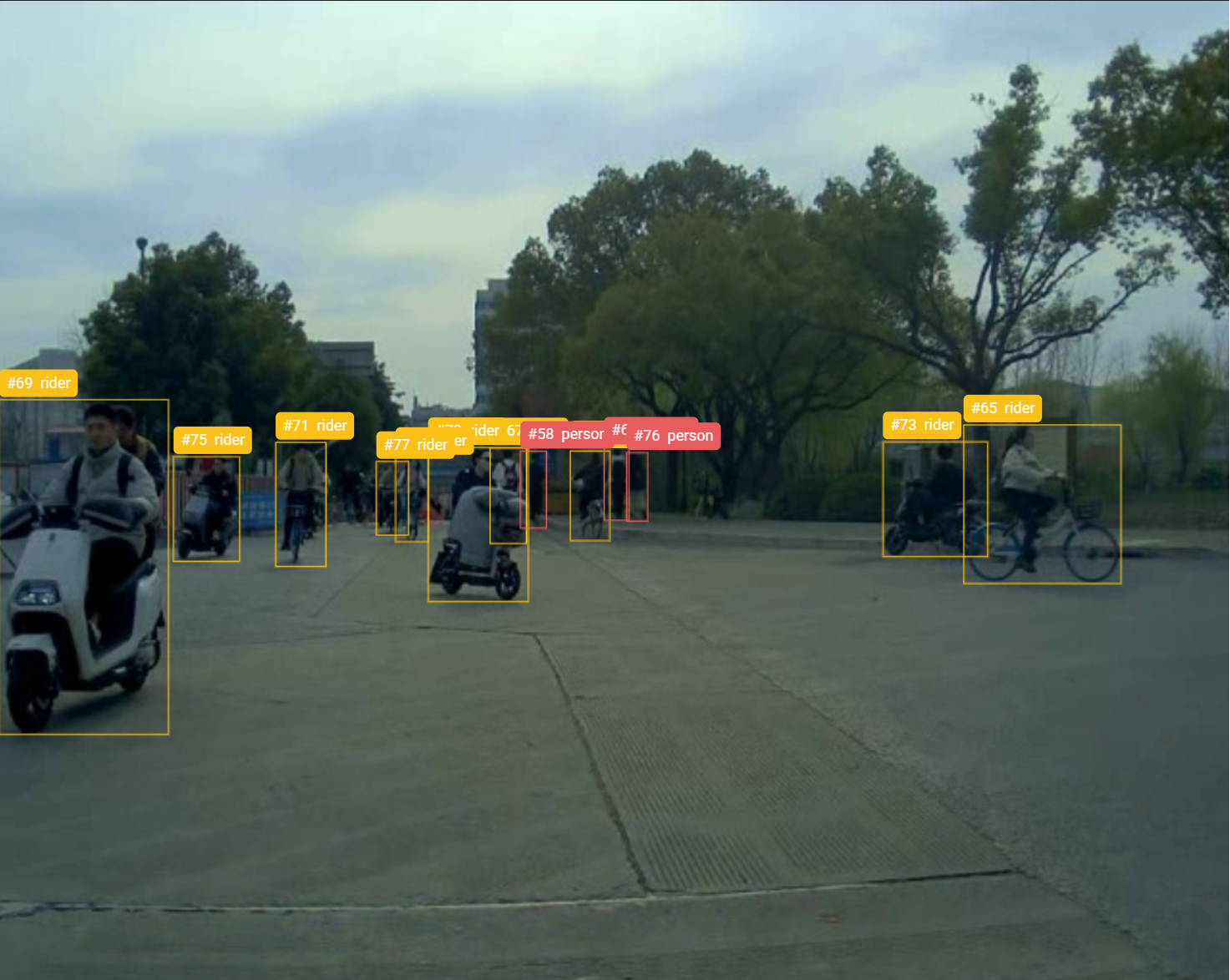}%
\hspace{1mm}}
\subfloat{\includegraphics[width=0.45\linewidth]{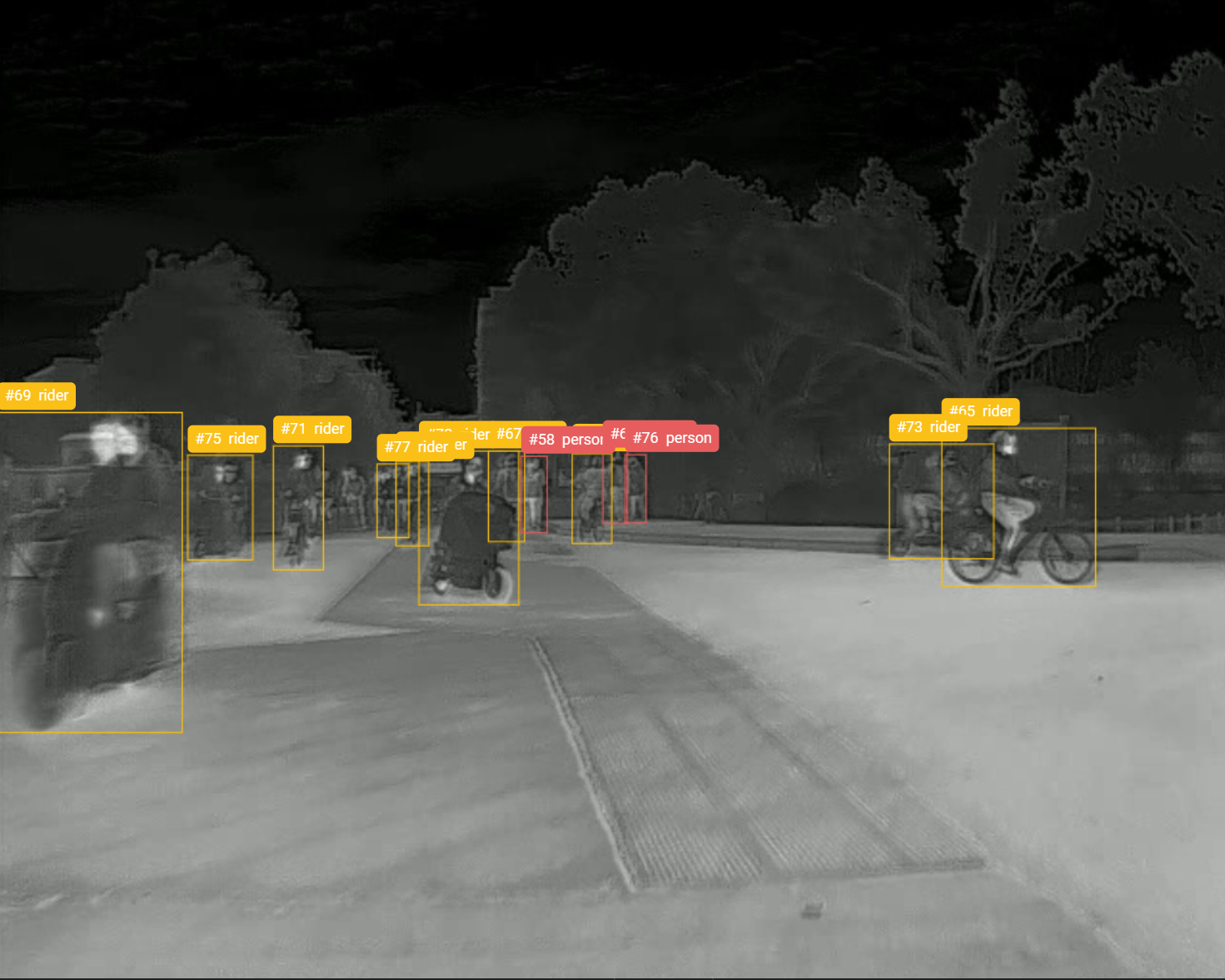}%
}
\caption{Image pairs in the SMOD dataset with strong light and occluded dense objects.}
\label{fig:dense and light}
\end{figure}

$\bullet$ Visible-infrared images are aligned strictly in time and space, and we annotate three other types of objects which are common in driving. Moreover, the image pairs contain rich illumination variations at night, such as the very low illumination and strong light shown in Fig.\ref{fig:dense and light}. Thus, image pairs can be used for other driving scene tasks other than object detection, such as image fusion, supervised image-to-image translation and low-light segmentation.

\section{Experiments}
\subsection{Dataset and Evaluation Metric}
\noindent \textbf{KAIST:} KAIST dataset\cite{hwang2015multispectral} is a widely used benchmark dataset in multispectral pedestrian detection. 
Since the original annotation of the dataset contains noise, we use the improved annotation proposed by \cite{li2018multispectral} and the sanitized annotation proposed by \cite{zhang2019weakly} for training. The test annotation we adopt is improved by \cite{liu2016multispectral}. The log-average Miss Rate ($\textit{MR}^{-2}$) is calculated by averaging the False Positive Per Image (FPPI) sampled within the range of [$10^{-2},10^0$], the lower the better. We adopt the $\textit{MR}^{-2}$ as our evaluation metric under the "reasonable" settings (excluding pedestrians that are occluded or shorter than 55 pixels).

\noindent \textbf{LLVIP:} LLVIP dataset\cite{jia2021llvip} is a challenging dataset for multispectral pedestrian detection including 12,025 training images and 3,463 test images. We use the COCO-style Average Precision\cite{lin2014microsoft} as evaluation metrics.

\noindent \textbf{SMOD:} We introduce this dataset in detail in Section.4. We divide 1876 pairs of images from a total of 8676 pairs as the test set, which contains 1178 pairs of daytime images and 698 pairs of nighttime images. We use both the COCO-style Average Precision\cite{lin2014microsoft} as evaluation metrics and log-average Miss Rate ($\textit{MR}^{-2}$) mentioned before.

\noindent\textbf{SUNRGB-D:} The SUNRGB-D\cite{song2015sun} dataset is a comprehensive dataset for 3D scene understanding, but it also provides valuable resources for 2D object detection. The dataset contains 10,335 pairs of well-aligned visible depth image pairs and provides bounding boxes for 2D object detection in the annotation file. Although the detection object is various types of furniture, we use it for simple generalizability verification.

\begin{figure}[h]
  \centering
  \includegraphics[width=\linewidth]{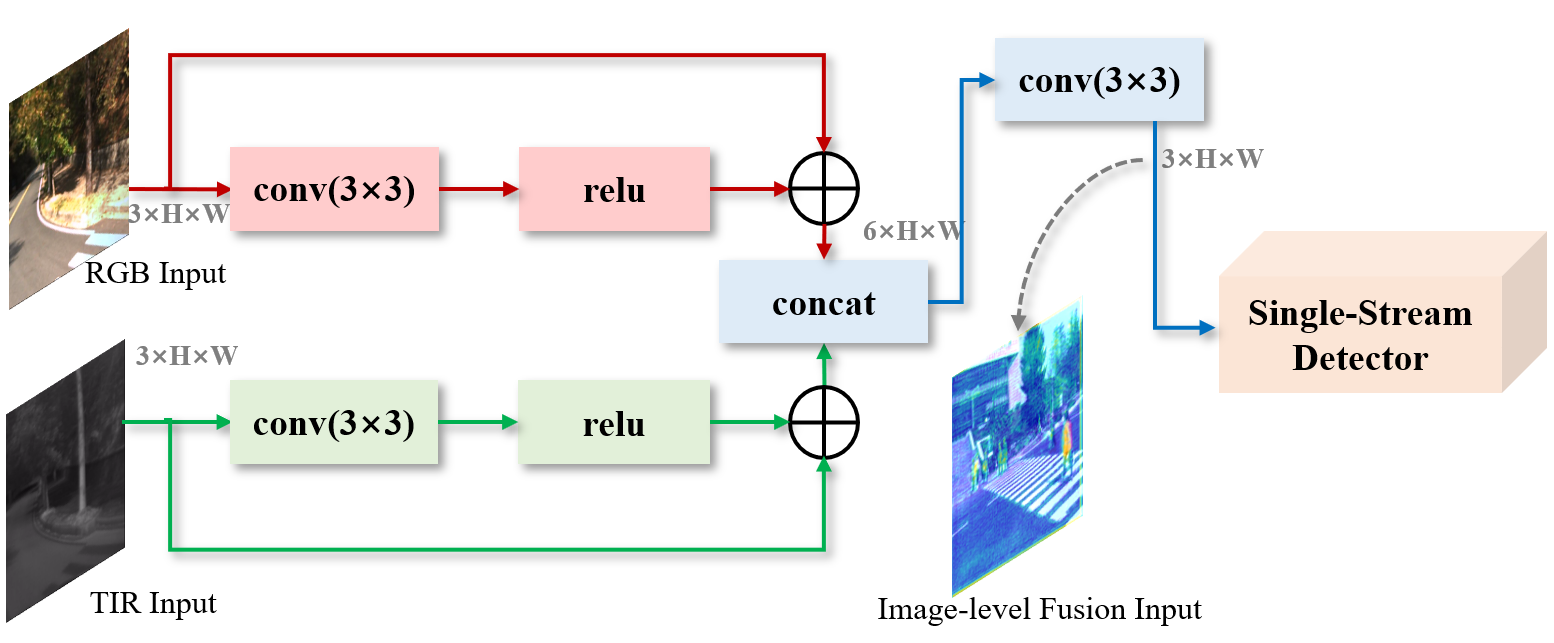}
  \caption{The structure of image-level fusion. It contains two residual modules and a concatenation module to get simple fusion results as the input.}
  \label{fig:IMF}
\end{figure}

\subsection{Implementation Details}
\noindent \textbf{Network Architecture.} We use MMDetection\cite{chen2019mmdetection}, a popular object detection toolkit based on PyTorch\cite{paszke2017automatic}, to implement our approach. 
We choose the DETR-based DINO\cite{zhangdino} detector, the classical two-stage network Faster-RCNN\cite{2017Faster} and one-stage detector RetinaNet\cite{lin2017focal} as baseline detectors. We use ResNet\cite{2016Deep} and Swin Transformer\cite{liu2021swin} as the backbone of the networks. The two-stream teacher network fuse the feature with the cross-modality attention transformer\cite{lee2022cross}. The fusion of one stream of students network is early image-level fusion by the fusion module shown in Fig.\ref{fig:IMF}. Note that when we do not specify the backbone of the network, we default to using ResNet50 for dual-stream teacher networks and ResNet18 for single-stream student networks.

\noindent \textbf{Training Details} All networks are trained on a single Nvidia GeForce GTX 1080Ti GPU. The backbone networks for all experiments are pre-trained on the ImageNet dataset\cite{russakovsky2015imagenet}. For all datasets, the learning rate is set to be $1\times10^{-4}$ with the $1\times10^{-4}$ weight decay AdamW optimizer. We adopt the hyper-parameters $\{\alpha_1 = \alpha_2 = 5\times10^{-5}, \gamma_1= \gamma2 = 5\times10^{-5}, \lambda_1 = \lambda_2 = 5\times10^{-7} \}$ for Faster-RCNN and DINO and $\{\alpha_1 = \alpha_2 = 1\times10^{-3}, \gamma_1= \gamma2 = 1\times10^{-3}, \lambda_1 = \lambda_2 = 5\times10^{-6} \}$ for RetinaNet.

\subsection{Distillation on KAIST Dataset}

\begin{table}[h]
\centering
\caption{The distillation results of AMFD on KAIST trained by the imporved annotations\cite{zhang2019weakly}.}
  \label{tab:distill_res}
\begin{tabular*}{0.475 \textwidth}{@{\extracolsep{\fill}}lllll@{}}
\toprule
                 Detectors  & All & Day & Night & Time\\ \midrule \hline
                 Faster-RCNN(Tea.) &   7.66 & 9.34 & 5.14  & 0.12s \\ \hline
                
\multicolumn{3}{l}{\emph{trained with the improved annotation}} & &  \\
               \makecell[l]{Faster-RCNN(Stu.)}   & 11.86 & 14.46 & 7.02 & \multirow{2}{*}{0.05s} \\
               \textbf{AMFD}+\textbf{Faster-RCNN}(Stu.) &  \textbf{7.23} & \textbf{9.83} & \textbf{2.18} &    \\ \hdashline
\makecell[l]{RetinaNet(Stu.)} & 13.80 & 15.43 & 9.99 & \multirow{2}{*}{\textbf{0.04s}} \\
               \textbf{AMFD}+\textbf{RetinaNet}(Stu.)   & 8.82 & 10.98 & 4.86 &  \\ \hline 
\multicolumn{3}{l}{\emph{trained with the sanitized annotation}} &  &\\
\makecell[l]{Faster-RCNN(Stu.)} & 14.54 & 18.99 & 6.90 & \multirow{2}{*}{0.05s} \\
              \textbf{AMFD}+\textbf{Faster-RCNN}(Stu.)   & \textbf{6.65} & \textbf{9.07} & \textbf{2.11} &    \\ \hdashline
\makecell[l]{RetinaNet (Stu.)} & 12.99 & 15.67 & 7.50 & \multirow{2}{*}{\textbf{0.04s}} \\
                \textbf{AMFD}+\textbf{RetinaNet}(Stu.)   & 7.66 & 10.30 & 3.16 &  \\ \hline \bottomrule
\end{tabular*}
\end{table}

\noindent \textbf{Analysis of distillation results.}
We fix the teacher network to the two-stream Faster-RCNN network described above and trained it using the improved annotation in all distillation experiments. According to the experimental results in Tab.\ref{tab:distill_res}, AMFD can reduce $\textit{MR}^{-2}$ of the student network on the All set by a factor of nearly two. When trained with the improved annotations, the inference time of the two-stage student network is reduced by 58.3\%, with even 0.4\% decrease in $\textit{MR}^{-2}$; meanwhile, the inference time of the one-stage student network is reduced by 66.7\% with only 2\% higher $\textit{MR}^{-2}$. In particular, we find that the student network distilled by AMFD has a low $\textit{MR}^{-2}$ on the \textbf{Night} set, i.e., the network distilled by AMFD has excellent extraction of TIR features. Student networks trained with the sanitized annotation even have better performance than those trained with the improved annotation.

\begin{table}[h]
\centering
\caption{Ablation study of the proposed fusion distillation architecture. The "RGB", "TIR" and "Fusion" denote whether distill the RGB, TIR and Fusion Feature respectively.}
\label{tab:ablation_multi}
\begin{tabular}{@{}l|ccc|lll@{}}
\toprule
              Detector    & RGB & TIR & Fusion & All & Day & Night \\ \midrule \hline
\multirow{3}{*}{Faster-RCNN} &  &  &  &11.86 & 14.46 & 7.02 \\
                  &  &  & \Checkmark & 11.03 & 14.07 & 5.64 \\
                  & \Checkmark & \Checkmark &  & \textbf{7.23} & \textbf{9.83} & \textbf{2.18 } \\ \hline
\multirow{3}{*}{RetinaNet} &  &  &  &13.80 & 15.43 & 9.99 \\
                  &  &  & \Checkmark  & 10.93 & 13.65 & 5.76 \\
                  & \Checkmark  & \Checkmark  &  & \textbf{8.82} & \textbf{10.98} & \textbf{4.86} \\ \hline \bottomrule
\end{tabular}
\end{table}

\begin{table}[h]
\centering
\caption{Ablation study of the proposed MEA. The "Global" and "Focal" denote the global feature extraction and focal feature extraction in MEA.}
\label{tab:ablation_mea}
\begin{tabular*}{0.475 \textwidth}{@{\extracolsep{\fill}}l|ll|lll@{}}
\toprule
                  Detector    & Global & Focal  & All & Day & Night \\ \midrule \hline
\multirow{4}{*}{Faster-RCNN} &  &   &11.86 & 14.46 & 7.02\\
                  &  & \Checkmark  &9.85  &13.35  &3.24 \\
                  & \Checkmark &  &7.51  &10.12  &3.25  \\
                  & \Checkmark &  \Checkmark & \textbf{7.23} & \textbf{9.83} & \textbf{2.18} \\ \hline
\multirow{4}{*}{RetinaNet} &  &  &  13.80 & 15.43 & 9.99\\
                  &  & \Checkmark & 11.34 &  13.97  & 6.37 \\
                  & \Checkmark & & 9.62 & 11.87 &  5.28 \\
                  & \Checkmark & \Checkmark &  \textbf{8.82} & \textbf{10.98} & \textbf{4.86} \\ \hline
\bottomrule
\end{tabular*}
\end{table}

\noindent \textbf{Ablation experiments and analysis.} 
Ablation experiments are conducted on the KAIST dataset to demonstrate the effectiveness of the proposed fusion distillation architecture and MEA module. In Tab.\ref{tab:ablation_multi}, distilling both the RGB and the TIR features denotes the fusion architecture, and only distilling the fusion features denotes the traditional architecture. For a fair comparison, the distillation modules we use in traditional architecture is also MEA module. The results of the experiment in Tab.\ref{tab:ablation_multi} show that the fusion architecture can greatly improve the performance of the student network. In Tab.\ref{tab:ablation_mea}, the "Global" and "Focal" denote global feature extraction and focal feature extraction in MEA. We can find that combining both local and global features is the more effective distillation method. Results of the ablation study demonstrate the effectiveness of the proposed distillation framework. 

\begin{table}[h]
\centering
\caption{Comparison of the teacher and student network with other multispectral distillation methods in terms of the log-average miss-rate on the KAIST test set}
\label{tab:compare_distill}
\begin{tabular}{@{}lccccc@{}}
\toprule
 Detector& Distill& \multicolumn{2}{c}{MR(All)}& \multicolumn{2}{c}{Time} \\ \midrule \hline
 \makecell[c]{DCRL\cite{liu2021deep} (Tea.)}& \textbf{--} & 9.16& \multirow{2}{*}{\textbf{---}} &  0.175s & \multirow{2}{*}{$\downarrow$ 26.8\%}\\
 \makecell[c]{DCRD\cite{liu2021deep} (Stu.)}&  \Checkmark& 12.58&  &  0.128s & \\ \hline
 MD\cite{zhang2022low} (Tea.)& \textbf{--} & 7.77&  & 11ms & \multirow{2}{*}{$\downarrow$ 36.3\%} \\
 MD\cite{zhang2022low} (Stu.)&   & 9.40& \multirow{2}{*}{$\downarrow$ 1.62\%} &  7ms &\\
 MD\cite{zhang2022low} (Stu.)&  \Checkmark  & 7.78 &  & 7ms  &\\ \hline
 \textbf{Ours}(Tea.)&  \textbf{--} & 7.66 & & 0.12s & \multirow{2}{*}{\textbf{$\downarrow$ 58.3\%}} \\
 \textbf{Ours}(Stu.)&   & 11.86 & \multirow{2}{*}{\textbf{$\downarrow$ 4.63\%}} & 0.05s & \\ 
  \textbf{Ours}(Stu.)&  \Checkmark & \textbf{7.23} & & \textbf{0.05s} &\\ 
 \hline \bottomrule
\end{tabular}
\end{table}

\begin{table*}[h]
\centering
\caption{Comparison of the student network distilled by the proposed AMFD with the state-of-the-art methods in terms of the log-average miss-rate on the KAIST test set. The "$^\star$" indicates the method is based on the one-stage detector and the rest are on the two-stage detector. Bolding indicates the lowest and underlining indicates the second lowest.}
\label{tab:compare}
\begin{tabular*}{0.80 \textwidth}{@{\extracolsep{\fill}}l|l|ll|ccc|l@{}}
\toprule
Methods & Publication Year & Backbone & GPU & \textbf{All} & Day & Night & Time \\ \midrule \hline
  CMM\cite{kim2024causal} & CVPR 2024 & ResNet50 & \textbf{--} & 8.54 & 9.60 & 5.93 & \textbf{--} \\
  MFPT-RetinaNet\cite{zhu2023multi}$^\star$ & TITS 2023 & ResNet50 & 1080Ti & 8.42 & 8.74 & 6.63 & 0.05 \\
  CMPD\cite{li2023multiscale} & TMM 2022 & ResNet50 & 1080Ti & 8.16 & 8.77 & 7.31 & 0.11 \\
 MBNet\cite{zhou2020improving}$^\star$ & ECCV2020 & ResNet50 & 1080Ti & 8.13 & 8.28 & 7.86 & 0.07 \\
  BAANet\cite{yang2022baanet} & ICRA 2022 & ResNet50 & 1080Ti & 7.92 & 8.37 & 6.98 & 0.07 \\
   UFF+UCG\cite{kim2021uncertainty} & TCSVT 2022 & ResNet50 & 1080Ti & 7.89 & 8.18 & 6.96 & 0.09 \\
   MFPT-Faster RCNN\cite{zhu2023multi}   & TITS 2023 & ResNet50 & 1080Ti & 7.72 & 8.26 & \underline{4.53} & 0.08 \\
   RITA\cite{liu2024region} & TIV 2024 & ResNet50 & \textbf{--} & 7.64 & \underline{7.73} & 7.11 & \textbf{--} \\ 
   AANet-RetinaNet\cite{chen2023attentive}$^\star$ & ACM MM 2023 & ResNet50 & 1080Ti & 7.51 & 7.74 & 7.39 & 0.06 \\
  SMPD\cite{li2023stabilizing} & TCSVT 2023 & VGG16 & 1080Ti & 7.44 & 8.21 & 6.23& 0.09 \\
 AANet-Faster RCNN\cite{chen2023attentive} & ACM MM 2023 & ResNet50 & 1080Ti & \textbf{6.91} & \textbf{6.66} & 7.31 & 0.10 \\ \hline
   \textbf{AMFD-Faster RCNN(ours)}  & \textbf{--} & ResNet18 & 1080Ti & \underline{7.23} & 9.83 & \textbf{2.18} &  \underline{0.05}  \\
   \textbf{AMFD-RetinaNet$^\star$(ours)}& \textbf{--} &  ResNet18  & 1080Ti & 8.82 & 10.98 & 4.86 & \textbf{0.04} \\ \hline
   \bottomrule
\end{tabular*}
\end{table*}

\noindent \textbf{Comparison with other distillation methods.} 
To further demonstrate the effectiveness of the AMFD, we also compare other distillation methods in the field of multispectral pedestrian detection. The results are shown in Tab.\ref{tab:compare_distill}. The MD\cite{zhang2022low} decreases the $\textit{MR}^{-2}$ of the student network by 1.62\% after distillation, which indicates that the student network is good relative to the Knowledge Transfer Module proposed by the author. The MD method reduces the inference time (without post-processing treatments) by 36.3\%. The DCRD\cite{liu2021deep} remains 3.42\% higher than its DCRL teacher network in terms of $\textit{MR}^{-2}$ has a 26. 8\% reduction in the inference time. In contrast, AMFD can significantly improve the performance of a simple student network, even a little better than that of teacher networks.

\begin{figure*}[ht]
  \centering
  \captionsetup[subfloat]{labelsep=none,format=plain,labelformat=empty}
        \subfloat{
        \includegraphics[width=0.247\linewidth]{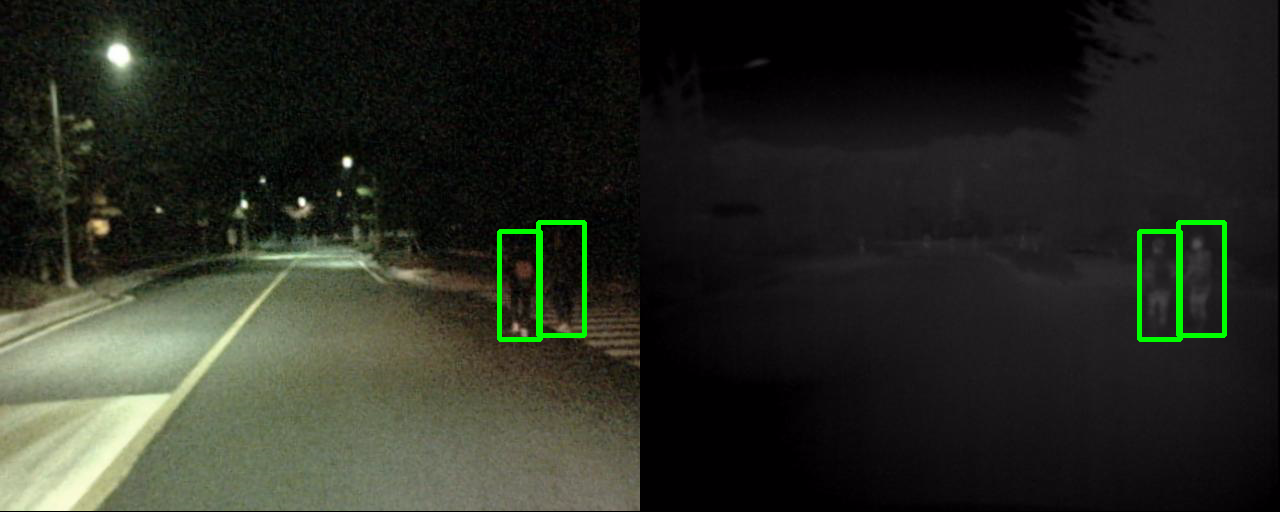}
        \hspace{-3.1mm}
        \label{fig: day1_GT}}
	\subfloat{
        \includegraphics[width=0.247\linewidth]{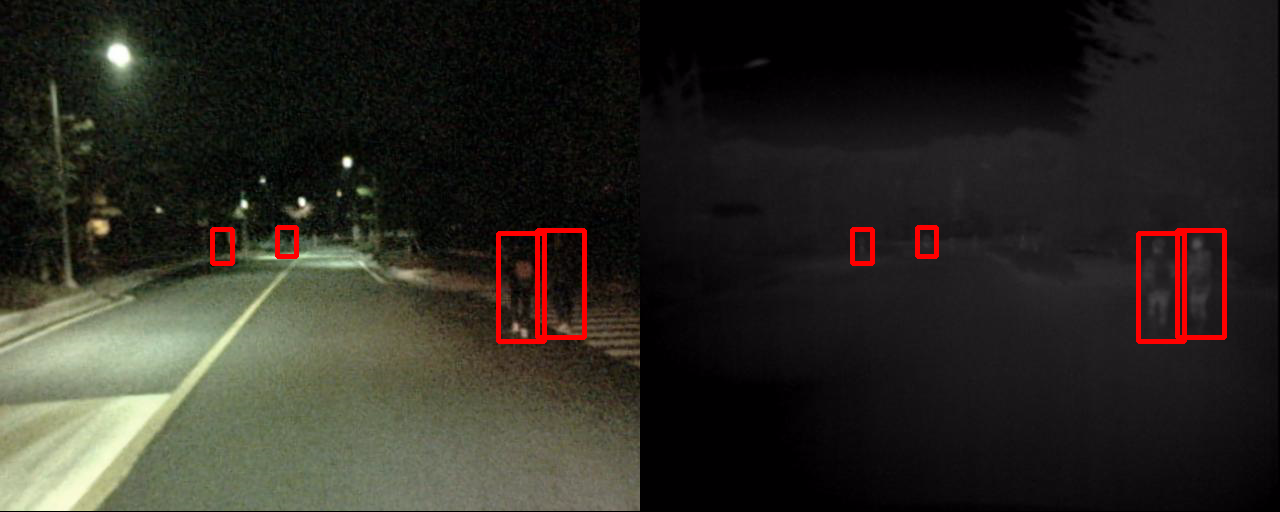}
        \hspace{-3.1mm}
        \label{fig: day1_MBNet}}
        \subfloat{
        \includegraphics[width=0.247\linewidth]{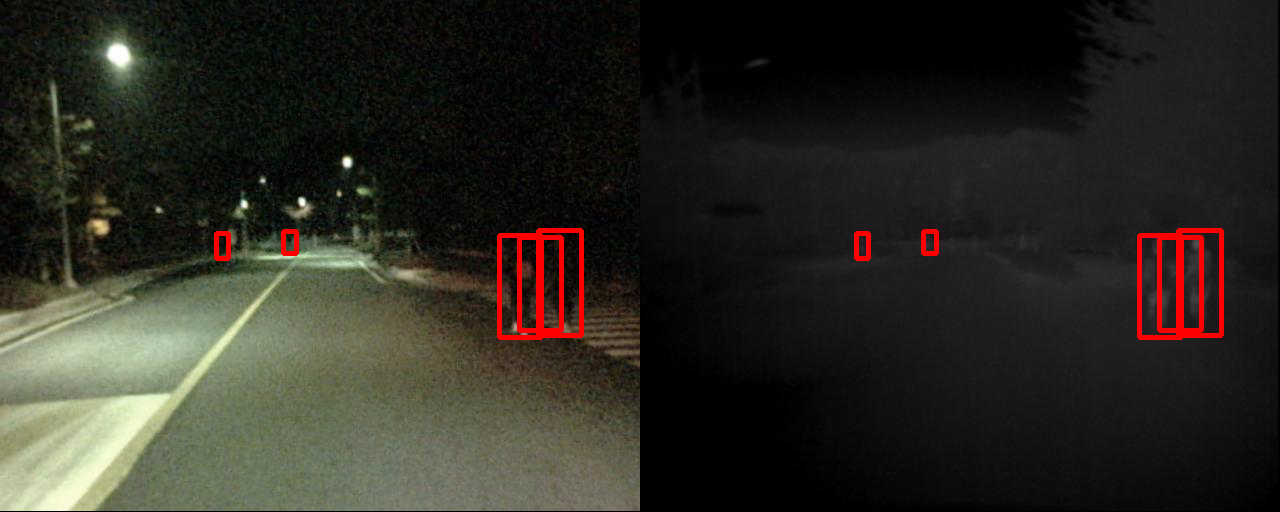}
        \hspace{-3.1mm}
        \label{fig: day1_T}}
        \subfloat{
        \includegraphics[width=0.247\linewidth]{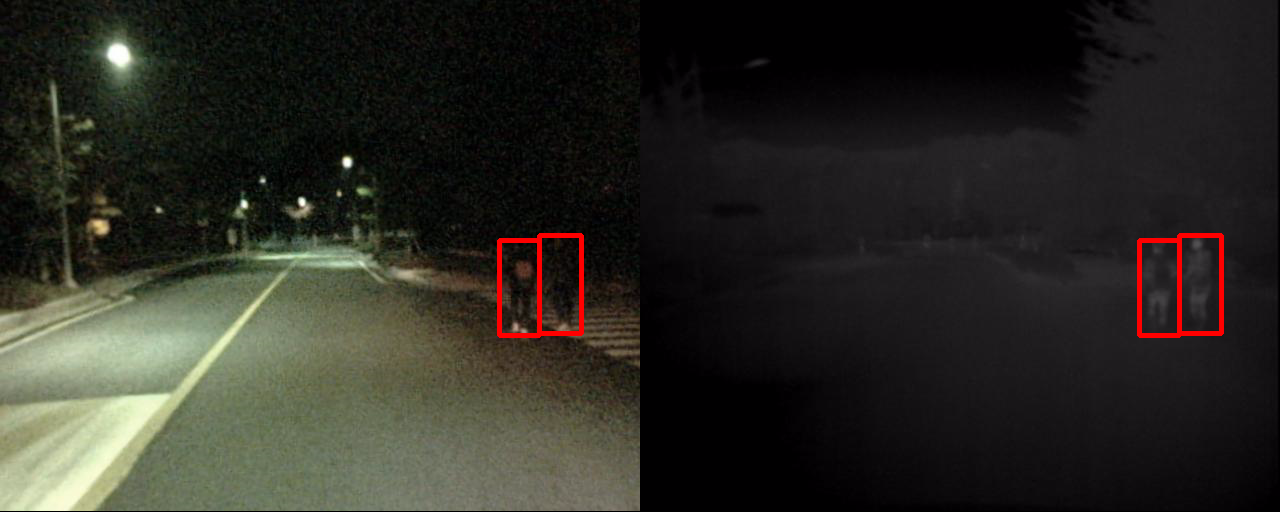}
        \hspace{-3.1mm}
        \label{fig: day1_SD}}
        \vskip -9.2pt
        \subfloat{
        \includegraphics[width=0.247\linewidth]{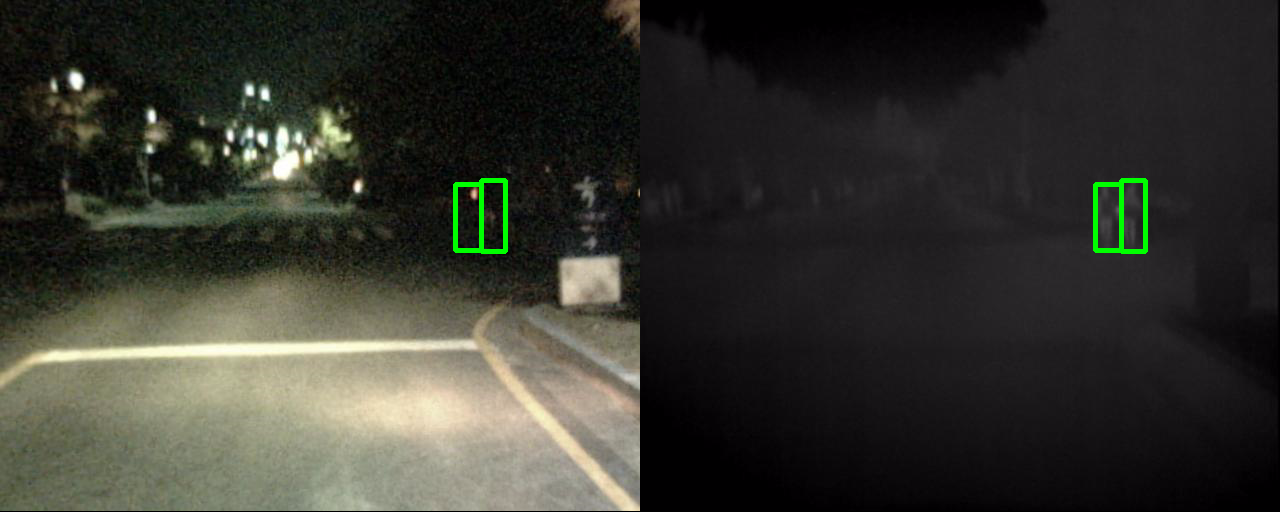}
        \hspace{-3.1mm}
        \label{fig: day2_GT}}
	\subfloat{
        \includegraphics[width=0.247\linewidth]{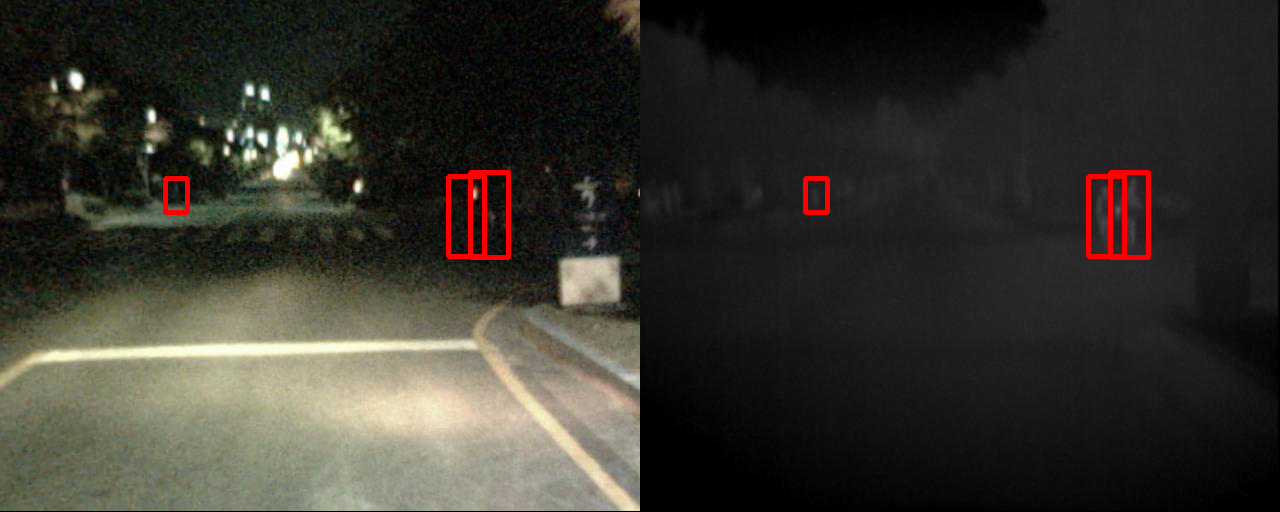}
        \hspace{-3.1mm}
        \label{fig: day2_MBNet}}
        \subfloat{
        \includegraphics[width=0.247\linewidth]{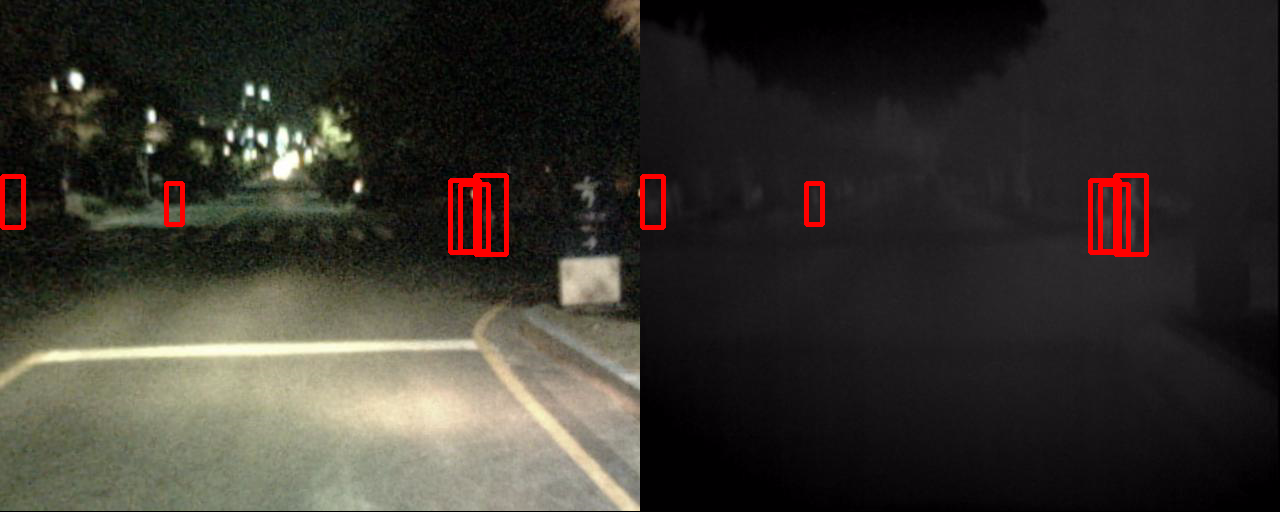}
        \hspace{-3.1mm}
        \label{fig: day2_T}}
        \subfloat{
        \includegraphics[width=0.247\linewidth]{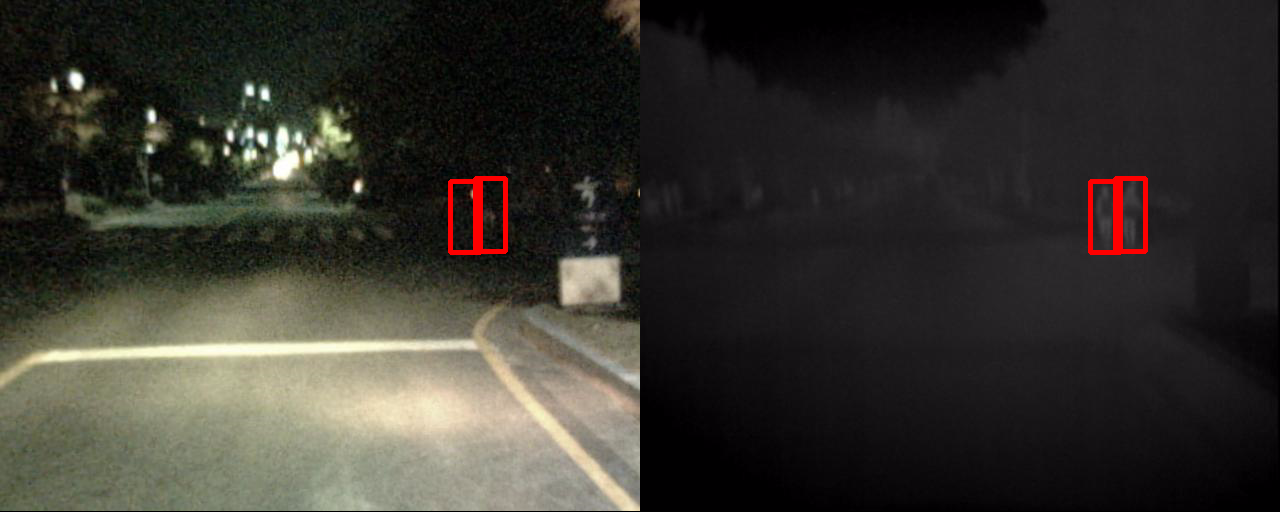}
        \hspace{-3.1mm}
        \label{fig: day2_SD}}
        \vskip -9.2pt
        \subfloat{
        \includegraphics[width=0.247\linewidth]{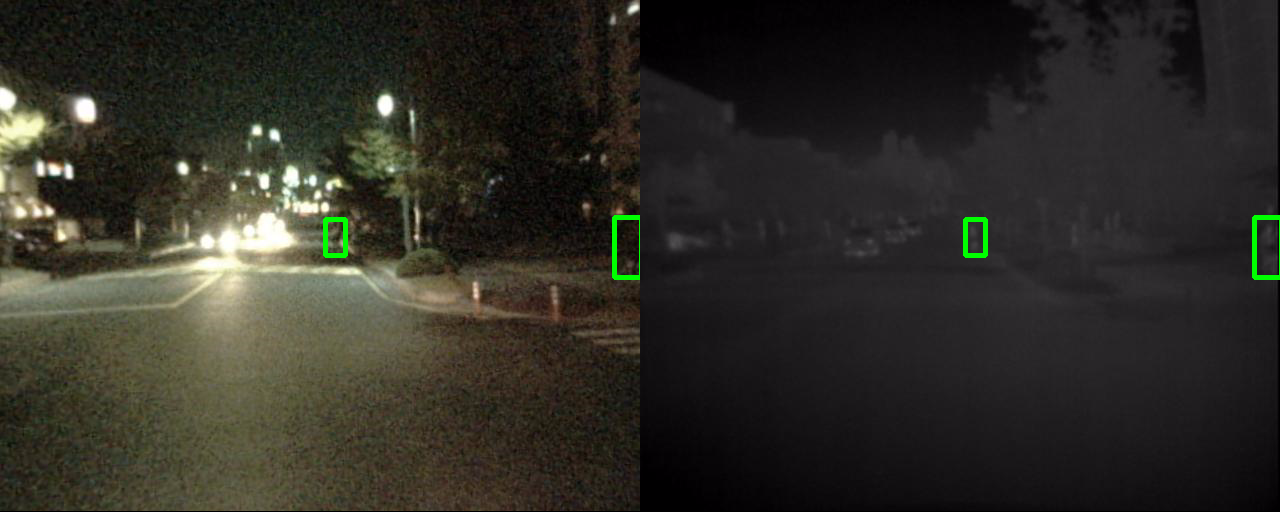}
        \hspace{-3.1mm}
        \label{fig: night1_GT}}
	\subfloat{
        \includegraphics[width=0.247\linewidth]{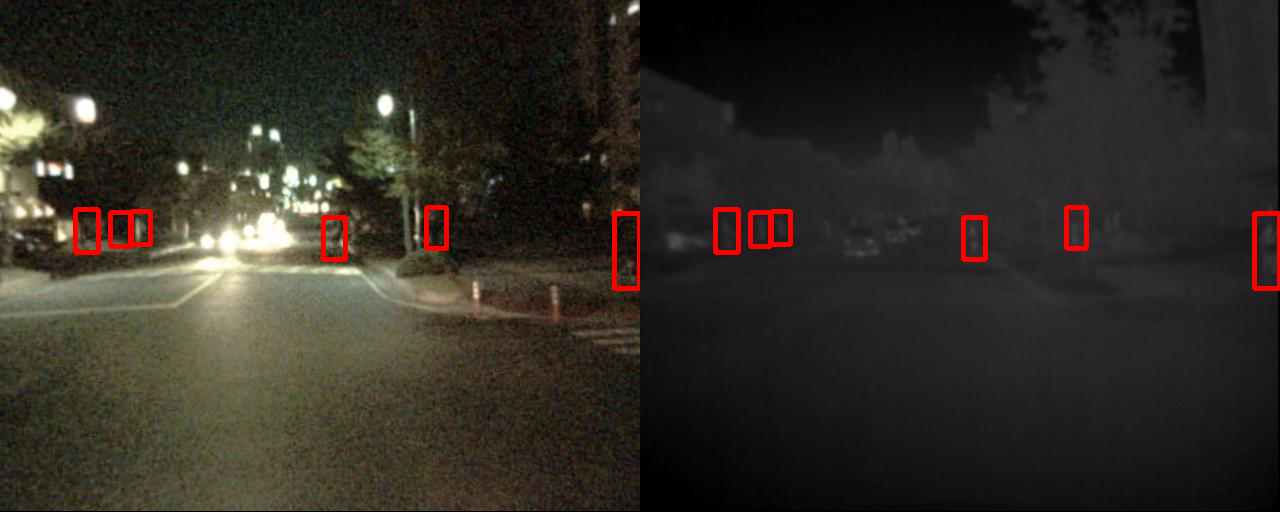}
        \hspace{-3.1mm}
        \label{fig: night1_MBNet}}
        \subfloat{
        \includegraphics[width=0.247\linewidth]{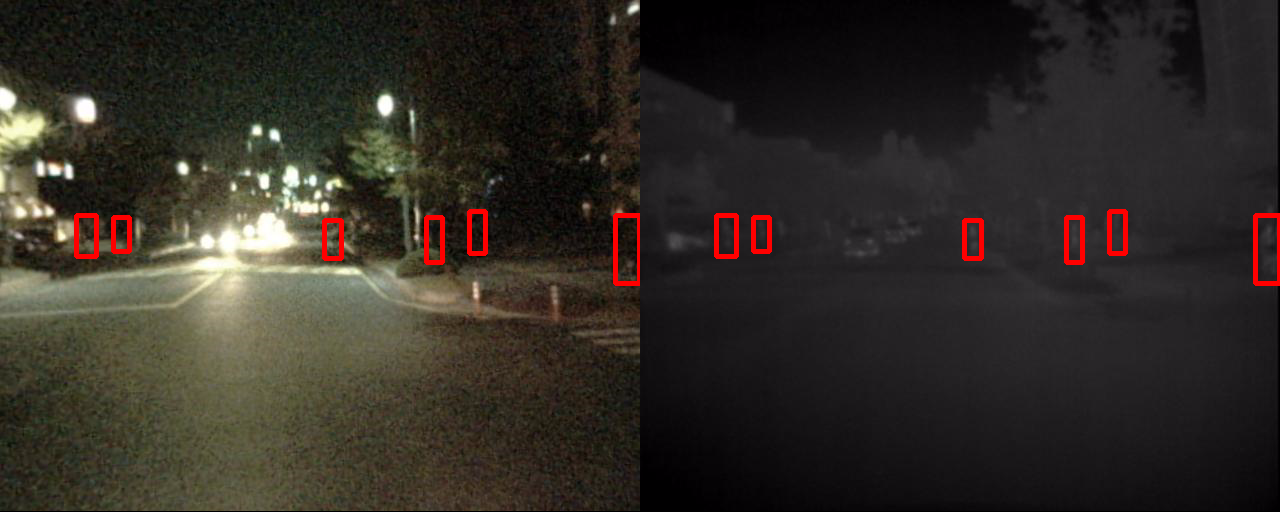}
        \hspace{-3.1mm}
        \label{fig: night1_T}}
        \subfloat{
        \includegraphics[width=0.247\linewidth]{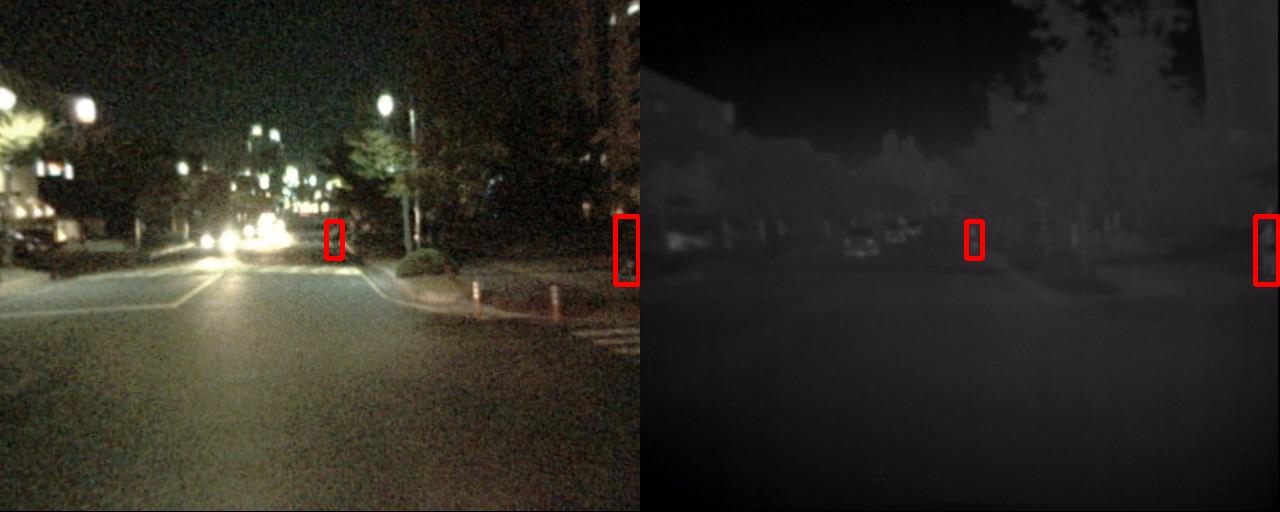}
        \hspace{-3.1mm}
        \label{fig: night1_SD}}
        \vskip -9.2pt
        \subfloat[(a) Ground Truth]{
        \includegraphics[width=0.247\linewidth]{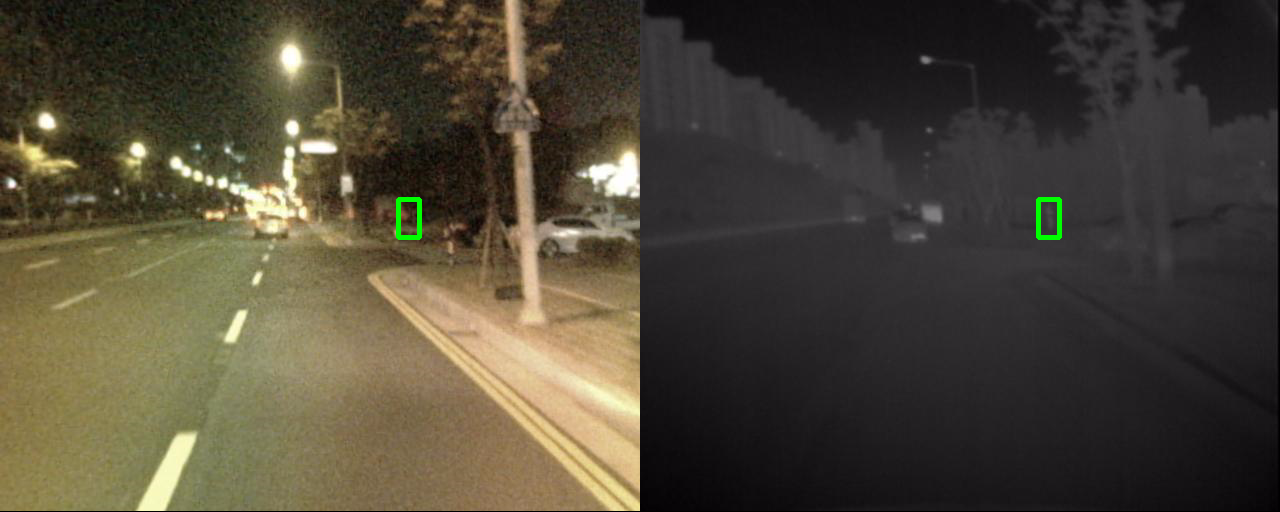}
        \hspace{-3.1mm}
        \label{fig: night1_GT}}
	\subfloat[(b) MBNet]{
        \includegraphics[width=0.247\linewidth]{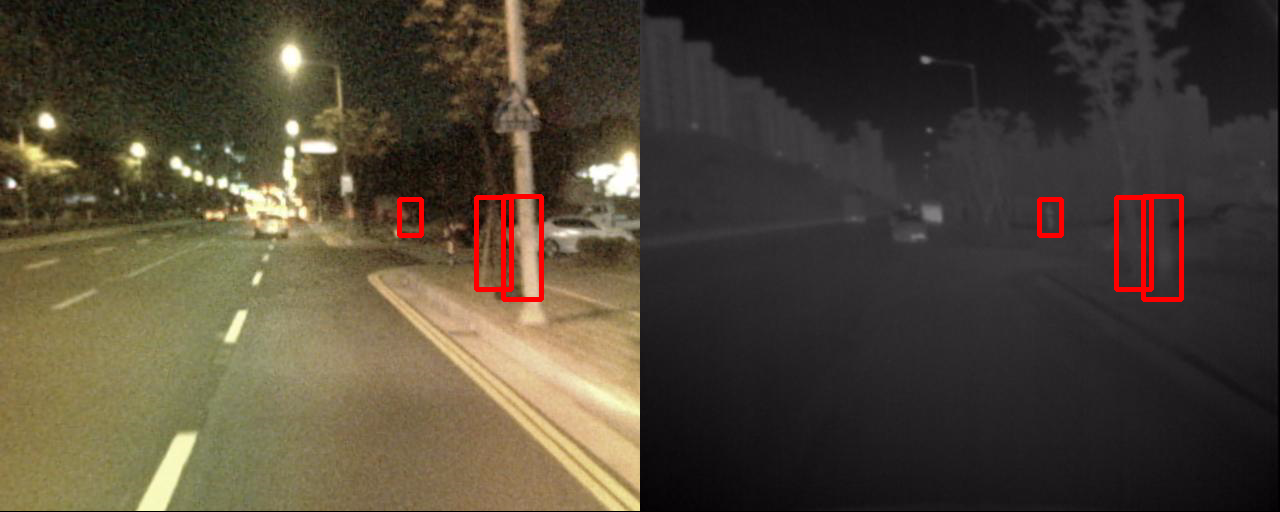}
        \hspace{-3.1mm}
        \label{fig: night1_MBNet}}
        \subfloat[(c) ProbEn3]{
        \includegraphics[width=0.247\linewidth]{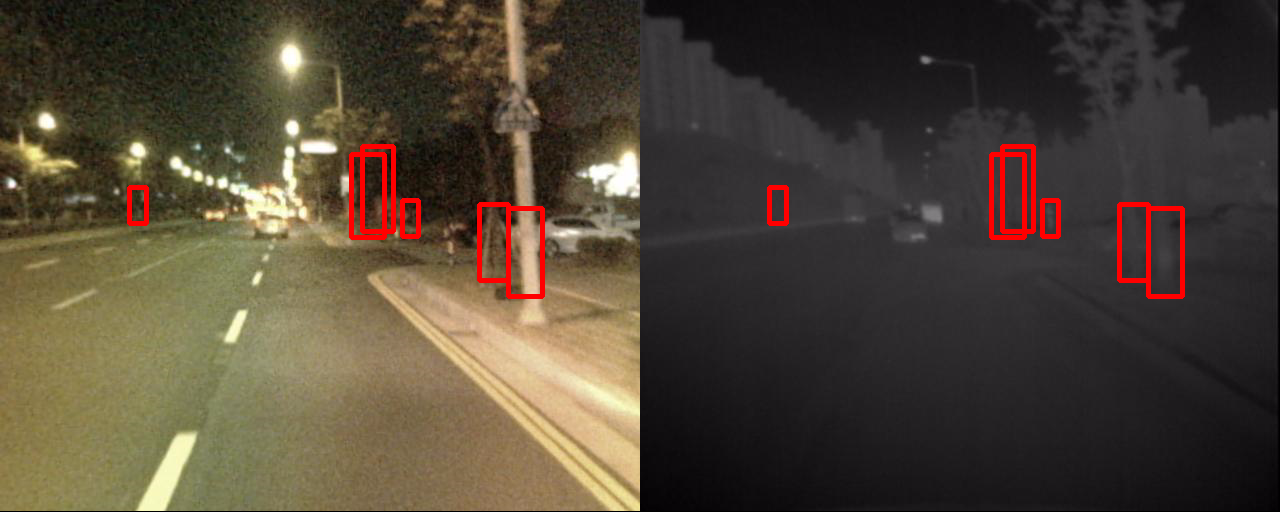}
        \hspace{-3.1mm}
        \label{fig: night1_T}}
        \subfloat[(d) AMFD-Faster RCNN(ours)]{
        \includegraphics[width=0.247\linewidth]{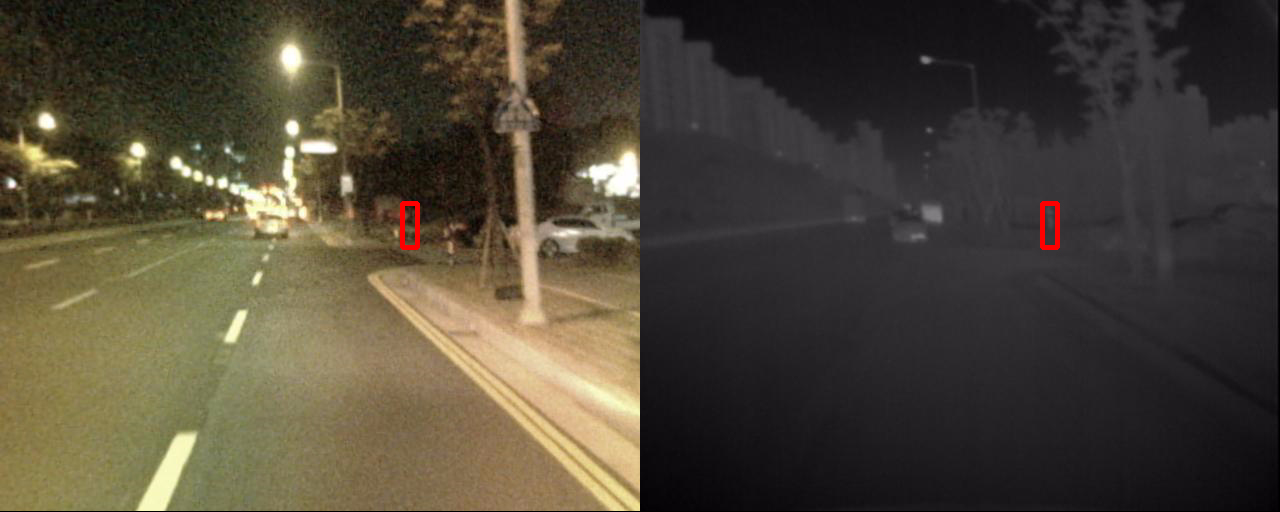}
        \hspace{-3.1mm}
        \label{fig: night1_SD}}
        \vspace{-1mm}
	\caption{Qualitative comparison of our (d) AMFD-Faster RCNN with (b) MBNet\cite{zhou2020improving} and (c) ProbEn3\cite{chen2022multimodal}. Green and red boxes indicate the ground-truth bounding box and detector predictions, respectively. From the figures, we can find that our distilled model is more robust and performs better in the night road scene.}
	\label{qualitative_vis_kaist}
\end{figure*}

\noindent \textbf{Comparison with the state-of-the-art.} 
To evaluate the superiority of our proposed method, we compare our method with some existing state-of-the-art multispectral pedestrian dectection methods on KAIST dataset trained by the improved annotation\cite{zhang2019weakly} namely CMM\cite{kim2024causal}, MFPT\cite{zhu2023multi}, CMPD\cite{li2023multiscale}, MBNet\cite{zhou2020improving}, BAANet\cite{yang2022baanet}, UFF+UCG\cite{kim2021uncertainty}, RITA\cite{liu2024region}, SMPD\cite{li2023stabilizing}, AANet\cite{chen2023attentive}. We compare the $\textit{MR}^{-2}$ under three subsets: \textbf{All}, \textbf{Day} and \textbf{Night} and inference time with the state-of-the-art methods. As shown in Tab.\ref{tab:compare}, among these methods, AANet-Faster RCNN\cite{chen2023attentive} has the best performance with achieving a log-average miss rate of 6.91\% on the \textbf{All} set.Though our AMFD-Faster RCNN with a log-average miss rate of 7.23\% and AMFD-RetinaNet with a log-average miss rate of 8.82\% are slightly worse than AANet\cite{chen2023attentive}, the 0.05s and 0.04s inference time already far outperform all existing methods. Comparisons show that AMFD strikes a good balance between model performance and model compression and achieves excellent results in both aspects.

\setlength{\parskip}{1em}
\noindent \textbf{Qualitative comparison.}
Fig.\ref{qualitative_vis_kaist} displays the qualitative comparisons in night scenarios between our AMFD-Faster RCNN with (b) MBNet\cite{zhou2020improving} and (c) ProbEn3\cite{chen2022multimodal}. The results show that images from these models produce unwanted false positive objects, such as red boxes in the background or replicated boxes on pedestrians, as well as false negative images (i.e. missed pedestrians). In contrast, our AMFD-Faster RCNN is more robust in detecting the night scene. This confirms what we present in Tab.\ref{tab:compare}, which shows that our model has the lowest $\textit{MR}^{-2}$ at night.

\subsection{Distillation on LLVIP Dataset}
\label{sec: LLVIP}
\setlength{\parskip}{0em}
\noindent \textbf{Comparison with other distillation methods.}
Many excellent methods have been proposed on how to perform efficient distillation in object detection tasks. In this section, we will compare the distillation results of MGD\cite{yang2022masked}, CWD\cite{shu2021channel}, PKD\cite{cao2022pkd}, DKRD\cite{ijcai2023p142}, FKD\cite{10198386} and our AMFD on the LLVIP dataset from the two-stream teacher network to the single-stream student network. We obtain results of state-of-the-art methods in traditional distillation for fusion features and our fusion distillation architecture. As shown in Tab.\ref{tab:distill_res_llvip}, under the fusion distillation architecture, our AMFD can increase mAP by 2.7\% which is better than other distillation methods. However, we found that not all distillation methods have better performance under the fusion distillation architecture such as DRKD\cite{ijcai2023p142} and MGD\cite{yang2022masked}, which suggests that the fusion distillation architecture and the MEA module are inseparable in our AMFD.

\begin{figure*}[h]
  \centering
  \captionsetup[subfloat]{labelsep=none,format=plain,labelformat=empty}
        \hspace{-4.0mm}
        \subfloat{
        \includegraphics[width=0.11\linewidth]{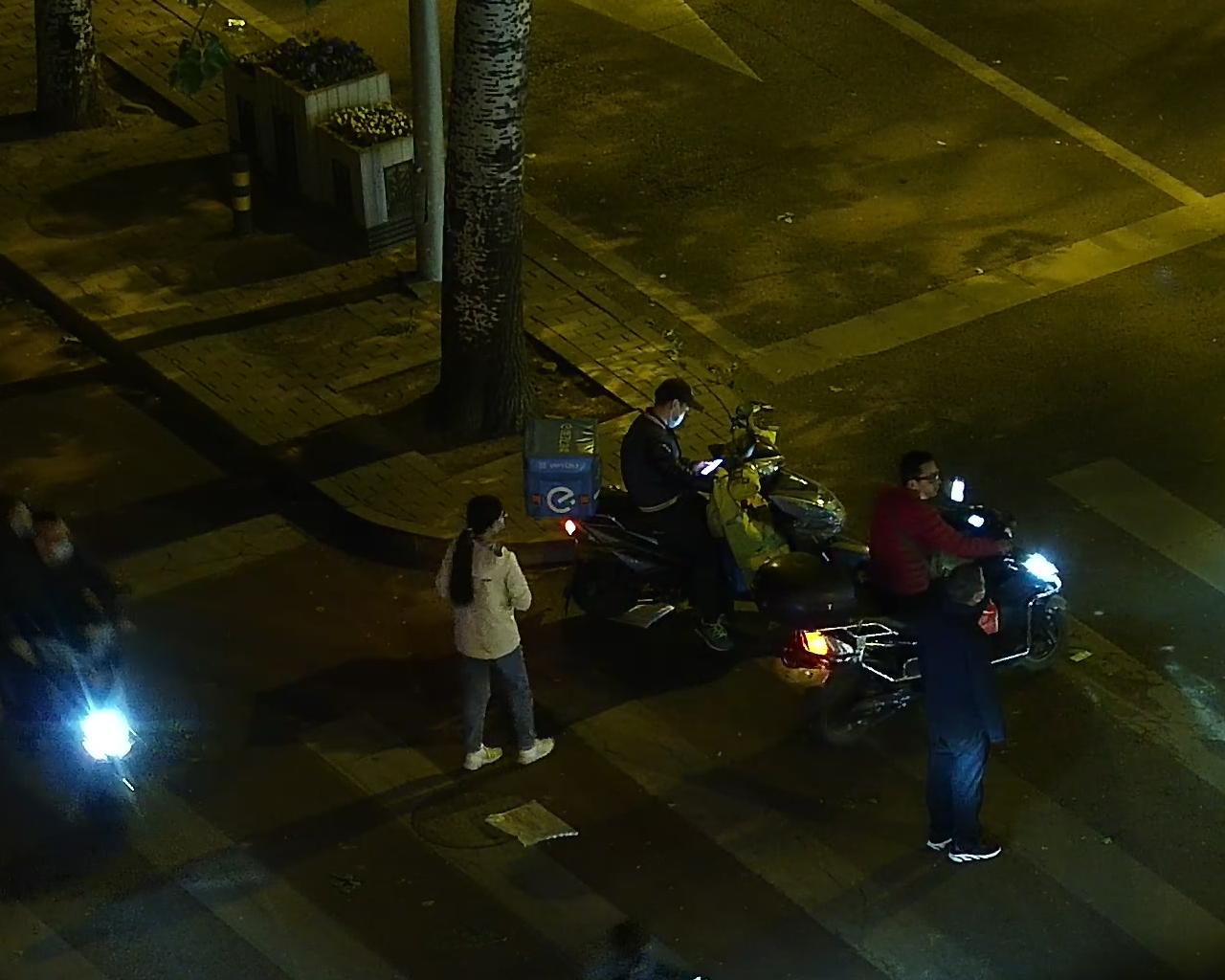}
        \hspace{-3.5mm}
        \label{fig: day_figure1}}
	\subfloat{
        \includegraphics[width=0.11\linewidth]{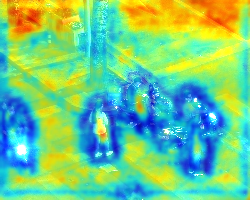}
        \hspace{-3.5mm}
        \label{fig: day_figure2}}
        \subfloat{
        \includegraphics[width=0.11\linewidth]{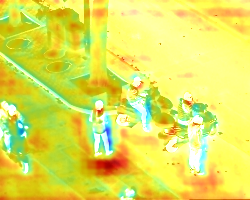}
        \hspace{-3.5mm}
        \label{fig: day_figure3}}
        \subfloat{
        \includegraphics[width=0.11\linewidth]{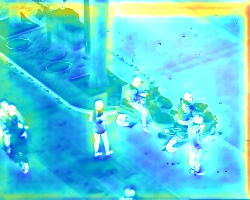}
        \hspace{-3.5mm}
        \label{fig: day_figure4}}
        \subfloat{
        \includegraphics[width=0.11\linewidth]{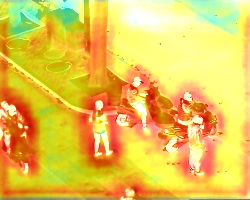}
        \hspace{-3.5mm}
        \label{fig: day_figure4}}
        \subfloat{
        \includegraphics[width=0.11\linewidth]{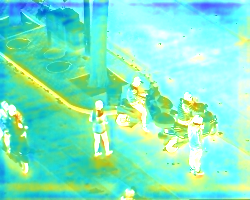}
        \hspace{-3.5mm}
        \label{fig: day_figure4}}
        \subfloat{
        \includegraphics[width=0.11\linewidth]{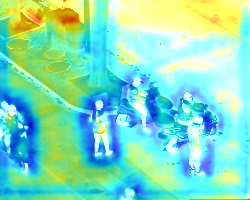}
        \hspace{-3.5mm}
        \label{fig: day_figure4}}
        \subfloat{
        \includegraphics[width=0.11\linewidth]{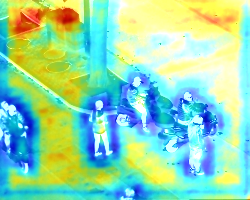}
        \hspace{-3.5mm}
        \label{fig: day_figure4}}
        \subfloat{
        \includegraphics[width=0.11\linewidth]{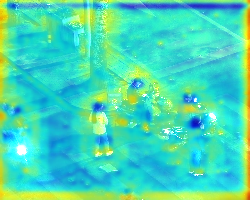}
        \hspace{-3.5mm}
        \label{fig: day_figure4}}
        \vskip -10pt
        \hspace{-4.0mm}
        \subfloat[RGB Input]{
        \includegraphics[width=0.11\linewidth]{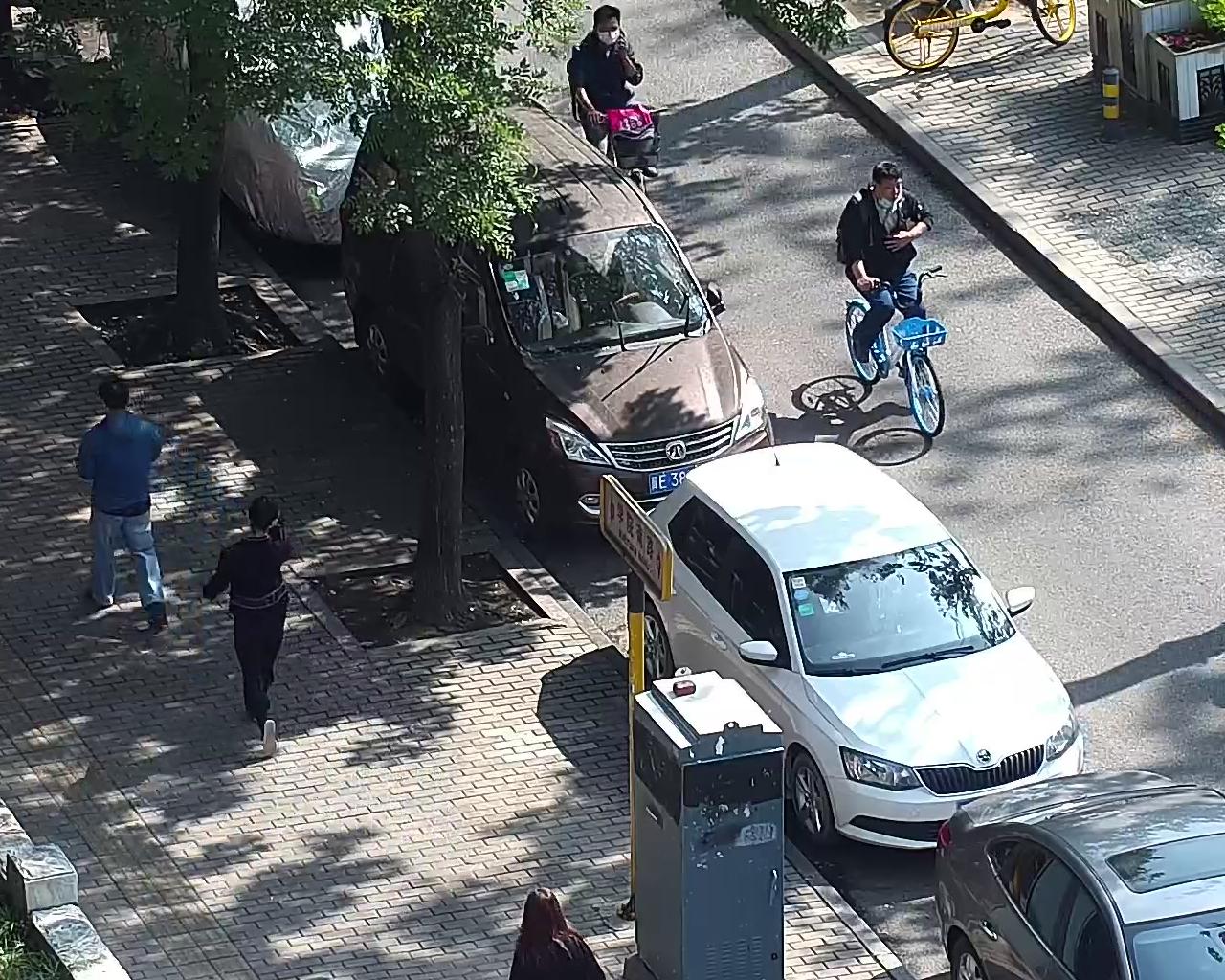}
        \hspace{-3.5mm}
        \label{fig: night_figure1}}
	\subfloat[Teacher]{
        \includegraphics[width=0.11\linewidth]{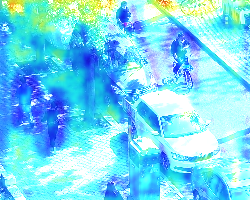}
        \hspace{-3.5mm}
        \label{fig: night_figure2}}
        \subfloat[Student]{
        \includegraphics[width=0.11\linewidth]{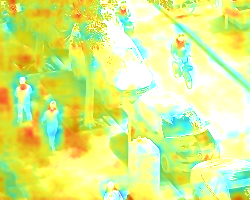}
        \hspace{-3.5mm}
        \label{fig: day_figure3}}
        \subfloat[w/DRKD($\uparrow$1.2)]{
        \includegraphics[width=0.11\linewidth]{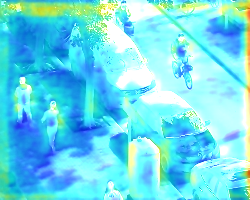}
        \hspace{-3.5mm}
        \label{fig: day_figure4}}
        \subfloat[w/FKD($\uparrow$1.4)]{
        \includegraphics[width=0.11\linewidth]{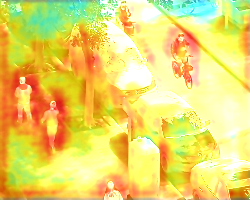}
        \hspace{-3.5mm}
        \label{fig: day_figure4}}
        \subfloat[w/MGD($\uparrow$1.6)]{
        \includegraphics[width=0.11\linewidth]{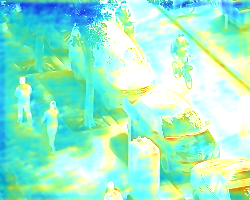}
        \hspace{-3.5mm}
        \label{fig: day_figure4}}
        \subfloat[w/CWD($\uparrow$1.9)]{
        \includegraphics[width=0.11\linewidth]{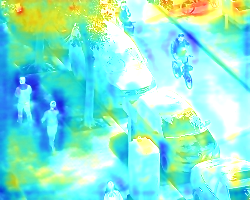}
        \hspace{-3.5mm}
        \label{fig: day_figure4}}
        \subfloat[w/PKD($\uparrow$2.0)]{
        \includegraphics[width=0.11\linewidth]{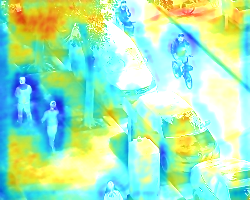}
        \hspace{-3.5mm}
        \label{fig: day_figure4}}
        \subfloat[\textbf{w/AMFD($\uparrow$2.7})]{
        \includegraphics[width=0.11\linewidth]{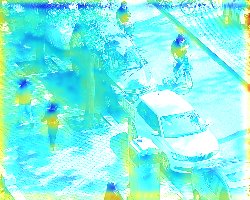}
        \hspace{-3.5mm}
        \label{fig: night_figure4}}
        \vspace{-0mm}
	\caption{Feature visualizations of fusion feature of the teacher network and the student networks with different distillation methods. It can be seen that the fusion feature of the student network distilled by AMFD is completely different from the other distillation methods. This suggests that AMFD enables student networks to generate fusion strategies independent of teacher networks.} 
	\label{feature_vis_llvip}
\end{figure*}

\begin{table}[h]
\centering
\caption{The comparison of distillation results of state-of-the-art distillation methods for detection on LLVIP dataset.} 
  \label{tab:distill_res_llvip}
\begin{tabular*}{0.475 \textwidth}{@{\extracolsep{\fill}}llllll@{}}
\toprule
                 Methods & Pub. Year & Backbone & mAP &  $\text{AP}_{50}$ & $\text{AP}_{75}$\\ \midrule \hline
                 \makecell[l]{F.RCNN \\\textbf{(Teacher)}} & \textbf{----}& \scriptsize{Res50 $\times2$} & \textbf{59.4} & 96.9 & 66.8 \\ 
                
               \makecell[l]{F.RCNN \\\textbf{(Student)}}  & \textbf{----}&\scriptsize{Res18 $\times1$} & 55.6 & 93.6& 60.4 \\ \hline
               \multicolumn{5}{l}{\emph{without fusion distillation architecture (distill fusion feature)}} & \\
               DRKD\cite{ijcai2023p142} & \scriptsize{IJCAI 2023} & \scriptsize{Res18 $\times1$} & 56.8(+1.2) & 94.6 & 62 \\
               FKD\cite{10198386} & \scriptsize{TPAMI 2023} & \scriptsize{Res18 $\times1$} & 57.0(+1.4) & 94.4 & 62.8 \\
               MGD\cite{yang2022masked} & \scriptsize{ECCV 2022}& \scriptsize{Res18 $\times1$} & 57.2(+1.6) & 94.5 & 62.7\\
               CWD\cite{shu2021channel}& \scriptsize{CVPR 2021} & \scriptsize{Res18 $\times1$} & 57.5 (+1.9) & 95.3 & 63.6 \\
               PKD\cite{cao2022pkd} & \scriptsize{NeurIPS 2022} & \scriptsize{Res18 $\times1$} & 57.6 (+2.0) & 94.5 &  64.3  \\ \hline 
                \multicolumn{3}{l}{\emph{with fusion distillation architecture}} & & & \\
                DRKD\cite{ijcai2023p142} & \scriptsize{IJCAI 2023} & \scriptsize{Res18 $\times1$} & 55.9(+0.3) & 94.2 & 60.4 \\
               MGD\cite{yang2022masked} & \scriptsize{ECCV 2022} & \scriptsize{Res18 $\times1$} & 56.3(+0.7) & 93.9 & 61.2\\
               FKD\cite{10198386} & \scriptsize{TPAMI 2023} & \scriptsize{Res18 $\times1$} & 57.1(+1.5) & 95.4 & 62.0 \\
               CWD\cite{shu2021channel} & \scriptsize{CVPR 2021} & \scriptsize{Res18 $\times1$} & 57.2 (+1.6) & 95.3 & 63.0 \\
               PKD\cite{cao2022pkd} & \scriptsize{NeurIPS 2022} & \scriptsize{Res18 $\times1$} & 57.7 (+2.1) & \textbf{95.5} &  63.0  \\ 
            \multicolumn{2}{l}{\hspace{-0.8em}\textbf{AFMD(ours)}} & \scriptsize{Res18 $\times1$} & \textbf{58.3 (+2.7)} & 95.2 & \textbf{64.9}\\\hline 
\bottomrule
\end{tabular*}
\end{table}

\noindent \textbf{Distillation on more detectors. }As shown in Tab\ref{tab:llvip_swin_detr}, we try distillation on a variety of different detectors. The results show that our distillation method is equally applicable to both Swin-based detectors and DETR-based detectors. We can find that distillation becomes less effective when the capacity gap between the teacher network and the student network is particularly large. 

\begin{table}[h]
    \centering
    \setlength\tabcolsep{4.5pt}
    \caption{Distillation results of various detectors on the LLVIP dataset. Bolding indicates the best distillation results for the specific student network.}
\label{tab:llvip_swin_detr}
    \begin{tabular}{l|l|lll}
    \toprule
                Teacher    & Student & mAP & $\text{\scriptsize{AP}}_{50}$ & $\text{\scriptsize{AP}}_{75}$ \\ \hline \midrule
    \multirow{6}{*}{\makecell[c]{Swin-L$\times$2 F.RCNN \\ (\textbf{mAP 62.1})}} & Swin-B$\times$1 F.RCNN & 57.7 & 94.0 & 65.6 \\
                      & w/AMFD & \textbf{59.3(+1.6)} & \textbf{96.0} & \textbf{66.7} \\ \cline{2-5}
    & Swin-S$\times$1 F.RCNN  & 57.2 & 94.8 & 63.7 \\
                      & w/AMFD & \textbf{60.4(+3.2)} & \textbf{96.0} & \textbf{69.2} \\ \cline{2-5}
            & Swin-T$\times$1 F.RCNN & 56.5 & 95.7 & 61.0 \\
            & w/AMFD & 57.3(+0.8) & 94.8 & 63.4 \\ \cline{1-5}
    \multirow{6}{*}{\makecell[c]{Swin-B$\times$2 F.RCNN \\ (\textbf{mAP 61.2})}} & Swin-B$\times$1 F.RCNN & 57.7 & 94.0 & 65.6 \\
                      & w/AMFD & 58.9(+1.2) & 94.8 & 66.4 \\ \cline{2-5}
    & Swin-S$\times$1 F.RCNN  & 57.2 & 94.8 & 63.7 \\
                      & w/AMFD & 59.4(+2.2) & 95.1 & 67.7 \\ \cline{2-5}
            & Swin-T$\times$1 F.RCNN & 56.5 & 95.7 & 61.0 \\
            & w/AMFD & 57.4(+0.9) & 95.0 & 63.8 \\ \cline{1-5}
    \multirow{4}{*}{\makecell[c]{Swin-S$\times$2 F.RCNN \\ (\textbf{mAP 60.6})}} & Swin-S$\times$1 F.RCNN & 57.2 & 94.8 & 63.7 \\
    & w/AMFD & 59.4(+2.2) & 95.3 & 67.8 \\ \cline{2-5}
    & Swin-T$\times$1 F.RCNN & 56.5 & 95.7 & 61.0 \\
            & w/ AMFD & \textbf{58.2(+1.7)} & \textbf{96.1} & \textbf{63.9} \\ \cline{1-5}
    \multirow{4}{*}{\makecell[c]{Res101$\times$2 DINO \\ (\textbf{mAP 65.5})}} & Res50$\times$1 DINO & 59.7 & 93.4 & 67.1 \\
                      & w/AMFD & \textbf{61.6(+1.9)} & \textbf{95.0} & \textbf{69.9}\\ \cline{2-5}
    & Res18$\times$1 DINO  & 59.1 & 93.7 & 66.3 \\
                      & w/AMFD & 60.6(+1.5) & 94.2 & 68.1 \\  \cline{1-5}
    \multirow{4}{*}{\makecell[c]{Res50$\times$2 DINO \\ (\textbf{mAP 63.1})}} & Res50$\times$1 DINO & 59.7 & 93.4 & 67.1 \\
                      & w/AMFD & 61.2(+1.5) & 94.7 & 70.0 \\ \cline{2-5}
    & Res18$\times$1 DINO  & 59.1 & 93.7 & 66.3 \\
                      & w/AMFD & \textbf{61.7(+2.6)} & \textbf{94.4} & \textbf{70.1} \\ \cline{1-5}
    \multirow{2}{*}{\makecell[c]{Swin-L$\times$2 DINO \\ (\textbf{mAP 66.2})}} & Swin-T$\times$1 DINO & 60.7 & 93.0 & 68.1 \\
                      & w/AMFD & \textbf{61.5(+0.8)} & \textbf{94.5} & \textbf{70.5} \\ \cline{1-5}\bottomrule
     \end{tabular}
    \end{table}

\begin{table*}[t]
\centering
\setlength\tabcolsep{4.5pt}
\caption{Comparison of distillation results between traditional and fusion distillation architectures on SMOD dataset. The "RGB", "TIR" and "Fusion" denote whether distill the RGB, TIR and Fusion Feature respectively.} 
\label{tab:SMOD_mAP_MR}
\begin{tabular}{@{}l|ccc|ll|lll@{}}
\toprule 
\multirow{2}{*}{Method} & \multicolumn{3}{c|}{Distill} & \multicolumn{2}{c|}{mAP} & \multirow{2}{*}{All} & \multirow{2}{*}{Day} & \multirow{2}{*}{Night} \\ \cline{2-4}

  \rule{0pt}{9pt}        & RGB & TIR & Fusion &All& NO &  &  & \\ \hline 
\makecell[l]{Faster-RCNN \textbf{(Teacher, 2$\times$R50)}} & \textbf{--} & \textbf{--} & \textbf{--} & 66.3 & 63.9  &2.17 & 2.91 & 1.55\\ 
\makecell[l]{Faster-RCNN \textbf{(Student, 1$\times$R18)}} & \textbf{--} & \textbf{--} & \textbf{--} & 60.6& 59.9  & 3.33 & 4.55 & 2.34 \\ \hline
\makecell[l]{Traditional Architecture(\textbf{MEA$\times 1$})} &  &  & \Checkmark & 63.5(+2.9)& 62.9  &  2.38($\downarrow$ 0.95) & 3.34 & 1.46\\
\makecell[l]{AMFD (\textbf{MEA$\times 2$},\textbf{ours})} & \Checkmark & \Checkmark &  & \textbf{64.1}(\textbf{+3.5})& \textbf{64.2}& \textbf{2.17($\downarrow$ 1.16)} & 3.05 & 1.44 \\ \hline\bottomrule
\end{tabular}
\end{table*}

\noindent \textbf{Qualitative comparison.} 
As shown in Fig.\ref{feature_vis_llvip}, we visualize the fusion feature maps of student networks distilled by previous state-of-the-art methods. We can see that pedestrians are well segmented by blue areas (lower values) in the feature map of the teacher network in the second column. However, student networks with CWD and PKD still remain feature fusion strategies based on the teacher network. The student network obtained by our AMFD in the last column represents the head of pedestrians with a blue region (low values) and the body of pedestrians with an orange region (high values), which is completely different from the fusion strategy of the teacher network and other distillation methods. This suggests that our AMFD can form a more robust and flexible fusion strategy independently from teacher networks, which better fits the structure of the student network than that of the teacher network.

\subsection{Distillation on SMOD Dataset}
\noindent \textbf{Analysis of distillation results.}
To validate the effectiveness of our method on our SMOD dataset, we compared the distillation results using the traditional distillation architecture with the distillation results using the AMFD. The traditional distillation architecture distills only the fusion features of the teacher network with one single MEA module. In Tab.\ref{tab:SMOD_mAP_MR} we can see that our AMFD improves the mAP of the student network by 3.5\%, which is 0.6\% more than the traditional distillation architecture that distills only fusion features. 
For $\textit{MR}^{-2}$, our distillation method also outperforms the traditional distillation architecture and obtains results that are on the same level as the teacher network. The above results illustrate the effectiveness of our proposed fusion distillation architecture. In addition to this we do separate evaluations for different occluded objects and the results show that the detection ability of the models are concentrated on "NO" objects.

\noindent \textbf{Heterogeneous distillation.}
AMFD gives greater independence to the student network than the usual feature distillation, which makes our distillation method very suitable for distillation between detectors with different architectures. From Tab.\ref{tab:smod_detr}, we can see that for different detector architectures and different backbones, AMFD is effective in improving the performance of the student network. This also illustrates the wide applicability of fusion distillation.

\begin{table}[h]
\centering
\caption{Heterogeneous distillation results of various detectors on the SMOD dataset. }
\label{tab:smod_detr}
    \begin{tabular}{l|l|lll}
    \toprule
                Teacher    & Student & mAP & $\text{AP}_{50}$ & $\text{AP}_{75}$ \\ \hline \midrule
    \multirow{4}{*}{\makecell[c]{Swin-B$\times$2 F.RCNN \\ (\textbf{mAP 65.9})}} & Res18$\times$1 F.RCNN & 60.6 & 88.6 & 71.3 \\
                      & w/AMFD & 61.9(+1.3) & 91.0 & 73.2 \\ \cline{2-5}
    & Res18$\times$1 Retina. & 59.8 & 89.5 & 69.1 \\
                      & w/AMFD & 62.0(+2.2) & 90.4 & 72.4 \\ \cline{1-5}
    \multirow{6}{*}{\makecell[c]{Res50$\times$2 F.RCNN \\ (\textbf{mAP 66.3})}} & Res18$\times$1 F.RCNN & 60.6 & 88.6 & 71.3 \\
                      & w/AMFD & 64.1(+3.5) & 95.0 & 69.9 \\ \cline{2-5}
    & Res18$\times$1 Retina. & 59.8 & 89.5 & 69.1 \\
                      & w/AMFD & 62.6(+2.8) & 90.6 & 73.2 \\ \cline{2-5}
                      & Res18$\times$1 DINO & 61.9 & 88.7 & 72.6 \\
                      & w/AMFD & 63.8(+1.9) & 89.5 & 74.5 \\\cline{1-5} \bottomrule
     \end{tabular}
    \end{table}

\begin{figure}[b]
  \centering
        \subfloat[Fused MEA]{
        \includegraphics[height=0.27\linewidth]{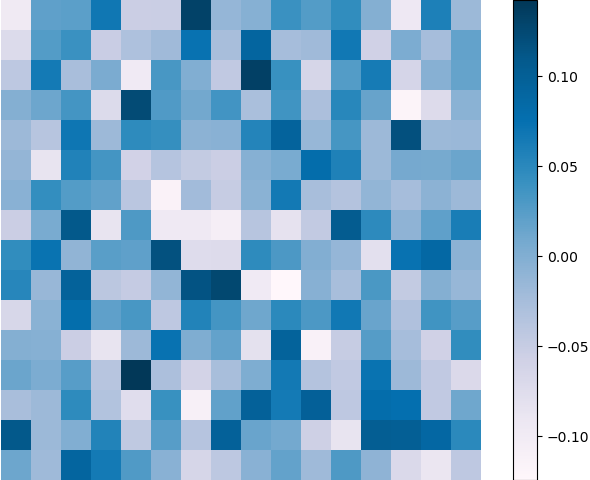}%
        \label{fig: channel_figure1}
        \hspace{-2mm}
        }
        \subfloat[TIR MEA]{
        \includegraphics[height=0.27\linewidth]{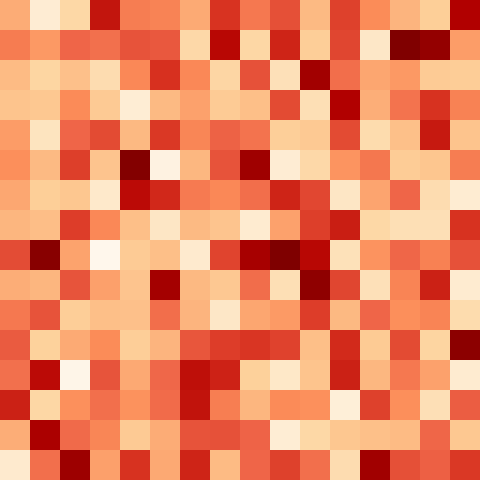}%
        \label{fig: channel_figure1}
        \hspace{-2mm}
        }
	\subfloat[RGB MEA]{
        \includegraphics[height=0.27\linewidth]{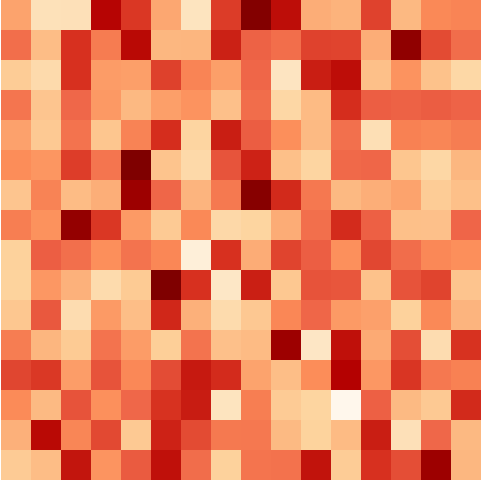}
        \hspace{-2mm}
        \label{fig: channel_figure2}}
        \subfloat{
        \includegraphics[height=0.27\linewidth]{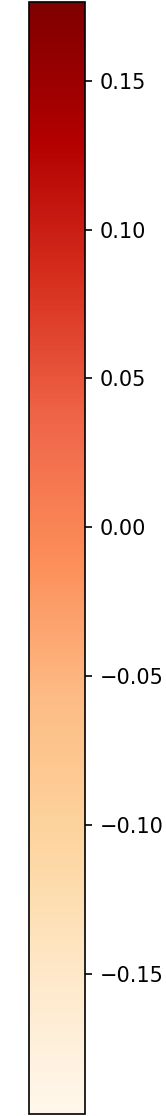}
        \hspace{-3mm}
        \label{fig: color_bar}}
	\caption{Channel weights visualizations of the Fused MEA module in traditional distillation and TIR, RGB MEA modules in our AMFD. This shows that in our fusion distillation architecture, the two MEA modules can adaptively extract different useful information in TIR and RGB modalities.}
	\label{mea_channel_vis_llvip}
\end{figure}

\noindent \textbf{Effectiveness of the proposed MEA module.} 
In the fusion distillation architecture, we use two MEA modules to force student networks to learn TIR and RGB features, respectively. In particular, we visualize the weights of the channels obtained in the MEA global feature extraction part (the $C \times 1 \times 1$ weights that we denote in Equation (\ref{equ: channel_weights}) by the symbol $\mathcal{W}(x)$). 
As shown in Fig.\ref{mea_channel_vis_llvip}, we can see that for different channels, there is a significant difference between the weights of the TIR MEA and the RGB MEA. This phenomenon of enhancement or weakness for different channels illustrates the effectiveness of our MEA module in fusion distillation architectures. In the traditional distillation architecture, we use only one MEA module to distill the fusion feature of teacher network. This generates only a single group of channel weights for the fusion features, which results in the missing useful latent information in both modalities.


\subsection{Distillation on SUNRGB-D Dataset}

\noindent  \textbf{Analysis of distillation results.}
To verify the generalizability of our method, we conduct simple experiments on a 2D object detection task where the inputs are visible and depth images. As shown in Tab.\ref{tab:distill_res_sunrgbd}, we can see that after distillation the performance of the student network is very close to that of the teacher network, with a 4.4\% increase in mAP compared to the baseline student network. The APs of Table, Chair and Desk even exceeded those of the teacher network, which reinforces the remarkable effectiveness of our distillation method. This suggests that our method is also applicable to the task of object detection in RGB-D. To ensure that our base model utilizes and fuses the two modalities for both depth and visible images, we conducted single-modality comparison experiments. This also ensures that our distillation method can indeed fuse multimodal features on the student network.

\begin{table*}[h]
\centering
\caption{The results of distillation and validation of multimodal data input effectiveness on the SUNRGB-D dataset.}
  \label{tab:distill_res_sunrgbd}
\begin{tabular*}{0.99 \textwidth}{@{\extracolsep{\fill}}llllllllll@{}}
\toprule
                 \multirow{2}{*}{Method} & \multirow{2}{*}{Backbone}& \multirow{2}{*}{Input} & \multicolumn{7}{c}{mAP} \\  
                 &   &  &  All & Bed & Table & Sofa & Chair & Toilet & Desk \\\midrule \hline
                  \makecell[l]{F.RCNN} & Res50 ($\times2$) & RGB & 30.6 & 44.2 & 28.5 & 32.1 & 36.6 & 51.9 & 10.7\\
                  \makecell[l]{F.RCNN} & Res50 ($\times2$) & Depth & 29.5 & 52.9 & 27.4 & 34.1 & 34.3 & 54.4 & 8.0\\
                 \makecell[l]{F.RCNN \textbf{(Tea.)}} &  Res50 ($\times2$) & RGB-D & \textbf{36.6} & \textbf{57.8} & \textbf{30.7} & \textbf{42.4} & \textbf{38.9} & \textbf{61.3} & \textbf{13.8}\\ \hline
                 \makecell[l]{F.RCNN} & Res18 ($\times1$) & RGB & 30.2 & 43.9 & 24.5 & 33.9 & 33.2 & \textbf{53.6} & 10.3\\
                 \makecell[l]{F.RCNN} & Res18 ($\times1$) & Depth & 30.5 & 46.3 & 25.7 & 34.2 & 33.3 & 53.3 & \textbf{11.4}\\
                 \makecell[l]{F.RCNN \textbf{(Stu.)}} &  Res18 ($\times1$) & RGB-D &32.1 & 50.9 & 28.3 & 36.5 & 34.9 & 53.3 & 11.1\\ 
                 F.RCNN \textbf{(Stu.)w/AMFD} &  Res18 ($\times1$) & RGB-D &\textbf{36.5}\textbf{(+4.4)} & \textbf{56.0} & \textbf{31.7} & \textbf{41.5} & \textbf{39.2} & \textbf{61.3} & \textbf{14.5} \\ \hline
                 
\bottomrule
\end{tabular*}
\end{table*}

\begin{figure}[h]
  \centering
  \captionsetup[subfloat]{format=plain}
        \subfloat[The visible image]{
        \includegraphics[width=0.325\linewidth]{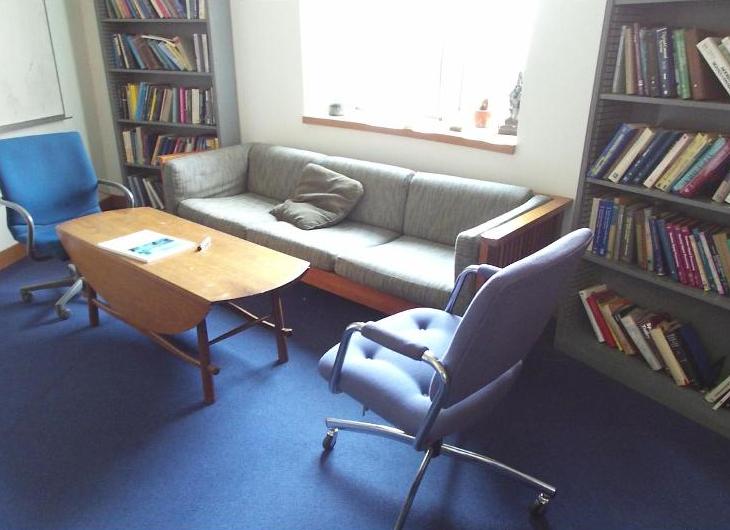}
        \hspace{-3.1mm}
        \label{fig: ori_rgb}}
	\subfloat[The depth image]{
        \includegraphics[width=0.325\linewidth]{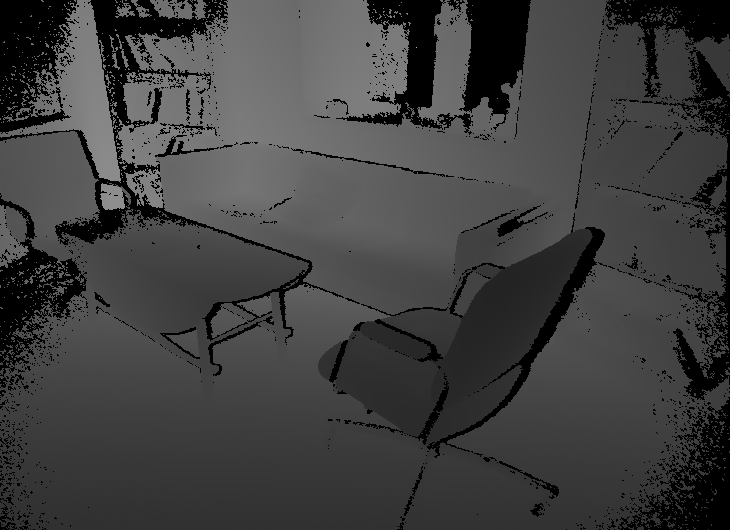}
        \hspace{-3.1mm}
        \label{fig: ori_depth}}
        \subfloat[Only RGB]{
        \includegraphics[width=0.325\linewidth]{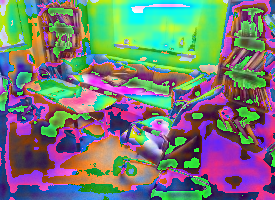}
        \hspace{-3.1mm}
        \label{fig: rgb}}
        \vskip -9.2pt
	\subfloat[Teacher]{
        \includegraphics[width=0.325\linewidth]{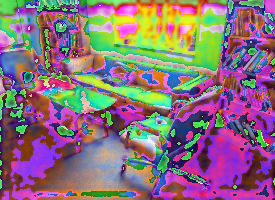}
        \hspace{-3.1mm}
        \label{fig: tea}}
        \subfloat[Student]{
        \includegraphics[width=0.325\linewidth]{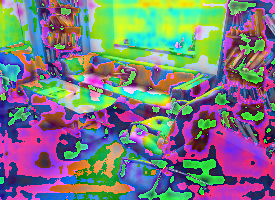}
        \hspace{-3.1mm}
        \label{fig: stu}}
        \subfloat[AMFD]{
        \includegraphics[width=0.325\linewidth]{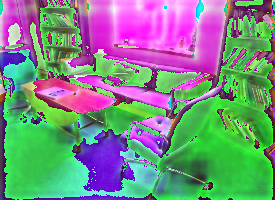}
        \hspace{-3.1mm}
        \label{fig: amfd}}
        \vspace{-1mm}
	\caption{The input image pairs and spatial attention maps for networks. "Only RGB" refers to student network using only RGB images as inputs. “AMFD” refers to the student network distilled by AMFD.}
	\label{vis_rgbd}
\end{figure}

\noindent\textbf{Qualitative comparison.} We perform visualizations to confirm the validity of AMFD in the RGB-D task. Purple and green represent high and low values, respectively. As shown in Fig.\ref{vis_rgbd}, the spatial attention representation of the student network distilled by the AMFD is quite different from that of the non-distilled networks (teacher networks, student networks, and student networks with only RGB inputs). The area where the table, chair and sofa are located of the distilled network is represented by low values (green). Attention is smoother and more focused than that of the non-distilled student network. This suggests that our method can be effectively migrated to other multispectral detection tasks.

\section{Conclusion}
In this paper, we proposed an adaptive modal fusion distillation (AMFD) framework. This framework adopts the fusion distillation architecture, which can significantly improve the performance of student network. This architecture allows for a fusion strategy of the student network is independent from the teacher network. The modal extraction alignment (MEA) module extracts the original modal feature based on focal and global attention mechanisms. This distillation method can efficiently improve the performance of the student network, leading to efficient compression of the teacher network, which can substantially reduce the inference time for multispectral networks. Experiments show that the simple student network with faster inference speed can perform as well as the teacher network with AMFD. This makes it more feasible to deploy multispectral pedestrian detection capabilities in embedded devices in the future.


\bibliographystyle{IEEEtran}
\bibliography{IEEEabrv, main.bib}

\begin{thebibliography}{10}
\providecommand{\url}[1]{#1}
\csname url@samestyle\endcsname
\providecommand{\newblock}{\relax}
\providecommand{\bibinfo}[2]{#2}
\providecommand{\BIBentrySTDinterwordspacing}{\spaceskip=0pt\relax}
\providecommand{\BIBentryALTinterwordstretchfactor}{4}
\providecommand{\BIBentryALTinterwordspacing}{\spaceskip=\fontdimen2\font plus
\BIBentryALTinterwordstretchfactor\fontdimen3\font minus \fontdimen4\font\relax}
\providecommand{\BIBforeignlanguage}[2]{{%
\expandafter\ifx\csname l@#1\endcsname\relax
\typeout{** WARNING: IEEEtran.bst: No hyphenation pattern has been}%
\typeout{** loaded for the language `#1'. Using the pattern for}%
\typeout{** the default language instead.}%
\else
\language=\csname l@#1\endcsname
\fi
#2}}
\providecommand{\BIBdecl}{\relax}
\BIBdecl

\bibitem{yang2018real}
Z.~Yang, J.~Li, and H.~Li, ``Real-time pedestrian and vehicle detection for autonomous driving,'' in \emph{IEEE Intell. Vehicles Symp.}\hskip 1em plus 0.5em minus 0.4em\relax IEEE, 2018, pp. 179--184.

\bibitem{bilal2016low}
M.~Bilal, A.~Khan, M.~U.~K. Khan, and C.-M. Kyung, ``A low-complexity pedestrian detection framework for smart video surveillance systems,'' \emph{IEEE Trans. Circuits Syst. Video Technol.}, vol.~27, no.~10, pp. 2260--2273, 2016.

\bibitem{hwang2015multispectral}
S.~Hwang, J.~Park, N.~Kim, Y.~Choi, and I.~So~Kweon, ``Multispectral pedestrian detection: Benchmark dataset and baseline,'' in \emph{Proc. IEEE Conf. Comput. Vis. Pattern Recognit.}, 2015, pp. 1037--1045.

\bibitem{2016Deep}
K.~He, X.~Zhang, S.~Ren, and J.~Sun, ``Deep residual learning for image recognition,'' in \emph{Proc. IEEE Conf. Comput. Vis. Pattern Recognit.}, 2016, pp. 770--778.

\bibitem{2017Faster}
S.~Ren, K.~He, R.~Girshick, and J.~Sun, ``Faster r-cnn: Towards real-time object detection with region proposal networks,'' \emph{IEEE Trans. Pattern Anal. Mach. Intell.}, vol.~39, no.~6, pp. 1137--1149, 2017.

\bibitem{lin2017focal}
T.-Y. Lin, P.~Goyal, R.~Girshick, K.~He, and P.~Doll{\'a}r, ``Focal loss for dense object detection,'' in \emph{Proc. IEEE Int. Conf. Comput. Vis.}, 2017, pp. 2980--2988.

\bibitem{xu2017learning}
D.~Xu, W.~Ouyang, E.~Ricci, X.~Wang, and N.~Sebe, ``Learning cross-modal deep representations for robust pedestrian detection,'' in \emph{Proc. IEEE Conf. Comput. Vis. Pattern Recognit.}, 2017, pp. 5363--5371.

\bibitem{zhang2021guided}
H.~Zhang, E.~Fromont, S.~Lef{\`e}vre, and B.~Avignon, ``Guided attentive feature fusion for multispectral pedestrian detection,'' in \emph{Proc. IEEE/CVF Winter Conf. Appl. Comput. Vis.}, 2021, pp. 72--80.

\bibitem{chen2022multimodal}
Y.-T. Chen, J.~Shi, Z.~Ye, C.~Mertz, D.~Ramanan, and S.~Kong, ``Multimodal object detection via probabilistic ensembling,'' in \emph{Proc. Eur. Conf. Comput. Vis.}\hskip 1em plus 0.5em minus 0.4em\relax Springer, 2022, pp. 139--158.

\bibitem{yang2022baanet}
X.~Yang, Y.~Qian, H.~Zhu, C.~Wang, and M.~Yang, ``Baanet: Learning bi-directional adaptive attention gates for multispectral pedestrian detection,'' in \emph{Proc. IEEE. Int. Conf. Rob. Autom.}\hskip 1em plus 0.5em minus 0.4em\relax IEEE, 2022, pp. 2920--2926.

\bibitem{chen2023attentive}
N.~Chen, J.~Xie, J.~Nie, J.~Cao, Z.~Shao, and Y.~Pang, ``Attentive alignment network for multispectral pedestrian detection,'' in \emph{Proc. ACM Int. Conf. Multimed.}, 2023, pp. 3787--3795.

\bibitem{zhang2023tfdet}
X.~Zhang, X.~Zhang, Z.~Sheng, and H.-L. Shen, ``Tfdet: Target-aware fusion for rgb-t pedestrian detection,'' \emph{arXiv preprint arXiv:2305.16580}, 2023.

\bibitem{li2023multiscale}
R.~Li, J.~Xiang, F.~Sun, Y.~Yuan, L.~Yuan, and S.~Gou, ``Multiscale cross-modal homogeneity enhancement and confidence-aware fusion for multispectral pedestrian detection,'' \emph{IEEE Trans. Multimedia}, 2023.

\bibitem{li2022confidence}
Q.~Li, C.~Zhang, Q.~Hu, H.~Fu, and P.~Zhu, ``Confidence-aware fusion using dempster-shafer theory for multispectral pedestrian detection,'' \emph{IEEE Trans. on Multimedia}, 2022.

\bibitem{kruthiventi2017low}
S.~S. Kruthiventi, P.~Sahay, and R.~Biswal, ``Low-light pedestrian detection from rgb images using multi-modal knowledge distillation,'' in \emph{Proc. IEEE Int. Conf. Image Process.}\hskip 1em plus 0.5em minus 0.4em\relax IEEE, 2017, pp. 4207--4211.

\bibitem{luo2018graph}
Z.~Luo, J.-T. Hsieh, L.~Jiang, J.~C. Niebles, and L.~Fei-Fei, ``Graph distillation for action detection with privileged modalities,'' in \emph{Proc. Eur. Conf. Comput. Vis.}, 2018, pp. 166--183.

\bibitem{garcia2019learning}
N.~C. Garcia, P.~Morerio, and V.~Murino, ``Learning with privileged information via adversarial discriminative modality distillation,'' \emph{IEEE Trans. Pattern Anal. Mach. Intell.}, vol.~42, no.~10, pp. 2581--2593, 2019.

\bibitem{liu2021deep}
T.~Liu, K.-M. Lam, R.~Zhao, and G.~Qiu, ``Deep cross-modal representation learning and distillation for illumination-invariant pedestrian detection,'' \emph{IEEE Trans. Circuits Syst. Video Technol.}, vol.~32, no.~1, pp. 315--329, 2021.

\bibitem{zhang2022low}
H.~Zhang, E.~Fromont, S.~Lef{\`e}vre, and B.~Avignon, ``Low-cost multispectral scene analysis with modality distillation,'' in \emph{Proc. IEEE/CVF Winter Conf. Appl. Comput. Vis.}, 2022, pp. 803--812.

\bibitem{mirzadeh2020improved}
S.~I. Mirzadeh, M.~Farajtabar, A.~Li, N.~Levine, A.~Matsukawa, and H.~Ghasemzadeh, ``Improved knowledge distillation via teacher assistant,'' in \emph{Proc. AAAI Conf. Artif. Intell.}, vol.~34, no.~04, 2020, pp. 5191--5198.

\bibitem{gao2021residual}
M.~Gao, Y.~Wang, and L.~Wan, ``Residual error based knowledge distillation,'' \emph{Neurocomputing}, vol. 433, pp. 154--161, 2021.

\bibitem{cao2019gcnet}
Y.~Cao, J.~Xu, S.~Lin, F.~Wei, and H.~Hu, ``Gcnet: Non-local networks meet squeeze-excitation networks and beyond,'' in \emph{Proc. IEEE Int. Conf. Comput. Vis.}, 2019, pp. 1971--1980.

\bibitem{jia2021llvip}
X.~Jia, C.~Zhu, M.~Li, W.~Tang, and W.~Zhou, ``Llvip: A visible-infrared paired dataset for low-light vision,'' in \emph{Proc. IEEE Int. Conf. Comput. Vis.}, 2021, pp. 3496--3504.

\bibitem{song2015sun}
S.~Song, S.~P. Lichtenberg, and J.~Xiao, ``Sun rgb-d: A rgb-d scene understanding benchmark suite,'' in \emph{Proc. IEEE Conf. Comput. Vis. Pattern Recognit.}, 2015, pp. 567--576.

\bibitem{zhou2020improving}
K.~Zhou, L.~Chen, and X.~Cao, ``Improving multispectral pedestrian detection by addressing modality imbalance problems,'' in \emph{Proc. Eur. Conf. Comput. Vis.}\hskip 1em plus 0.5em minus 0.4em\relax Springer, 2020, pp. 787--803.

\bibitem{li2018multispectral}
C.~Li, D.~Song, R.~Tong, and M.~Tang, ``Multispectral pedestrian detection via simultaneous detection and segmentation,'' \emph{arXiv preprint arXiv:1808.04818}, 2018.

\bibitem{xie2022learning}
J.~Xie, R.~M. Anwer, H.~Cholakkal, J.~Nie, J.~Cao, J.~Laaksonen, and F.~S. Khan, ``Learning a dynamic cross-modal network for multispectral pedestrian detection,'' in \emph{Proc. ACM Int. Conf. Multimed.}, 2022, pp. 4043--4052.

\bibitem{hinton2015distilling}
G.~Hinton, O.~Vinyals, and J.~Dean, ``Distilling the knowledge in a neural network,'' \emph{arXiv preprint arXiv:1503.02531}, 2015.

\bibitem{adriana2015fitnets}
R.~Adriana, B.~Nicolas, K.~S. Ebrahimi, C.~Antoine, G.~Carlo, and B.~Yoshua, ``Fitnets: Hints for thin deep nets,'' \emph{Proc. Int. Conf. Learn. Represent.}, vol.~2, no.~3, p.~1, 2015.

\bibitem{zagoruyko2016paying}
S.~Zagoruyko and N.~Komodakis, ``Paying more attention to attention: Improving the performance of convolutional neural networks via attention transfer,'' \emph{arXiv preprint arXiv:1612.03928}, 2016.

\bibitem{wang2019distilling}
T.~Wang, L.~Yuan, X.~Zhang, and J.~Feng, ``Distilling object detectors with fine-grained feature imitation,'' in \emph{Proc. IEEE Conf. Comput. Vis. Pattern Recognit.}, 2019, pp. 4933--4942.

\bibitem{guo2021distilling}
J.~Guo, K.~Han, Y.~Wang, H.~Wu, X.~Chen, C.~Xu, and C.~Xu, ``Distilling object detectors via decoupled features,'' in \emph{Proc. IEEE Conf. Comput. Vis. Pattern Recognit.}, 2021, pp. 2154--2164.

\bibitem{gonzalez2016pedestrian}
A.~Gonz{\'a}lez, Z.~Fang, Y.~Socarras, J.~Serrat, D.~V{\'a}zquez, J.~Xu, and A.~M. L{\'o}pez, ``Pedestrian detection at day/night time with visible and fir cameras: A comparison,'' \emph{Sensors}, vol.~16, no.~6, p. 820, 2016.

\bibitem{liu2022target}
J.~Liu, X.~Fan, Z.~Huang, G.~Wu, R.~Liu, W.~Zhong, and Z.~Luo, ``Target-aware dual adversarial learning and a multi-scenario multi-modality benchmark to fuse infrared and visible for object detection,'' in \emph{Proc. IEEE Conf. Comput. Vis. Pattern Recognit.}, 2022, pp. 5802--5811.

\bibitem{lin2017feature}
T.-Y. Lin, P.~Doll{\'a}r, R.~Girshick, K.~He, B.~Hariharan, and S.~Belongie, ``Feature pyramid networks for object detection,'' in \emph{Proc. IEEE Conf. Comput. Vis. Pattern Recognit.}, 2017, pp. 2117--2125.

\bibitem{zhang2019weakly}
L.~Zhang, X.~Zhu, X.~Chen, X.~Yang, Z.~Lei, and Z.~Liu, ``Weakly aligned cross-modal learning for multispectral pedestrian detection,'' in \emph{Proc. IEEE Int. Conf. Comput. Vis.}, 2019, pp. 5127--5137.

\bibitem{liu2016multispectral}
J.~Liu, S.~Zhang, S.~Wang, and D.~N. Metaxas, ``Multispectral deep neural networks for pedestrian detection,'' \emph{arXiv preprint arXiv:1611.02644}, 2016.

\bibitem{lin2014microsoft}
T.-Y. Lin, M.~Maire, S.~Belongie, J.~Hays, P.~Perona, D.~Ramanan, P.~Doll{\'a}r, and C.~L. Zitnick, ``Microsoft coco: Common objects in context,'' in \emph{Proc. Eur. Conf. Comput. Vis.}\hskip 1em plus 0.5em minus 0.4em\relax Springer, 2014, pp. 740--755.

\bibitem{chen2019mmdetection}
K.~Chen, J.~Wang, J.~Pang, Y.~Cao, Y.~Xiong, X.~Li, S.~Sun, W.~Feng, Z.~Liu, J.~Xu \emph{et~al.}, ``Mmdetection: Open mmlab detection toolbox and benchmark,'' \emph{arXiv preprint arXiv:1906.07155}, 2019.

\bibitem{paszke2017automatic}
A.~Paszke, S.~Gross, S.~Chintala, G.~Chanan, E.~Yang, Z.~DeVito, Z.~Lin, A.~Desmaison, L.~Antiga, and A.~Lerer, ``Automatic differentiation in pytorch,'' 2017.

\bibitem{zhangdino}
\BIBentryALTinterwordspacing
H.~Zhang, F.~Li, S.~Liu, L.~Zhang, H.~Su, J.~Zhu, L.~Ni, and H.-Y. Shum, ``{DINO}: {DETR} with improved denoising anchor boxes for end-to-end object detection,'' in \emph{The Eleventh International Conference on Learning Representations}, 2023. [Online]. Available: \url{https://openreview.net/forum?id=3mRwyG5one}
\BIBentrySTDinterwordspacing

\bibitem{liu2021swin}
Z.~Liu, Y.~Lin, Y.~Cao, H.~Hu, Y.~Wei, Z.~Zhang, S.~Lin, and B.~Guo, ``Swin transformer: Hierarchical vision transformer using shifted windows,'' in \emph{Proc. IEEE Int. Conf. Comput. Vis.}, 2021, pp. 10\,012--10\,022.

\bibitem{lee2022cross}
W.-Y. Lee, L.~Jovanov, and W.~Philips, ``Cross-modality attention and multimodal fusion transformer for pedestrian detection,'' in \emph{Proc. Eur. Conf. Comput. Vis.}\hskip 1em plus 0.5em minus 0.4em\relax Springer, 2022, pp. 608--623.

\bibitem{russakovsky2015imagenet}
O.~Russakovsky, J.~Deng, H.~Su, J.~Krause, S.~Satheesh, S.~Ma, Z.~Huang, A.~Karpathy, A.~Khosla, M.~Bernstein \emph{et~al.}, ``Imagenet large scale visual recognition challenge,'' \emph{Int. J. Comput. Vision}, vol. 115, pp. 211--252, 2015.

\bibitem{kim2024causal}
T.~Kim, S.~Shin, Y.~Yu, H.~G. Kim, and Y.~M. Ro, ``Causal mode multiplexer: A novel framework for unbiased multispectral pedestrian detection,'' in \emph{Proc. IEEE Conf. Comput. Vis. Pattern Recognit.}, 2024, pp. 26\,784--26\,793.

\bibitem{zhu2023multi}
Y.~Zhu, X.~Sun, M.~Wang, and H.~Huang, ``Multi-modal feature pyramid transformer for rgb-infrared object detection,'' \emph{IEEE Trans. Intell. Transp. Syst.}, vol.~24, no.~9, pp. 9984--9995, 2023.

\bibitem{kim2021uncertainty}
J.~U. Kim, S.~Park, and Y.~M. Ro, ``Uncertainty-guided cross-modal learning for robust multispectral pedestrian detection,'' \emph{IEEE Trans. Circuits Syst. Video Technol.}, vol.~32, no.~3, pp. 1510--1523, 2021.

\bibitem{liu2024region}
Y.~Liu, C.~Hu, B.~Zhao, Y.~Huang, and X.~Zhang, ``Region-based illumination-temperature awareness and cross-modality enhancement for multispectral pedestrian detection,'' \emph{IEEE Trans. Intell. Veh.}, 2024.

\bibitem{li2023stabilizing}
Q.~Li, C.~Zhang, Q.~Hu, P.~Zhu, H.~Fu, and L.~Chen, ``Stabilizing multispectral pedestrian detection with evidential hybrid fusion,'' \emph{IEEE Trans. Circuits Syst. Video Technol.}, 2023.

\bibitem{yang2022masked}
Z.~Yang, Z.~Li, M.~Shao, D.~Shi, Z.~Yuan, and C.~Yuan, ``Masked generative distillation,'' in \emph{Proc. Eur. Conf. Comput. Vis.}\hskip 1em plus 0.5em minus 0.4em\relax Springer, 2022, pp. 53--69.

\bibitem{shu2021channel}
C.~Shu, Y.~Liu, J.~Gao, Z.~Yan, and C.~Shen, ``Channel-wise knowledge distillation for dense prediction,'' in \emph{Proc. IEEE Int. Conf. Comput. Vis.}, 2021, pp. 5311--5320.

\bibitem{cao2022pkd}
W.~Cao, Y.~Zhang, J.~Gao, A.~Cheng, K.~Cheng, and J.~Cheng, ``Pkd: General distillation framework for object detectors via pearson correlation coefficient,'' \emph{Adv. neural inf. proces. syst.}, vol.~35, pp. 15\,394--15\,406, 2022.

\bibitem{ijcai2023p142}
\BIBentryALTinterwordspacing
Z.-L. Ni, F.~Yang, S.~Wen, and G.~Zhang, ``Dual relation knowledge distillation for object detection,'' in \emph{Proceedings of the Thirty-Second International Joint Conference on Artificial Intelligence, {IJCAI-23}}, E.~Elkind, Ed.\hskip 1em plus 0.5em minus 0.4em\relax Int. Joint Conf. Artif. Intell. Org., 8 2023, pp. 1276--1284, main Track. [Online]. Available: \url{https://doi.org/10.24963/ijcai.2023/142}
\BIBentrySTDinterwordspacing

\bibitem{10198386}
L.~Zhang and K.~Ma, ``Structured knowledge distillation for accurate and efficient object detection,'' \emph{IEEE Trans. Pattern Anal. Mach. Intell.}, vol.~45, no.~12, pp. 15\,706--15\,724, 2023.

\end{thebibliography}











\vspace{-10mm}
\begin{IEEEbiography}[{\includegraphics[width=1in,height=1.25in,clip,keepaspectratio]{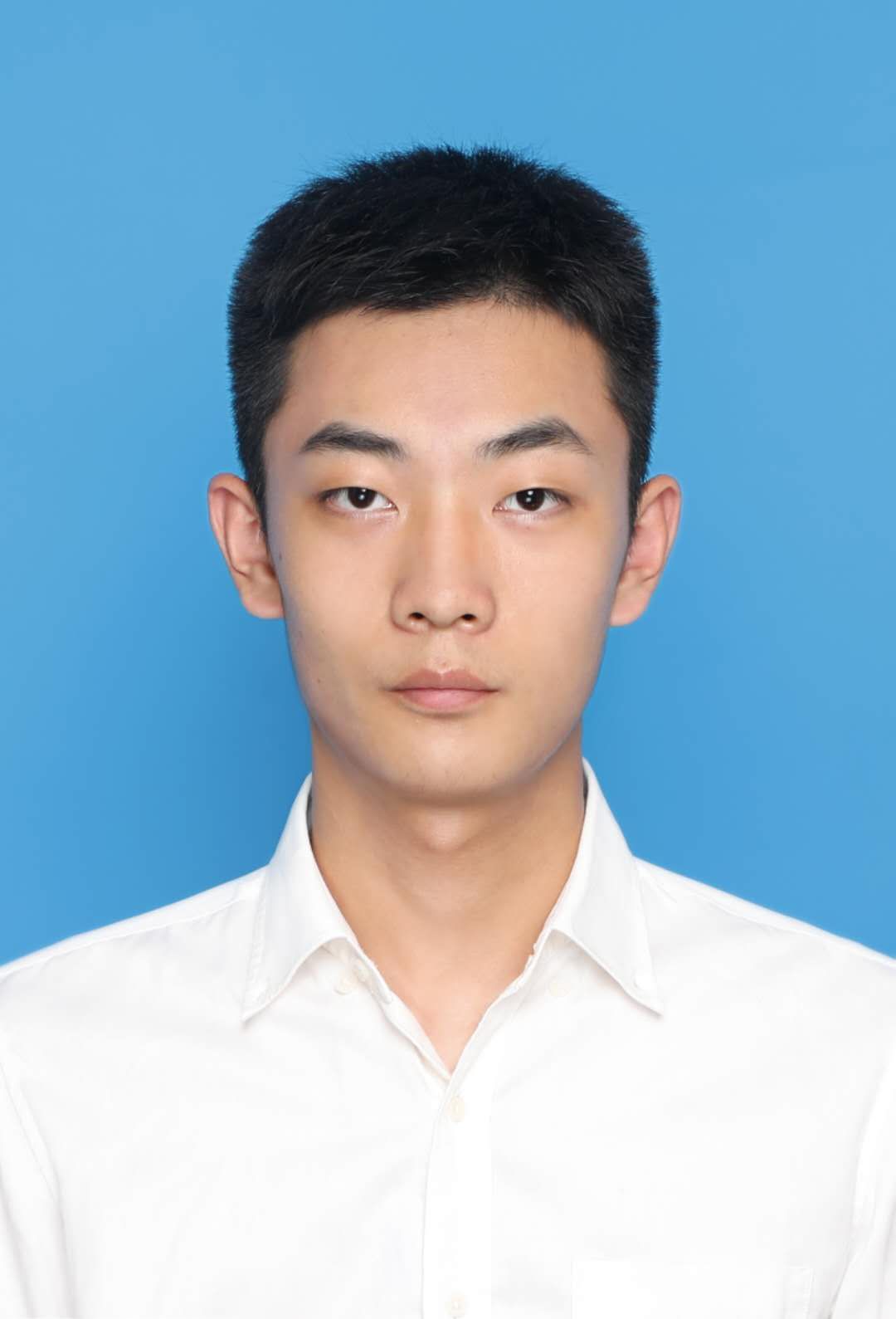}}]{Zizhao Chen}
received the B.S. degree in automation from Shanghai Jiao Tong University, Shanghai,
China, in 2024. He is currently working toward the M.S. degree in Institute of Artificial Intelligence and Robotics of Xi'an Jiao Tong University.

His main research interests include computer vision, pattern recognition, machine learning, and their applications in intelligent transportation systems.
\end{IEEEbiography}
\vspace{-10mm}
\begin{IEEEbiography}[{\includegraphics[width=1in,height=1.25in,clip,keepaspectratio]{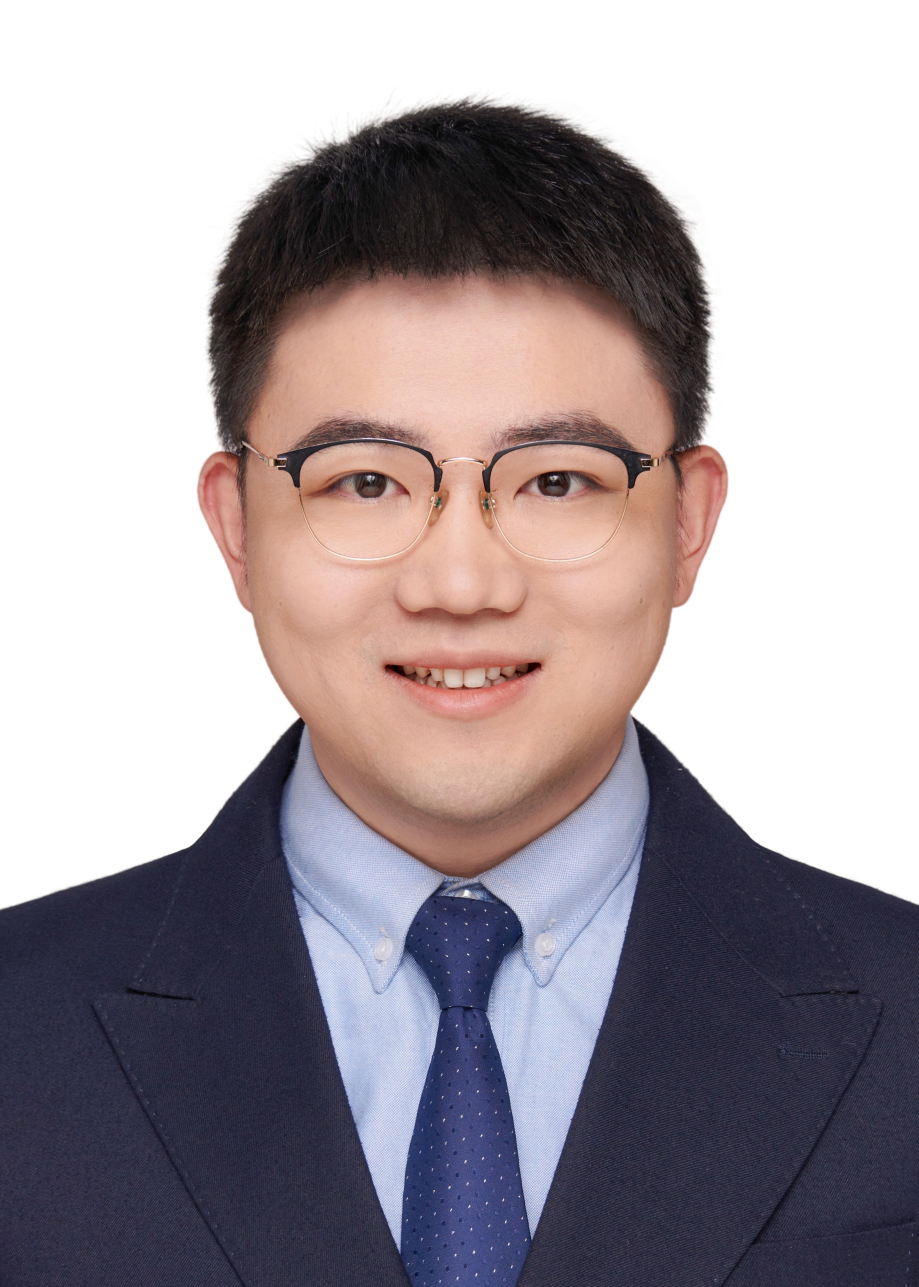}}]{Yeqiang Qian}
received a Ph.D. degree in control science and engineering from Shanghai Jiao Tong
University, Shanghai, China, in 2020. He is currently a Tenure Track Associate Professor in the Department of Automation at Shanghai Jiao Tong University, Shanghai, China.

His main research interests include computer vision, pattern recognition, machine learning, and their applications in intelligent transportation systems.
\end{IEEEbiography}
\vspace{-10mm}
\begin{IEEEbiography}[{\includegraphics[width=1in,height=1.25in,clip,keepaspectratio]{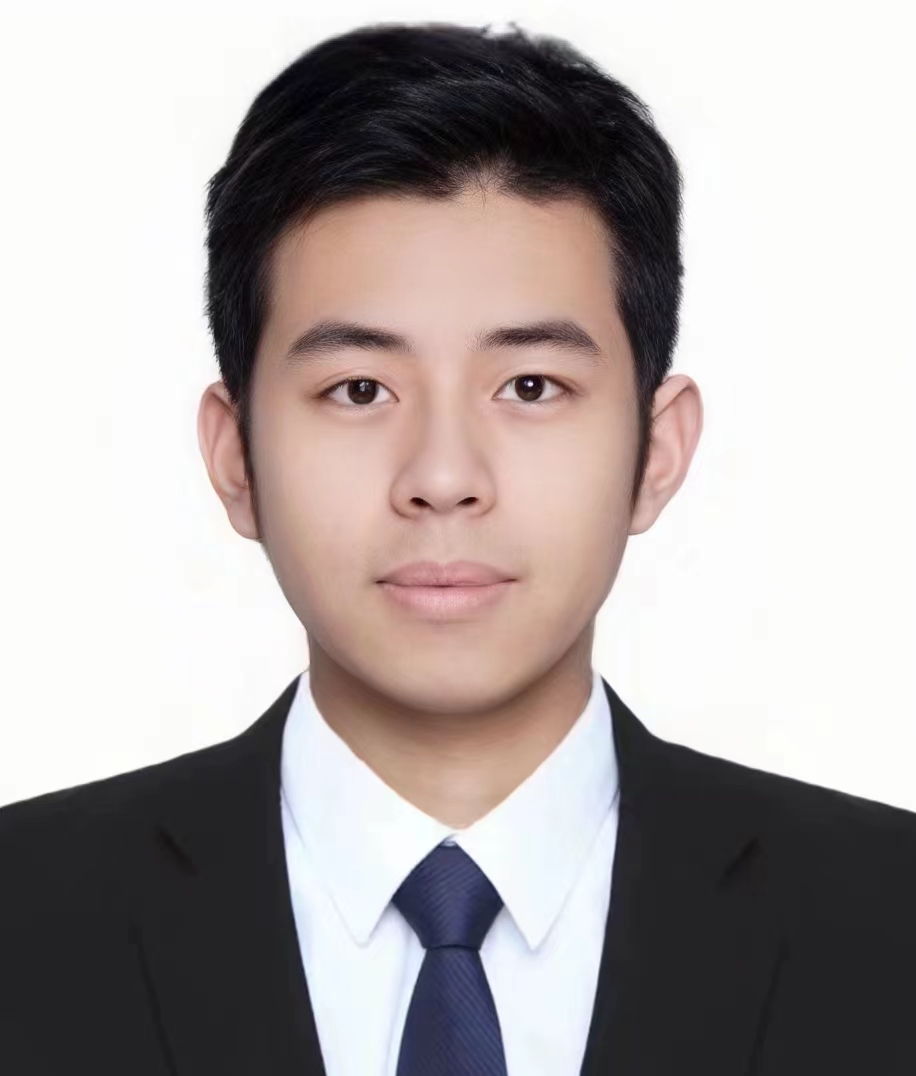}}]{Xiaoxiao Yang}
 received the B.S. degree in automation from Tongji University, Shanghai, China, in 2021. He is currently pursuing the M.S. degree in Shanghai Jiao Tong University, Shanghai, China.

His main research interests include computer vision,  deep learning, and their applications in intelligent transportation systems.
\end{IEEEbiography}
\vspace{-10mm}

\begin{IEEEbiography}[{\includegraphics[width=1in,height=1.25in,clip,keepaspectratio]{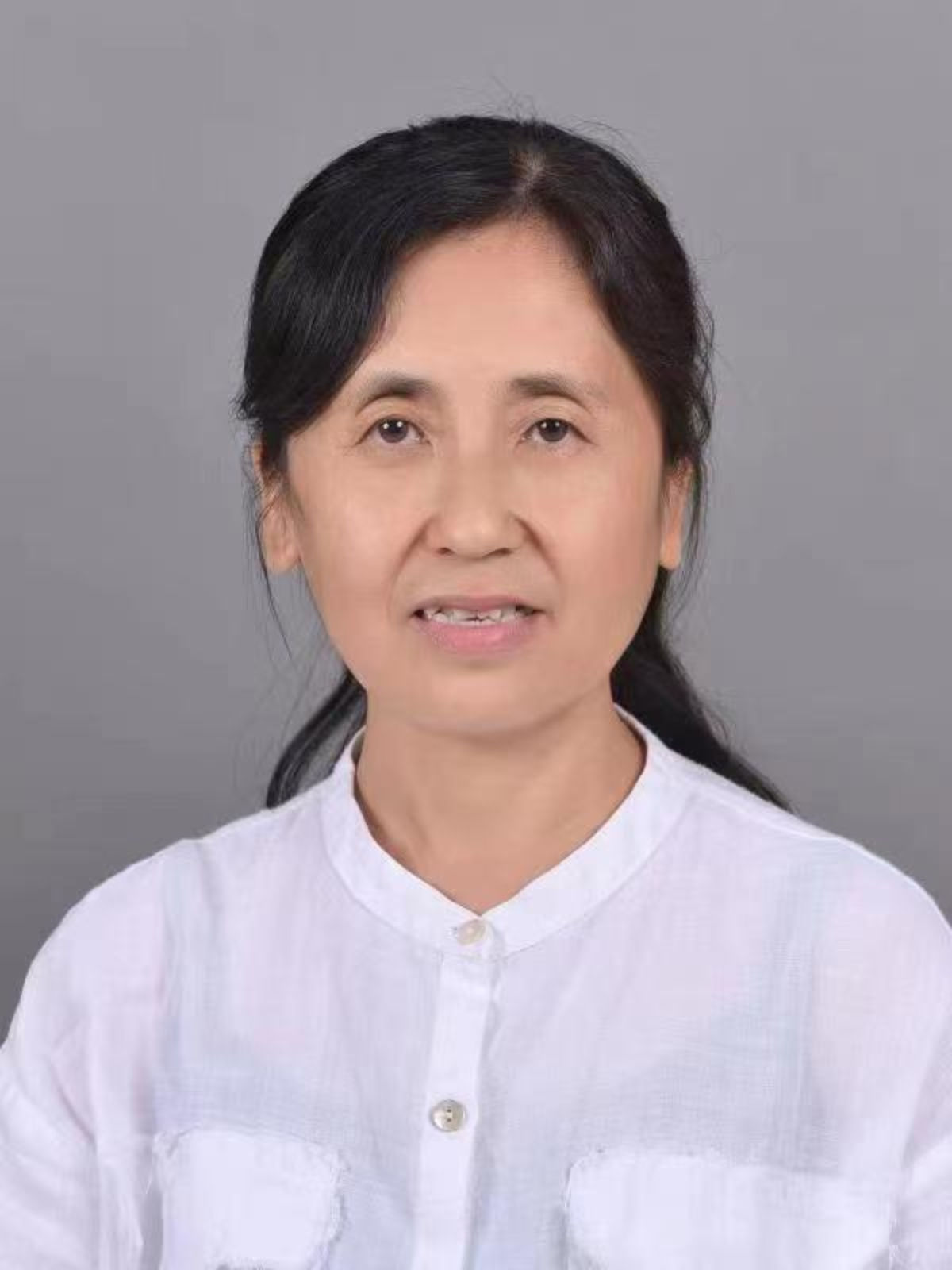}}]{Chunxiang Wang}
received his Ph.D. degree in mechanical engineering from the Harbin Institute of
Technology, Harbin, China, in 1999.

She is currently an Associate Professor with the Department of Automation at Shanghai Jiao Tong University, Shanghai, China. Her research interests include robotic technology and electromechanical integration.
\end{IEEEbiography}
\vspace{-10mm}

\begin{IEEEbiography}[{\includegraphics[width=1in,height=1.25in,clip,keepaspectratio]{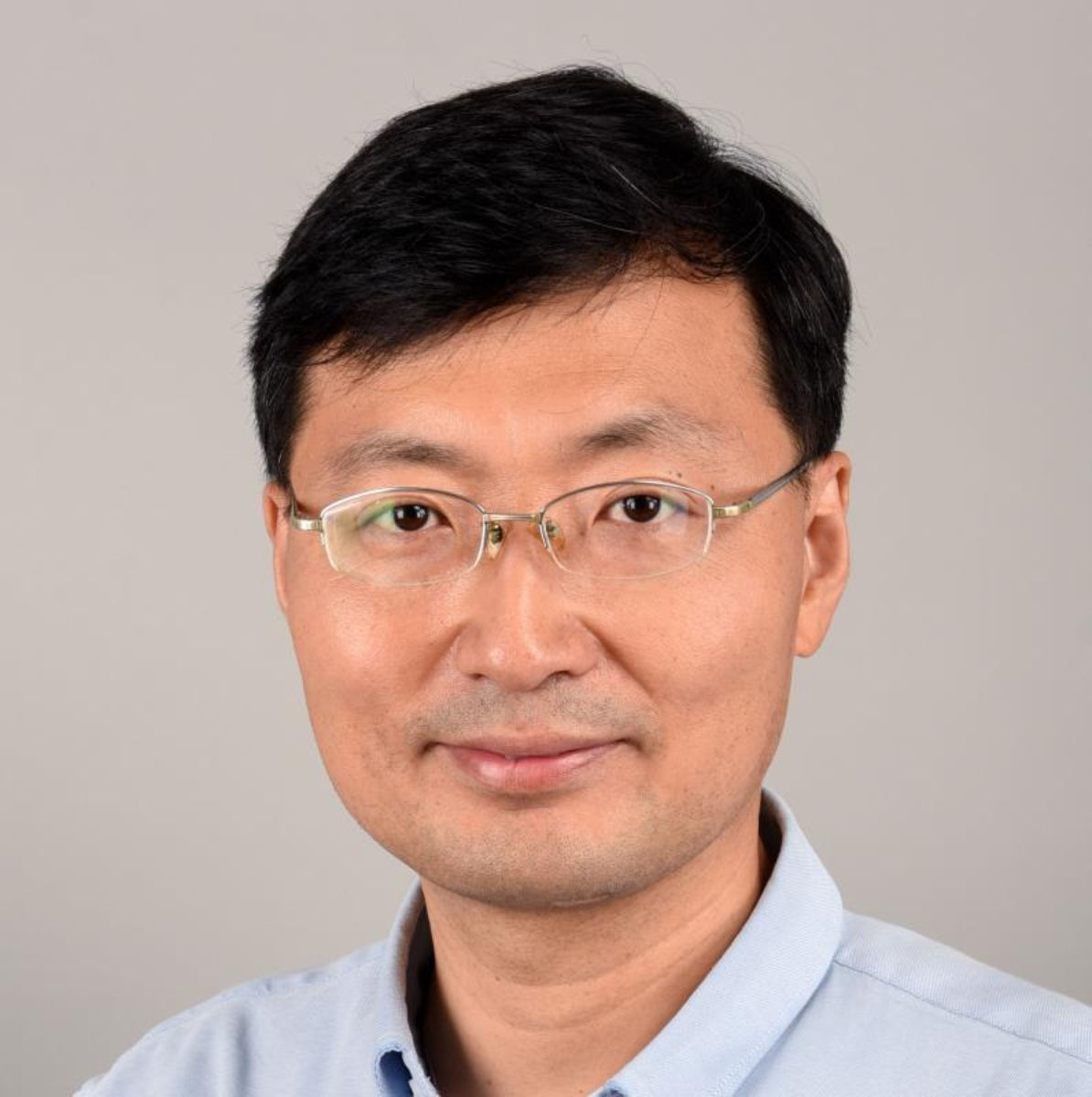}}]{Ming Yang}
received Master and Ph.D. degrees from Tsinghua University, Beijing, China, in 1999 and 2003, respectively.

He is currently the Full Tenure Professor at Shanghai Jiao Tong University, the deputy director of the Innovation Center of Intelligent Connected Vehicles.
He has been working in the field of intelligent vehicles for more than 20 years. He participated in several related research projects, such as the THMRV project (first intelligent vehicle in China), European CyberCars and CyberMove projects, CyberC3
project, CyberCars-2 project, ITER transfer cask project, AGV, etc.
\end{IEEEbiography}

\end{document}